\documentclass[journal]{IEEEtran}

\usepackage{cite}

\usepackage{microtype}
\usepackage{graphicx}
\usepackage{booktabs}
\usepackage{multirow}
\usepackage{array}
\usepackage{amsmath}
\usepackage{amssymb}
\usepackage{amsfonts}
\usepackage{amsthm}
\usepackage{mathtools}
\usepackage{textcomp}
\usepackage{xspace}
\usepackage{url}
\usepackage{adjustbox}
\usepackage{rotating}
\usepackage[para,online,flushleft]{threeparttable}
\usepackage{bbding}
\usepackage{makecell}
\usepackage{subfigure}
\usepackage[linesnumbered,ruled,lined]{algorithm2e}
\usepackage[colorlinks,linkcolor=black,citecolor=black,urlcolor=black]{hyperref}

\newcommand{\method}{\textbf{CEP}\xspace}

\newcommand{\fires}[1]{\textbf{#1}}
\newcommand{\secres}[1]{\underline{#1}}

\def\Eqref#1{Equation~\eqref{#1}}

\newcommand{\x}[0]{\mathbf{x}}
\newcommand{\y}[0]{\mathbf{y}}
\newcommand{\D}[0]{\mathcal{D}}

\newcommand{\z}[0]{\mathbf{z}}

\theoremstyle{plain}

\theoremstyle{definition}

\theoremstyle{remark}

\begin{document}

\title{Continuous Evolution Pool: \\ Taming Recurring Concept Drift in Online Time Series Forecasting}

\author{
    Tianxiang Zhan,
    Ming Jin,
    Yuanpeng He,
    Yuxuan Liang,
    and~Shirui~Pan
    \thanks{T. Zhan, M. Jin and S. Pan are with Griffith University,
        Nathan, QLD, Australia (e-mail: ztxtech.time@gmail.com;
        ming.jin@griffith.edu.au; shirui.pan@griffith.edu.au).}
    \thanks{Y. He is with Peking University, Beijing, China
        (e-mail: heyuanpeng@pku.edu.cn).}
    \thanks{Y. Liang is with the Hong Kong University of
        Science and Technology (Guangzhou), Guangzhou, China
        (e-mail: yuxuan.liang@hkust-gz.edu.cn).}
    \thanks{Corresponding authors: Ming Jin and Shirui Pan.}
    \thanks{The source code is available at \url{https://github.com/ztxtech/cep_ts}.}
}

\maketitle

\begin{abstract}
    Recurring concept drift is pervasive in real-world online time series, where the underlying data-generating process repeatedly alternates between a small set of regimes, most notably daily or seasonal cycles that dominate energy, traffic, and weather patterns, and is therefore a central obstacle to reliable long-horizon forecasting. This problem poses a dual challenge in online time series forecasting: mitigating catastrophic forgetting while operating under strict constraints that prevent storing or replaying historical raw samples. Existing approaches predominantly rely on parameter updates or experience replay, which inevitably suffer from knowledge overwriting or stale replay buffers. To address this, the Continuous Evolution Pool (\method), a replay-free framework that maintains a dynamic pool of specialized forecasters, is proposed. Instead of storing raw samples, \method utilizes lightweight statistical genes to decouple concept identification from forecasting. Specifically, it employs a retrieval mechanism to identify the nearest concept based on gene similarity, an evolution strategy to spawn new forecasters upon detecting distribution shifts, and an elimination policy to prune obsolete models under memory constraints. Experiments on real-world datasets demonstrate that \method significantly outperforms state-of-the-art baselines, reducing forecasting error by up to $24\%$ on datasets with pronounced recurring drift without accessing historical ground truth.
\end{abstract}


\section{Introduction}

\IEEEPARstart{T}{ime} series forecasting, a fundamental task in fields such as finance, energy management, traffic prediction~\cite{ijcai2022p534}, and environmental monitoring, empowers better decision-making, optimized resource allocation, and effective risk management~\cite{jin2024survey}. In many of these applications, data does not arrive as a static, complete dataset but as a continuous stream, which gives rise to the problem of online time series forecasting: the model must predict future values as new observations arrive, and update itself incrementally without retraining from scratch. The central difficulty of this setting is concept drift, which refers to the fact that the relationship between input variables and their true values changes over time~\cite{ditzler2015learning,10.1145/2523813}, so a model trained on past data gradually becomes outdated.

Research on online forecasting under concept drift has progressed through several stages, each addressing part of the problem while leaving a new limitation behind. Early work built on online convex optimization and shallow models~\cite{Wu2007Top1A,read2018concept}, which adapt quickly but lack the capacity to model complex temporal patterns. In the broader data-stream community, drift detectors such as ADWIN~\cite{bifet2009adaptive} focused on signaling when a distribution change occurs, online learning of temporal association rules~\cite{10623290} tracked recurring patterns in multivariate streams, while ensemble classifiers like the Adaptive Random Forest~\cite{gomes2017adaptive} and dynamically expanding continual learning methods~\cite{elwell2011incremental,kim2024quilt,GONCALVESJR20131018} maintained and grew pools of models for classification tasks. These approaches demonstrated the value of explicit drift awareness and model diversity, but their mechanisms are tied to classification or single-step prediction and do not transfer directly to multi-step time series forecasting, where long-range temporal dependencies dominate. A second line of work imported techniques from continual learning~\cite{9349197} into forecasting, most notably Experience Replay (ER)~\cite{er}, its distillation-based variant DER++~\cite{derpp}, which store historical samples in a buffer and interleave them with new data to mitigate forgetting. However, replay buffers are bounded in size and quickly become stale under drift, and storing raw samples is simply not permitted in replay-free deployment scenarios where historical data cannot be retained. More recently, deep architectures designed specifically for online forecasting have emerged. FSNet~\cite{fsnet}, inspired by the Complementary Learning Systems theory, couples a slowly-learned backbone with per-layer adapters and an associative memory to balance fast adaptation against recall of old patterns. OneNet~\cite{onenet} instead maintains two branches modeling cross-time and cross-variable dependencies and ensembles them online with dynamically adjusted weights. Following this line, ODGNet~\cite{11459376} learns dynamic variable associations to track evolving dependencies in the stream, and TOE~\cite{11626574} tackles distribution shifts with a three-perspective online ensemble that fuses complementary forecasting signals. Parallel efforts have also explored normalizing or decomposing the input to make models inherently more robust to distribution shift, such as RevIN~\cite{revin}, AdaRNN~\cite{adarnn}, and Non-stationary Transformers~\cite{liu2023nonstationarytransformersexploringstationarity}. Beyond accuracy-oriented adaptation, uncertainty-aware online multi-step forecasting has also been developed for cloud systems~\cite{11435627}. In parallel, Jhin et al.~\cite{jhin2024addressing} showed that the evaluation protocol used by these methods is subtly unfair, because models were given partial access to the ground truth of the next step, and introduced the delayed-feedback setting that is also adopted in this work.

\begin{figure*}[!t]
    \centering
    \includegraphics[width=0.9\linewidth]{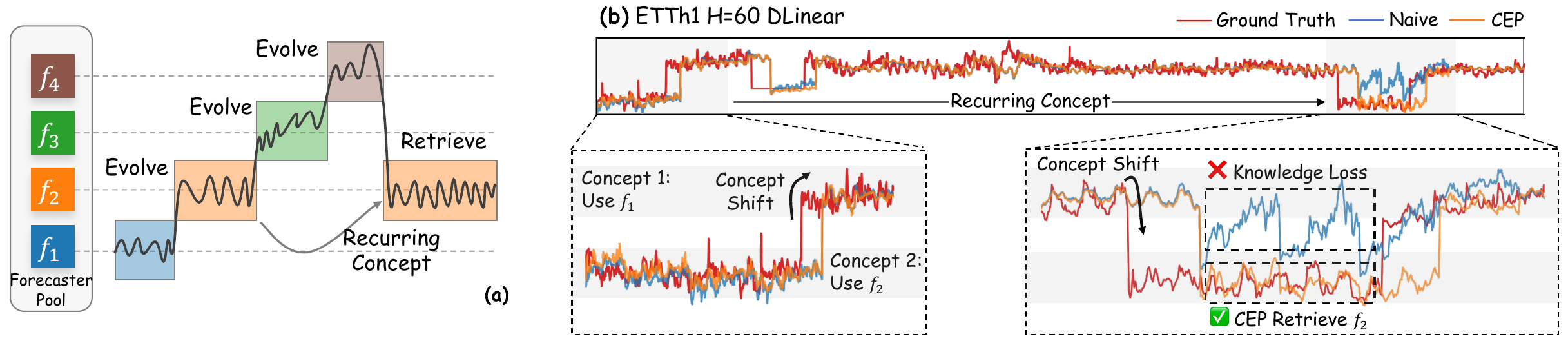}
    \caption{Concept recurrence is a notable phenomenon in time series data. For instance, in (a), the second concept recurs following the emergence of the fourth concept. During online updates and concept drifts, forecasters often suffer from forgetting previously learned concepts, as in (b). To address this challenge and effectively track recurring concepts, the Continuous Evolution Pool employs an evolving and retrieval strategy, assigning distinct forecasters to manage different concepts.}
    \label{fig:recurring}
\end{figure*}

Despite this progress, a fundamental gap remains, and it becomes most visible under recurring concept drift~\cite{alippi2013just}, a prevalent form of drift in which previously observed patterns reappear after a period of absence, as shown in Figure~\ref{fig:recurring}(a). Seasonal electricity consumption is a canonical example: similar consumption patterns recur annually or monthly, and these recurring patterns often encapsulate sensitive user behaviors, so historical raw samples cannot be stored or replayed when analyzing such recurrence~\cite{ijcai2017p504,johansson2025privacy}. Under this setting, all of the above paradigms share a common structural weakness. Parameter-updating methods such as FSNet and OneNet continuously overwrite a single set of weights, so knowledge of an old concept is gradually erased during its absence and must be relearned from scratch when it recurs. Replay-based methods such as ER and DER++ in principle retain past knowledge, but their buffers are too small to cover all concepts, become outdated under delayed feedback, and are disallowed outright under replay-free constraints. Neither paradigm maintains an explicit, addressable knowledge base at the level of individual concepts. To close this gap, an online forecasting framework must address two questions simultaneously: how to filter macro distribution shifts without overwriting specialized knowledge, and how to identify recurring concepts and manage them under bounded memory.

The key insight is that concept drift is, by definition, a shift in the data distribution; therefore, low-order statistical moments of the data are sufficient to address a concept, even though they are far from sufficient to model it. This suggests a division of labor: cheap statistics decide which concept the current sample belongs to, and dedicated neural forecasters handle how to predict within each concept. Based on this insight, the Continuous Evolution Pool (\method) is proposed, a pooling mechanism that maintains multiple forecasters, each specialized for one concept. \method summarizes every concept with a lightweight statistical gene, which is defined as online-updatable mean and variance and serves as the concept's address in a gene space. When a new test sample arrives, \method retrieves the forecaster whose gene is nearest to the sample's gene and uses it for prediction. If no existing forecaster is close enough, a distribution shift is declared, and a new forecaster evolves from the nearest one and is added to the pool, preserving knowledge of the emerging concept. An elimination mechanism removes forecasters that stay idle for too long, keeping memory bounded and filtering out noise. Unlike Time Series Foundation Models (TSFMs) that rely on massive pre-training resources~\cite{chronos}, \method focuses on learning from scratch in data isolated environments. The main contributions of this paper are as follows:

\begin{enumerate}
    \item The structural insufficiency of parameter-updating and replay-based paradigms under recurring concept drift with replay-free constraints is identified.
    \item A decoupled adaptation mechanism is designed, using lightweight statistics to filter macro distribution shifts while leaving micro-pattern learning to neural gradient descent.
    \item A gene-based identification-and-management scheme is proposed, enabling accurate detection of recurring concepts and bounded-memory pool management.
\end{enumerate}

This paper is organized as follows. Related work is reviewed in Section~\ref{sec:related}, and the gap that motivates the present study is identified. The architecture and the underlying mathematical principles of \method are then presented in Section~\ref{sec:method}. Section~\ref{sec:experiments} reports the experimental results, covering gradual and recurring concept drift, comparisons with base forecasters, online methods, and time series foundation models, as well as ablation and hyperparameter analyses. The paper concludes in Section~\ref{sec:conclusion} with a summary of the key findings. Complete per-dataset results and additional visualizations are deferred to the appendix.

\section{Related Work}\label{sec:related}

\subsection{Concept Drift}

Unlike conventional time series analysis, where concept drift might be less critical, it is a central concern in online time series forecasting. Suppose the input variables $x$ and their ground truth $y$ follow a distribution $p_{t}$ over time as in~\Eqref{eq:xy_dis}. For two different time points $t_{1}$ and $t_{2}$, the distribution shift can be formally defined by~\Eqref{eq:xy_cd}~\cite{read2018concept}.

\begin{equation}\label{eq:xy_dis}
    (x,y) \sim p_{t}(x,y)
\end{equation}

\begin{equation}\label{eq:xy_cd}
    p_{t_{1}}(y | x) \neq p_{t_{2}}(y | x)
\end{equation}

Concept drift was formally defined~\cite{tsymbal2004problem}, and emphasized for continuous adaptation in data streams~\cite{read2018concept}. Subsequent work includes the AdaRNN~\cite{adarnn}, which struggled with complex relationships, RevIN~\cite{revin} designed for distribution shift, and the exploration into online drift detection~\cite{wan2024online}. Despite these advancements, significant challenges remain in online time series forecasting, necessitating continuous research.

\subsection{Recurring Concept Drift}

Building on the formulation in~\Eqref{eq:xy_dis} and~\Eqref{eq:xy_cd}, recurring concept drift describes the case in which a distribution that appeared at an earlier time becomes active again after a period of absence. Formally, let the active concept at time $t$ be drawn from a set of distinct concepts $\{C_1, C_2, \ldots\}$, each associated with a stationary distribution $p_{C_k}(\x, \y)$. Recurring drift occurs when the sequence of active concepts revisits a previously seen concept, i.e., there exist time points $t_1 < t_2 < t_3$ such that $C(t_1) = C(t_3) \neq C(t_2)$. This pattern is prevalent in real-world time series, where seasonal or periodic factors cause similar distributions to reappear cyclically~\cite{alippi2013just}.

The recurring setting imposes a requirement that neither abrupt nor gradual drift does: the model must not only adapt to the current concept but also retain the knowledge of previously encountered concepts so that they can be reused immediately upon recurrence. Parameter-updating methods overwrite a single set of weights and therefore forget old concepts during their absence. Replay-based methods retain raw samples, but their buffers are too small to cover all concepts and are disallowed under replay-free constraints. This gap motivates a mechanism that maintains an explicit, addressable knowledge base at the level of individual concepts.

\subsection{Mixture of Experts and Forecaster Assignment}

A natural candidate for maintaining such a concept-level knowledge base is the Mixture of Experts (MoE) paradigm~\cite{10.1162/neco.1991.3.1.79,moe}, which keeps a set of specialized expert networks and routes each input to a subset of them through a learned gating function. At first glance, MoE appears to offer exactly what recurring concept drift requires, since different experts could specialize in different concepts and the gate could reactivate the relevant expert when a concept recurs. Formally, the MoE output for an input $\x$ is the gated combination shown in~\Eqref{eq:moe_out}.
\begin{equation}
    \hat{\y}_{\text{MoE}}(\x) = \sum_{i=1}^{N} g_i(\x) \cdot f_i(\x)
    \label{eq:moe_out}
\end{equation}
where $f_i$ is the $i$-th expert and $g_i(\x) \in [0, 1]$ is the gating weight with $\sum_i g_i(\x) = 1$.

However, standard MoE architectures are designed for a stationary training distribution, not for online adaptation under drift. Two structural limitations arise. First, the experts are predefined at training time, so a MoE cannot create a new expert when a previously unseen concept emerges. Second, the predominant soft-gating mechanisms, which include Softmax Gating and Noisy Top-K Gating, update all experts with non-zero weights on every sample. Under recurring drift, this means an expert that learned a seasonal concept is continuously modified by gradients from the currently active, unrelated concept. The per-step parameter update for expert $f_i$ takes the form of~\Eqref{eq:moe_grad}, which exposes the gradient pollution problem.
\begin{equation}
    \theta_i \leftarrow \theta_i - \eta \, g_i(\x_t) \, \nabla_{\theta_i} \mathcal{L}(f_i(\x_t), \y_t)
    \label{eq:moe_grad}
\end{equation}
Because $g_i(\x_t) > 0$ for multiple experts on every sample in soft gating, an expert specialized in a latent or inactive concept receives a non-zero gradient from the current concept, which gradually erases the specialized knowledge that expert was supposed to preserve.

\section{Methodology}\label{sec:method}

\begin{figure*}[!t]
    \centering
    \includegraphics[width=0.95\linewidth]{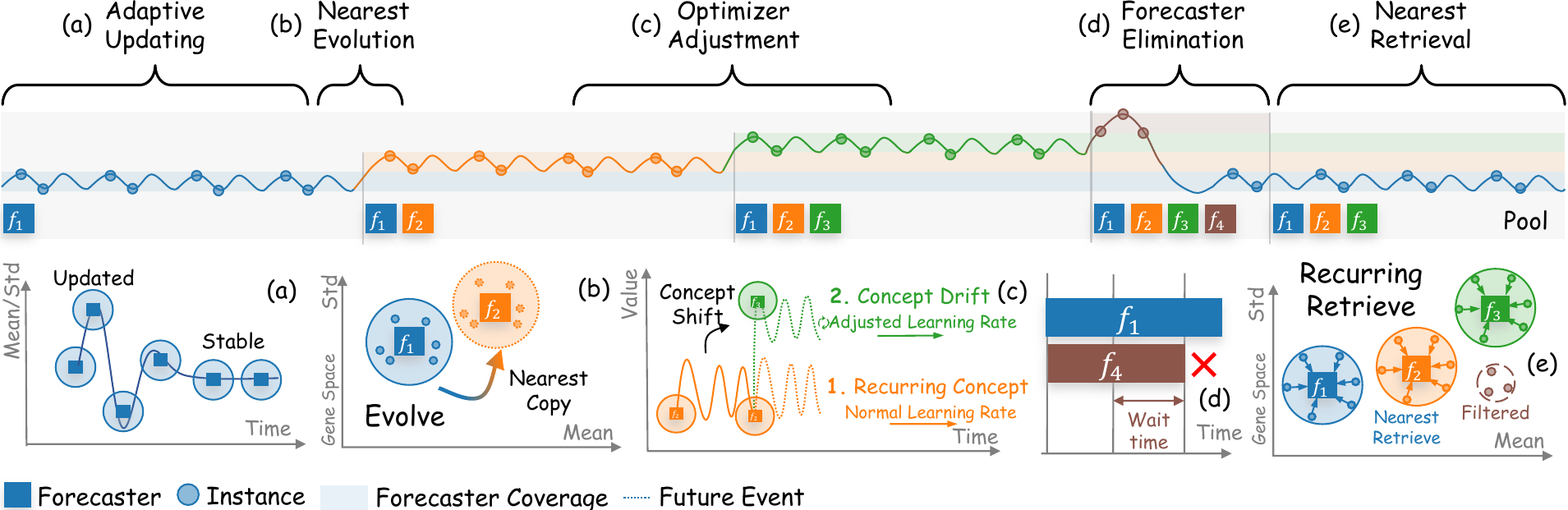}
    \caption{Illustration of the proposed \method mechanism:  (a) \textbf{Adaptive Updating}: Continuously update the position of the evolved forecaster in the gene space as new instances arrive. (b) \textbf{Nearest Evolution}: If the instance surpasses the forecaster's evolution threshold, evolve and replicate the closest forecaster at the instance. (c) \textbf{Optimizer Adjustment}: Adjust the learning rate for the shifted concept to ensure accurate adaptation. (d) \textbf{Forecaster Elimination}: The forecaster $f_4$ associated with the rarely-occurring noise concept may be removed due to prolonged inactivity. (e) \textbf{Nearest Retrieval}: Identify the nearest forecaster in the gene space when an input instance is encountered.}
    \label{fig:evo_details}
\end{figure*}

\begin{figure}[!t]
    \centering
    \includegraphics[width=0.9\linewidth]{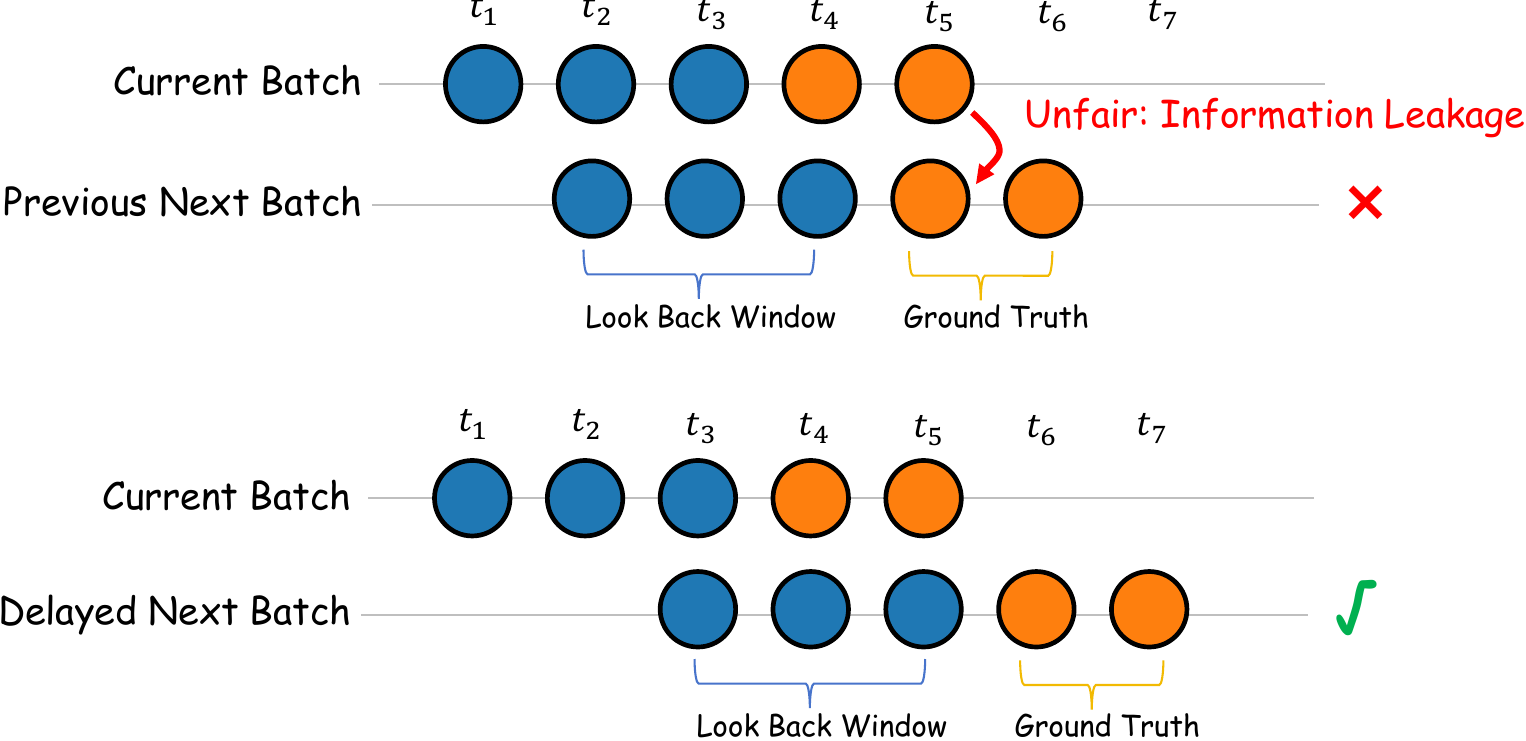}
    \caption{Delayed feedback setting. The upper case is unfair: the previous setting lets the model see part of the next ground truth beforehand. The lower delayed setting removes this lookahead and requires predicting strictly future values.}
    \label{fig:delay}
\end{figure}

In this section, the proposed \method is introduced. It helps partition samples from different distributions, enabling the updating of different models for evolution. The dynamic evolution mechanism for the recurring concept scenario is described as follows.

\subsection{Data Format and Problem Formulation}

A univariate time series is represented as $x = (x_{1}, x_{2}, \ldots, x_{N})$. Consider a look-back window of length $\mathbf{L}$, which is denoted as $\x_i = (x_i, x_{i+1},\ldots,x_{i+\mathbf{L}-1})$. The corresponding ground truth $\y_i$ with a forecasting horizon of $\mathbf{H}$ is defined as $\y_i=(x_{i + \mathbf{L}}, x_{i+\mathbf{L}+1},\ldots,x_{i+\mathbf{L}+\mathbf{H}-1})$. Each pair of $\x_i$ and $\y_i$ forms an instance pair expressed as $\D_i = (\x_i, \y_i)$.

In contrast to traditional time series forecasting, online forecasting is partitioned into two distinct stages: (1) the warm-up stage and (2) the online stage. Once the dataset $x$ is split into these two stages, sample pairs are constructed for each of them. The warm-up stage setting is consistent with traditional time series forecasting. In the online stage, the step size of the instance pair is set as forecasting horizon in Figure~\ref{fig:delay}. This setting effectively prevents the forecasting of the ground truth in the overlapping parts, which is in line with the setup of delayed feedback adopted in previous research~\cite{fsnet, onenet}. The batch size and epoch for the online stage are both set to $1$. For univariate time series without auxiliary information from other channels, it is necessary to focus on the conceptual changes of the single channel itself. Previous research~\cite{onenet} has shown that time-channel independent online forecasting can reduce interference from too many channels. The complete workflow of \method is presented in Algorithm~\ref{alg:cep}.

\begin{algorithm}[t]
    \caption{Continuous Evolution Pool (\method)}
    \label{alg:cep}
    \SetKwInput{KwInput}{Input}
    \SetKw{KwEvolve}{\textbf{Evolve:}}
    \SetKw{KwRetrieve}{\textbf{Retrieve:}}
    \DontPrintSemicolon
    \KwInput{Dataset $x$, Forecaster $f$}
    $\D_{warm}, \D_{online} \leftarrow \text{split}(x)$ \tcp*{Problem Formulation}
    Initialize $Pl = \{(f(\theta_1), \z_1)\}$ \;
    \BlankLine
    \For{$\D_i \in \D_{warm}$ \tcp*{$\D_i = (\x_i, \y_i)$}}{
        $\z_x = \text{Gene}(\x_i)$ \tcp*{\Eqref{eq:x_gene}}
        $\mathcal{L} = \text{Loss}(f(\x_i, \theta_1),\y_i)$ \;
        $\theta_1 \leftarrow \theta_1 - lr \cdot \nabla \mathcal{L}$ \;
        $\z_1 \leftarrow \text{Update}(\z_1, \z_x)$ \tcp*{Update gene, \Eqref{eq:f_g}-\Eqref{eq:f_gg}}
    }
    \BlankLine
    \For{$\D_i \in \D_{online}$ \tcp*{$\D_i = (\x_i, \y_i)$}}{
        $\z_x = \text{Gene}(\x_i)$ \tcp*{\Eqref{eq:x_gene}}
        $(f(\theta_n), \z_n) = \text{Nearest}(Pl, \z_x)$ \tcp*{\Eqref{eq:d_g}, \Eqref{eq:f_select}} \;
        \eIf{$\text{Evolution}(\z_x, \z_n)$ \tcp*{\Eqref{eq:mean_thre}}}{
            \KwEvolve \;
            $\theta_c = \text{Copy}(\theta_n)$ \;
            $Pl \leftarrow Pl \cup \{(f(\theta_c), \z_x)\}$ \;
            $lr \leftarrow \text{Adjust}(lr)$ \tcp*{Adjust optimizer, \Eqref{eq:lr_dec}}
        }{
            \KwRetrieve \;
            $\theta_c = \theta_n$ \;
        }
        $\hat{\y}_i = f(\x_i, \theta_c)$ \;
        $\z_y = \text{Gene}(\y_i)$ \tcp*{\Eqref{eq:x_gene}} \;
        \If{not $\text{Evolution}(\z_y, \z_n)$ \tcp*{Prevent gradient pollution}}{
            $\mathcal{L} = \text{Loss}(\hat{\y}_i, \y_i)$ \;
            $\theta_c \leftarrow \theta_c - lr \cdot \nabla \mathcal{L}$ \;
            $\z_c \leftarrow \text{Update}(\z_c, \z_x)$ \tcp*{\Eqref{eq:f_g}-\Eqref{eq:f_gg}}
        }
        $Pl \leftarrow \text{Eliminate}(Pl)$ \tcp*{\Eqref{eq:f_e}}
    }
\end{algorithm}

\subsection{Warm-Up Stage}

\subsubsection{Pool Initialization}

A pool set $Pl = \{(f(\theta_1), \z_1)\}$ is first initialized with a single element. Here, $f(\theta_1)$ represents the initial forecaster with parameter $\theta_1$, and $\z_1$ is the corresponding gene vector that stores the features of the instances. The gene, serving as a statistical signature for the underlying data distribution, is composed of two parts with a gene ratio $\tau_{g}$, as defined in~\Eqref{eq:f_g}: the global gene vector $\z_{g}$ and the local gene vector $\z_{l}$. The global gene $\z_{g}$ is designed to capture and record long-term macroscopic features, which are crucial for forecasting. In contrast, the local gene $\z_{l}$ is responsible for documenting short-term features that are strongly correlated with the current time. This design adopts a decoupled strategy: genes capture high-level regime shifts, ensuring the forecaster focuses on learning complex dependencies without being distracted by statistical noise. During the warm-up stage, which has the same task as typical time series forecasting, \method employs the initial forecaster to make forecasts and accumulate knowledge for the online stage with recurring concepts.

\begin{equation} \label{eq:f_g}
    \z =  \tau_{g} \cdot \z_{l} + (1-\tau_{g}) \cdot \z_{g}
\end{equation}

\subsubsection{Adaptive Updating}

There are several necessary conditions for the gene implementation:
\begin{itemize}
    \item In the absence of historical samples, genes can be dynamically updated.
    \item The distances between genes are statistically grounded.
    \item Gene computation is highly efficient.
\end{itemize}

Therefore, the gene of each instance $\x$ is designed to use statistical moments in~\Eqref{eq:x_gene}, where $\mu$ is the mean operator, $\sigma$ is the standard deviation operator, $S$ limits the look-back window scope. This is a simple, yet effective approach, instead of considering more complex features such as spectral domain or periodic patterns. While spectral features could offer richer representations, this is hard to dynamically update under the previous conditions. The genes are adaptively updated as instances arrive, stabilizing over time in Figure~\ref{fig:evo_details}(a).

Local genes are updated by Exponential Moving Average (EMA) with ratio $\tau_l$ for short-term adaptation in~\Eqref{eq:f_gl}, while global genes employ statistical formulas for overall trend estimation in~\Eqref{eq:f_gg}. EMA introduces an implicit soft transition, allowing the forecaster to smoothly track gradual drifts before a distinct concept split is triggered. The statistical basis of both updates is detailed in Section~\ref{sec:gene_foundation}.

\begin{equation}\label{eq:x_gene}
    \z_x = \text{Gene}(x) = (\mu(\x[-S:]), \sigma (\x[-S:]))
\end{equation}

\begin{equation} \label{eq:f_gl}
    \z_{l} \leftarrow  \tau_l  \cdot  \z_x + (1 - \tau_l) \cdot  \z_{l}
\end{equation}

\begin{equation} \label{eq:f_gg}
    \resizebox{0.95\linewidth}{!}{$
            \z_{g} \leftarrow \left(\frac{n \cdot \z_{g, \mu} + \z_{x,\mu}}{n+1}, \sqrt{\frac{n}{n+1}\z_{g, \sigma}^2 + \frac{n}{(n+1)^2}(\z_{g, \mu} - \z_{x, \mu})^2} \right)
        $}
\end{equation}

Additionally, adaptive updating is not only limited to the warm-up stage but also applies to the online stage. In the case of continuous concept drift shown in Section~\ref{sec:gradual_drift}, the forecaster can adapt naively. This prevents the unnecessary evolution of additional forecasters, which would otherwise cause sample dispersion and hinder effective training. The updating behaviour can enable the forecaster to follow the slight and continuous changes in the data.

\subsection{Statistical Foundation of Gene Representation}\label{sec:gene_foundation}

The gene retains the first two moments of each input window because they can be updated without retaining historical raw samples. Let $a_i=\z_{x,\mu}^{(i)}$ be the mean of the $i$-th instance gene. The global mean $m_n$ and population variance $v_n$ after $n$ assigned instances are $m_n=\frac{1}{n}\sum_{i=1}^{n}a_i$ and $v_n=\frac{1}{n}\sum_{i=1}^{n}(a_i-m_n)^2$. The two online recurrences are proved below.

For the mean, adding the new instance mean to the previous sum gives
\begin{align}
    m_{n+1}
     & = \frac{1}{n+1}\left(\sum_{i=1}^{n}a_i+a_{n+1}\right) \notag \\
     & = \frac{n m_n+a_{n+1}}{n+1}.
    \label{eq:mean_update_proof}
\end{align}
For the variance, define the second raw moment $M_{2,n}=v_n+m_n^2$. Since $M_{2,n+1}=\frac{nM_{2,n}+a_{n+1}^2}{n+1}$, substitution of the mean update gives
\begin{align}
    v_{n+1}
     & = M_{2,n+1}-m_{n+1}^2 \notag                  \\
     & = \frac{n(v_n+m_n^2)+a_{n+1}^2}{n+1}
    -\left(\frac{n m_n+a_{n+1}}{n+1}\right)^2 \notag \\
     & = \frac{n}{n+1}v_n+
    \frac{n}{(n+1)^2}(m_n-a_{n+1})^2.
    \label{eq:variance_update_proof}
\end{align}
The initialization is $m_1=a_1$ and $v_1=0$. Equations~\Eqref{eq:mean_update_proof} and~\Eqref{eq:variance_update_proof} give the global update in~\Eqref{eq:f_gg}. The global gene stores only the current count, mean, and standard deviation of the instance-window mean sequence. Updating this state costs $O(1)$ time and memory per assigned instance. Computing $\z_x$ from a length-$S$ window costs $O(S)$ when the window statistics are recomputed directly, or $O(1)$ when rolling sums are maintained. The local gene complements this global estimate through EMA, so the combined gene retains both persistent and recent moment information.

The moment representation also gives a statistical motivation for concept comparison. Suppose the observations in an arriving window come from a single concept $C_k$ that follows a Gaussian $\mathcal{N}(\mu_k,\sigma_k^2)$. Let the window contain $n$ scalar observations $x_1,\ldots,x_n$, and let $\hat{\mu}_\x=\frac{1}{n}\sum_{i=1}^n x_i$ and $\hat{\sigma}_\x^2=\frac{1}{n}\sum_{i=1}^n (x_i-\hat{\mu}_\x)^2$ denote their sample mean and population variance. These two quantities are exactly the components of the gene $\z_x$, so the likelihood of the window under $C_k$ can be written using only the gene.

To connect the likelihood to the moments, the sum of squared deviations from $\mu_k$ is split around the sample mean $\hat{\mu}_\x$. Writing $x_i-\mu_k=(x_i-\hat{\mu}_\x)+(\hat{\mu}_\x-\mu_k)$ and expanding the square, the cross term vanishes because $\sum_{i=1}^n(x_i-\hat{\mu}_\x)=0$, which gives the identity in~\Eqref{eq:sum_sq_decomp}.

\begin{equation}
    \resizebox{0.95\linewidth}{!}{$
            \begin{aligned}
                \sum_{i=1}^n(x_i-\mu_k)^2
                 & = \sum_{i=1}^n(x_i-\hat{\mu}_\x)^2
                +2(\hat{\mu}_\x-\mu_k)\sum_{i=1}^n(x_i-\hat{\mu}_\x)
                +n(\hat{\mu}_\x-\mu_k)^2                         \\
                 & = n\hat{\sigma}_\x^2+n(\hat{\mu}_\x-\mu_k)^2.
            \end{aligned}
        $}
    \label{eq:sum_sq_decomp}
\end{equation}

The per-sample negative log-likelihood of the window under $C_k$ can then be written in terms of the gene alone, as in~\Eqref{eq:gene_nll_deriv}.

\begin{align}
    \mathcal{J}_k
     & = -\frac{1}{n}\sum_{i=1}^n\log\mathcal{N}(x_i\mid\mu_k,\sigma_k^2) \notag \\
     & = \frac{1}{2}\log(2\pi)+\log\sigma_k
    +\frac{1}{2n\sigma_k^2}\sum_{i=1}^n(x_i-\mu_k)^2 \notag                      \\
     & = \frac{1}{2}\log(2\pi)+\log\sigma_k
    +\frac{\hat{\sigma}_\x^2+(\hat{\mu}_\x-\mu_k)^2}{2\sigma_k^2},
    \label{eq:gene_nll_deriv}
\end{align}
where the second step substitutes the Gaussian density and the third step applies~\Eqref{eq:sum_sq_decomp} with $\frac{1}{n}\sum_{i=1}^n(x_i-\hat{\mu}_\x)^2=\hat{\sigma}_\x^2$. Dropping the constant term $\frac{1}{2}\log(2\pi)$, which does not depend on $k$, gives the compact objective in~\Eqref{eq:gene_nll}.

\begin{equation}
    \mathcal{J}_k=
    \log\sigma_k+
    \frac{\hat{\sigma}_\x^2+(\hat{\mu}_\x-\mu_k)^2}{2\sigma_k^2}.
    \label{eq:gene_nll}
\end{equation}

This expression motivates retaining both location and scale in a concept summary. It does not make the Euclidean distance in~\Eqref{eq:d_g} identical to the Gaussian likelihood: the latter also weights deviations by $\sigma_k$ and includes $\log\sigma_k$. Moreover, the Gaussian model is a working approximation, while the global gene tracks the cross-instance variation of window means. \method therefore uses gene-space Euclidean distance as a lightweight bounded-memory retrieval heuristic, whose empirical performance is evaluated in Section~\ref{sec:assignment_analysis}.

\subsection{Online Stage}\label{sec:online_stage}

\subsubsection{Nearest Retrieval}

For a new instance $\x$ with gene $\z_x$, the closest forecaster is retrieved using the Euclidean distance in~\Eqref{eq:d_g} and~\Eqref{eq:f_select}, whose moment-level motivation is given in Section~\ref{sec:gene_foundation}. The assignment is sparse by design. Soft-gating ensembles update multiple experts with non-zero weights, so an expert specialized for an inactive concept can be altered by the current concept. \method activates only the nearest forecaster and keeps inactive forecasters isolated, allowing each model to remain specialized until its concept recurs. The EMA update provides continuity before a discrete switch, while sparse retrieval prevents gradient pollution after the switch. The comparison with differentiable gating mechanisms is reported in Section~\ref{sec:assignment_analysis}. Euclidean distance is deliberately chosen instead of shape-based metrics such as Dynamic Time Warping to stay replay-free, since raw historical data cannot be retained, making gene-space retrieval a necessary proxy.

\begin{equation}\label{eq:d_g}
    d(\z_x,\z)=\left\lVert \z_x -\z \right\rVert_2
\end{equation}

\begin{equation}\label{eq:f_select}
    (f(\theta_N), \z_N) = \underset{(f(\theta_i), \z_i) \in Pl}{\text{argmin}} \ d(\z_x, \z_i)
\end{equation}

\subsubsection{Nearest Evolution}

Given that the instances in online time series forecasting arrive incrementally, the features of different concepts may exhibit instability during the initial stages. To ensure the stability of the gene, each forecaster in the pool $Pl$ is assigned a safety period $\tau_{safe}$. After this period, the standardized mean deviation in~\Eqref{eq:mean_thre} provides a proactive screening rule for a macro distribution shift, without waiting for delayed forecasting errors to accumulate. Under an approximately stable Gaussian concept, $\tau_{\mu}=3$ has the usual three-sigma interpretation; in \method it is used as a practical threshold rather than a calibrated hypothesis test because both the window statistic and the reference moments are estimated online. If the threshold is exceeded, the evolution process in~\Eqref{eq:evo} is initiated. The new forecaster copies the parameters of the nearest forecaster and receives the current input gene, as shown in Figure~\ref{fig:evo_details}(b). This initialization transfers reusable structure from the closest concept while reserving a separate parameter set for the new regime.

\begin{equation} \label{eq:mean_thre}
    |\z_{x,\mu} - \z_{N, \mu} | > \tau_{\mu} \cdot \z_{N,\sigma}
\end{equation}

\begin{equation} \label{eq:evo}
    Pl \leftarrow Pl \cup \{(f(\theta_c), \z_x)\}, \quad \theta_c = \text{Copy}(\theta_N)
\end{equation}

Detailed statistical justification is provided in Section~\ref{sec:gene_foundation}. Subsequently, the split forecaster will be tasked with forecasting the instance $\x$ and updating its parameters. It is important to note that in online forecasting, the normalization layer typically maintains the current concept. For \method, the evolution mechanism serves a similar purpose. On one hand, the evolution mechanism ensures that the samples learned by the forecasters in the pool belong to individual concepts. On the other hand, \method does not focus on the internal design of the forecaster, aiming to enhance generalization and accommodate a wider range of forecasters.

\subsubsection{Forecaster Elimination}

Forecasters that remain idle for extended periods are removed to mitigate errors from time series fluctuations and conserve storage, as illustrated in Figure~\ref{fig:evo_details}(d). A threshold $\tau_e$ triggers elimination when $n_{wait} > \tau_e \cdot n_{pred}$ in~\Eqref{eq:f_e}, where $n_{wait}$ is idle time which refers to the number of time points that have passed since the last prediction and $n_{pred}$ is total predictions. This mechanism optimizes the system's active working memory for edge deployment; removed concepts can theoretically be offloaded to hierarchical cold storage, such as a disk, to handle long-term recurrence without consuming runtime resources. This mechanism promptly removes erroneously evolved forecasters and prevents noise corruption. Strategies such as the First-in-First-out (FIFO) method can be employed to control the maximum forecasters in the pool.

\begin{equation}\label{eq:f_e}
    n_{wait} > \tau_e \cdot n_{pred}
\end{equation}

\subsubsection{Optimizer Adjustment and Gradient Abandonment}

For abrupt concept shifts in Figure~\ref{fig:evo_details}(c), the learning rate is initialized to $lr = \tau_{lr} \cdot lr_{raw}$ and then gradually restored toward $lr_{raw}$ by the exponential decay in~\Eqref{eq:lr_dec}, where $t_{lr}$ is the warming time. The $\max$ lower bound prevents the effective learning rate from drifting below the configured base $lr_{raw}$. This balances rapid adaptation with stable convergence.

\begin{equation}\label{eq:lr_dec}
    lr \leftarrow \max(lr_{raw},  \tau_{lr}^{-\frac{1}{t_{lr}}} \cdot lr)
\end{equation}

Complementarily, if the ground truth gene $\mathbf{g}_y$ triggers a split in~\Eqref{eq:mean_thre}, parameter updating is skipped to prevent concept contamination from anticipated distribution shifts, preserving forecaster specialization.

\subsection{Regret Analysis}\label{sec:regret}

To clarify why assigning each recurring concept its own forecaster helps, the regret of \method is sketched under a few simplifying assumptions. Regret measures how much more loss an online method accumulates than an oracle that always uses the best forecaster for the current concept~\cite{krichene2015hedge,LITTLESTONE1994212}. Assume the stream contains a finite number $K$ of distinct concepts that recur over time, each concept is stationary while active, the per-step loss is bounded by $L_{\max}$, and the genes of different concepts are well separated so that the active concept is identified correctly with high probability. Under these assumptions, the total regret of \method over $T$ steps can be decomposed into three parts in~\Eqref{eq:regret_decomp}.

\begin{equation}
    R_T^{\text{\method}} = R_{\text{id}} + R_{\text{est}} + R_{\text{mgmt}}
    \label{eq:regret_decomp}
\end{equation}

The three terms differ in how they scale with $T$, as summarized in~\Eqref{eq:regret_terms}.

\begin{align}
    R_{\text{id}}   & \le \epsilon_{\text{id}} \, T L_{\max}, \notag         \\
    R_{\text{est}}  & \le \sum_{k=1}^{K} c\sqrt{T_k} \le c\sqrt{K T}, \notag \\
    R_{\text{mgmt}} & \le K W L_{\max}.
    \label{eq:regret_terms}
\end{align}

For identification, let $\epsilon_{\text{id}}$ be the per-step probability of routing to the wrong forecaster. Each misrouted step costs at most $L_{\max}$, and there are $T$ steps, so $R_{\text{id}} \le \epsilon_{\text{id}} T L_{\max}$. Because the genes are well separated, $\epsilon_{\text{id}}$ decays exponentially with the gene gap, so this term is negligible in practice. For estimation, let $T_k$ be the number of steps on which concept $C_k$ is active, with $\sum_k T_k = T$. Each forecaster learns only on its own stationary concept, where online gradient methods~\cite{zinkevich2003online} give regret at most $c\sqrt{T_k}$ per concept; summing over the $K$ concepts and applying the concavity of the square root gives $R_{\text{est}} \le c\sqrt{K T}$. For management, each of the $K$ concepts pays a one-time warm-up cost when it first appears or is relearned after elimination, and each such event lasts at most $W$ steps bounded by the safety period, so $R_{\text{mgmt}} \le K W L_{\max}$, a constant independent of $T$. Substituting the three terms into~\Eqref{eq:regret_decomp} gives the overall bound in~\Eqref{eq:regret_bound}.

\begin{equation}
    R_T^{\text{\method}} \le c\sqrt{K T} + O(1)
    \label{eq:regret_bound}
\end{equation}

The dominant term grows only as $O(\sqrt{T})$. By contrast, a single model that updates all of its parameters on every sample forgets a concept each time the distribution switches, so its error on a recurring concept resets and its regret grows linearly as $\Omega(T)$. The pooling mechanism of \method therefore turns a linear regret into a sublinear one in the recurring concept drift setting.

\section{Experiments}\label{sec:experiments}

\subsection{Experiment Setting}

\subsubsection{Datasets}

\method is evaluated on a diverse array of real-world datasets, including the ECL dataset~\cite{ecl}, the ETT dataset~\cite{informer}, the Exchange dataset~\cite{autoformer}, the Traffic dataset~\cite{autoformer}, and the WTH dataset~\cite{autoformer}. The ETT dataset covers two years and monitors the oil temperature and six power load characteristics. Its four subsets, ETTh1, ETTh2, ETTm1, and ETTm2, log data hourly and at 15-minute intervals, respectively, providing multiple granularities for the analysis of electricity-related time series. The ECL dataset monitors the electricity consumption of 321 clients from 2012 to 2014, which facilitates the exploration of intricate consumption patterns. The WTH dataset monitors 11 climatic aspects at around 1,600 spots in the United States every hour from 2010 to 2013. The Traffic dataset, sourced from the California Department of Transportation, supplies hourly freeway occupancy rates in the San Francisco Bay Area, which is beneficial for traffic flow forecasting. The Exchange dataset, spanning from 1990 to 2016 and containing the daily exchange rates of eight countries, is instrumental in predicting the fluctuations in the foreign exchange market. All datasets are used with a single channel under recurring concept drift, and the statistics are summarized in Table~\ref{tab:data_stat}. The data is split between the warm-up stage and the online stage in a ratio of $25:75$.

\begin{table}[!ht]
    \centering
    \caption{Dataset statistics.}\label{tab:data_stat}

    \setlength{\tabcolsep}{25pt}
    \resizebox{\columnwidth}{!}{
        \begin{tabular}{lcc}
            \toprule
            Data     & Feature            & Timestep \\
            \midrule
            ECL      & MT\_104            & 26,304   \\
            ETTh1    & LULL               & 14,400   \\
            ETTh2    & LUFL               & 14,400   \\
            ETTm1    & LULL               & 14,400   \\
            ETTm2    & MUFL               & 14,400   \\
            Exchange & OT                 & 7,588    \\
            Traffic  & 16                 & 17,544   \\
            WTH      & DewPointFahrenheit & 35,064   \\
            \bottomrule
        \end{tabular}}
\end{table}

\subsubsection{Implementation Details}

The experiments were conducted on a computer equipped with an AMD Ryzen 9 7950X CPU with 16 cores at 5.1 GHz and 64 GB of memory. All experiments were executed using PyTorch~\cite{pytorch} with the AdamW optimizer~\cite{adam,10480574}. In all benchmarks, the length of the look-back window was set to $\mathbf{L}=60$, and the forecast horizon was varied as $\mathbf{H} \in \{30, 60\}$. The metric used for evaluation was Mean Squared Error (MSE). The default hyperparameters of \method are listed in Table~\ref{tab:default_hypara}.

\begin{table}[!ht]
    \centering
    \caption{Default hyperparameters of \method.}
    \label{tab:default_hypara}
    \setlength{\tabcolsep}{20pt}
    \resizebox{\columnwidth}{!}{
        \begin{tabular}{cccccc}
            \toprule
            Hyperparameter & $\tau_\mu$ & $\tau_{g}$ & $\tau_l$ & $\tau_{safe}$ & $\tau_e$ \\
            \midrule
            Value          & 3.0        & 0.8        & 0.2      & 15            & 1.5      \\
            \bottomrule
        \end{tabular}}
\end{table}

\subsubsection{Baselines}

Multiple methods across continual learning, time series forecasting, and online learning are evaluated. The Experience Replay (ER) method~\cite{er} stores historical data in a buffer and interleaves it with new samples during learning. DER++~\cite{derpp} incorporates knowledge distillation to enhance performance. FSNet~\cite{fsnet} utilizes a fast-slow learning mechanism to capture short-term changes and long-term patterns in online forecasting. OneNet~\cite{onenet} improves forecasting under concept drift through online ensembling, which combines multiple models to adapt to evolving data distributions. For time series forecasting, models based on various backbones are screened, including DLinear~\cite{dlinear}, TimesNet~\cite{timesnet}, PatchTST~\cite{patchtst}, SegRNN~\cite{segrnn}, iTransformer~\cite{itransformer}, and TimeMixer~\cite{timemixer}.

\subsection{Gradual Concept Drift}\label{sec:gradual_drift}

\begin{table*}[!ht]
    \centering
    \caption{MSE errors of gradual concept shift experiments; percentages in parentheses show the relative change. Best and second-best performances are highlighted.}
    \label{tab:full_p1}
    \setlength{\tabcolsep}{10pt}
    \renewcommand{\arraystretch}{1}
    \resizebox{\linewidth}{!}{
        \begin{threeparttable}
            \begin{small}
                \begin{tabular}{ccccccccccccccccc}
                    \toprule
                    \multirow{2}{*}{Data} & \multicolumn{2}{c}{ECL} & \multicolumn{2}{c}{ETTh1} & \multicolumn{2}{c}{ETTh2} & \multicolumn{2}{c}{ETTm1} & \multicolumn{2}{c}{ETTm2} & \multicolumn{2}{c}{Exchange} & \multicolumn{2}{c}{Traffic} & \multicolumn{2}{c}{WTH}                                                                                                     \\ \cmidrule{2-17}
                                          & MSE                     & Std                       & MSE                       & Std                       & MSE                       & Std                          & MSE                         & Std                     & MSE            & Std   & MSE            & Std   & MSE            & Std   & MSE            & Std   \\ \midrule
                    ER                    & 0.124                   & 0.004                     & 0.084                     & 0.001                     & \secres{0.693}            & 0.006                        & \secres{0.089}              & 0.000                   & \secres{0.050} & 0.000 & 0.189          & 0.025 & 0.606          & 0.361 & 0.091          & 0.015 \\
                    DER++                 & 0.107                   & 0.004                     & \secres{0.081}            & 0.001                     & \fires{0.663}             & 0.006                        & \fires{0.086}               & 0.000                   & \fires{0.048}  & 0.000 & 0.168          & 0.020 & 0.507          & 0.321 & 0.072          & 0.012 \\
                    FSNet                 & 0.208                   & 0.026                     & 0.091                     & 0.001                     & 0.767                     & 0.010                        & 0.123                       & 0.009                   & 0.062          & 0.001 & 1.175          & 0.425 & 0.618          & 0.065 & 0.096          & 0.004 \\
                    OneNet                & 0.109                   & 0.004                     & 0.084                     & 0.001                     & 0.737                     & 0.021                        & 0.115                       & 0.005                   & 0.058          & 0.002 & 0.248          & 0.321 & 0.303          & 0.015 & 0.065          & 0.002 \\ \midrule
                    TCN                   & \secres{0.070}          & 0.003                     & \fires{0.079}             & 0.001                     & 0.701                     & 0.006                        & 0.094                       & 0.001                   & 0.053          & 0.001 & \fires{0.033}  & 0.001 & \fires{0.190}  & 0.011 & \fires{0.045}  & 0.000 \\
                    +\method              & \secres{0.070}          & 0.003                     & \fires{0.079}             & 0.001                     & 0.701                     & 0.006                        & 0.094                       & 0.001                   & 0.053          & 0.001 & \fires{0.033}  & 0.001 & \fires{0.190}  & 0.011 & \fires{0.045}  & 0.000 \\ \midrule
                    TimesNet              & 0.096                   & 0.009                     & 0.110                     & 0.010                     & 0.888                     & 0.045                        & 0.185                       & 0.012                   & 0.093          & 0.010 & 0.077          & 0.009 & 0.329          & 0.028 & 0.077          & 0.005 \\
                    +\method              & 0.096                   & 0.009                     & 0.110                     & 0.010                     & 0.901                     & 0.071                        & 0.192                       & 0.014                   & 0.093          & 0.010 & 0.077          & 0.008 & 0.329          & 0.028 & 0.077          & 0.005 \\ \midrule
                    DLinear               & 0.160                   & 0.000                     & 0.109                     & 0.000                     & 1.026                     & 0.001                        & 0.218                       & 0.000                   & 0.069          & 0.000 & 0.077          & 0.000 & \secres{0.251} & 0.000 & 0.061          & 0.000 \\
                    +\method              & 0.160                   & 0.000                     & 0.109                     & 0.000                     & 1.026                     & 0.001                        & 0.223                       & 0.001                   & 0.069          & 0.000 & 0.077          & 0.000 & \secres{0.251} & 0.000 & 0.061          & 0.000 \\ \midrule
                    PatchTST              & 0.105                   & 0.013                     & 0.096                     & 0.007                     & 0.966                     & 0.108                        & 0.142                       & 0.019                   & 0.068          & 0.004 & 0.066          & 0.009 & 0.284          & 0.039 & 0.072          & 0.017 \\
                    +\method              & 0.105                   & 0.013                     & 0.095                     & 0.006                     & 0.993                     & 0.102                        & 0.151                       & 0.026                   & 0.071          & 0.008 & 0.067          & 0.007 & 0.284          & 0.039 & 0.067          & 0.015 \\ \midrule
                    iTransformer          & 0.126                   & 0.033                     & 0.117                     & 0.009                     & 0.956                     & 0.042                        & 0.239                       & 0.047                   & 0.089          & 0.007 & 0.083          & 0.010 & 0.308          & 0.015 & 0.108          & 0.015 \\
                    +\method              & 0.123                   & 0.023                     & 0.116                     & 0.009                     & 0.950                     & 0.050                        & 0.221                       & 0.052                   & 0.089          & 0.007 & 0.095          & 0.029 & 0.305          & 0.012 & 0.108          & 0.015 \\ \midrule
                    TimeMixer             & \fires{0.066}           & 0.005                     & 0.091                     & 0.005                     & 0.701                     & 0.018                        & 0.113                       & 0.007                   & 0.060          & 0.002 & 0.040          & 0.004 & 0.263          & 0.016 & \secres{0.050} & 0.003 \\
                    +\method              & \fires{0.066}           & 0.005                     & 0.091                     & 0.005                     & 0.702                     & 0.016                        & 0.114                       & 0.007                   & 0.060          & 0.002 & \secres{0.039} & 0.003 & 0.263          & 0.016 & 0.051          & 0.002 \\ \bottomrule
                \end{tabular}
            \end{small}
        \end{threeparttable}}
\end{table*}

To assess model adaptability to gradual distribution shifts, the forecast horizon is set to $\mathbf{H}=1$, so each input instance advances by only one time step and the overall input distribution changes mildly. Experiments were conducted with previous online methods and base forecasters with different common backbones, with \method additionally applied to the base forecasters. The results are presented in Table~\ref{tab:full_p1}. DER++~\cite{derpp} achieves the lowest MSE on the three ETT subsets, achieving $0.663$ on ETTh2 and $0.086$ on ETTm1, while the base forecasters TimeMixer and TCN lead on the remaining four datasets, reaching $0.066$ on ECL, $0.050$ on WTH, $0.033$ on Exchange, and $0.190$ on Traffic; FSNet~\cite{fsnet} and OneNet~\cite{onenet} consistently trail behind. Since all models produced low and acceptable errors, sophisticated techniques for handling gradual concept drift are unnecessary for short-term forecasting.

\subsection{Comparison with Base Forecasting Models}\label{sec:base_forecasting}

\begin{table*}[!h]
   \centering
   \caption{Averaged MSE errors of different forecasters when using \method; percentages in parentheses show the relative change after adding \method. Best and second-best performances are highlighted. Full results are presented in Appendix Table~\ref{tab:full_baseline}.}
   \label{tab:baseline}
   \setlength{\tabcolsep}{5pt}
   \renewcommand{\arraystretch}{1}
   \resizebox{\linewidth}{!}{
      \begin{threeparttable}
         \begin{small}
            \begin{tabular}{c|cccccccccccccc}
               \toprule
               Data                      & TimeMixer & +\method         & iTransformer & +\method         & PatchTST & +\method         & DLinear & +\method          & SegRNN & +\method         & TimesNet & +\method         & TCN   & +\method          \\ \midrule
               \multirow{2}{*}{ECL}      & 0.290     & 0.271            & 0.275        & 0.256            & 0.310    & 0.297            & 0.307   & 0.299             & 0.278  & 0.276            & 0.305    & 0.289            & 0.344 & 0.309             \\
                                         & -         & -6.62\% & -            & -6.98\% & -        & -4.29\% & -       & -2.60\%  & -      & -0.79\% & -        & -5.34\% & -     & -10.16\% \\ \midrule
               \multirow{2}{*}{ETTh1}    & 0.335     & 0.322            & 0.326        & 0.323            & 0.332    & 0.327            & 0.337   & 0.308             & 0.303  & 0.301            & 0.351    & 0.341            & 0.463 & 0.459             \\
                                         & -         & -3.91\% & -            & -0.80\% & -        & -1.75\% & -       & -8.70\%  & -      & -0.76\% & -        & -2.96\% & -     & -0.76\%  \\ \midrule
               \multirow{2}{*}{ETTh2}    & 2.630     & 2.508            & 2.566        & 2.542            & 2.594    & 2.458            & 2.718   & 2.630             & 2.628  & 2.584            & 2.676    & 2.560            & 3.166 & 2.804             \\
                                         & -         & -4.65\% & -            & -0.93\% & -        & -5.23\% & -       & -3.22\%  & -      & -1.66\% & -        & -4.37\% & -     & -11.44\% \\ \midrule
               \multirow{2}{*}{ETTm1}    & 0.755     & 0.737            & 0.760        & 0.740            & 0.759    & 0.721            & 0.849   & 0.752             & 0.719  & 0.716            & 0.837    & 0.793            & 0.981 & 0.801             \\
                                         & -         & -2.39\% & -            & -2.64\% & -        & -5.00\% & -       & -11.53\% & -      & -0.32\% & -        & -5.21\% & -     & -18.40\% \\ \midrule
               \multirow{2}{*}{ETTm2}    & 0.237     & 0.235            & 0.240        & 0.238            & 0.237    & 0.236            & 0.235   & 0.227             & 0.232  & 0.228            & 0.258    & 0.255            & 0.271 & 0.264             \\
                                         & -         & -0.76\% & -            & -0.79\% & -        & -0.55\% & -       & -3.36\%  & -      & -1.85\% & -        & -1.05\% & -     & -2.48\%  \\ \midrule
               \multirow{2}{*}{Exchange} & 0.356     & 0.342            & 0.385        & 0.367            & 0.371    & 0.350            & 0.408   & 0.377             & 0.295  & 0.294            & 0.422    & 0.394            & 0.542 & 0.412             \\
                                         & -         & -4.02\% & -            & -4.73\% & -        & -5.61\% & -       & -7.44\%  & -      & -0.51\% & -        & -6.59\% & -     & -23.90\% \\ \midrule
               \multirow{2}{*}{Traffic}  & 0.478     & 0.478            & 0.486        & 0.486            & 0.586    & 0.586            & 0.628   & 0.628             & 0.812  & 0.812            & 0.544    & 0.544            & 0.691 & 0.691             \\
                                         & -         & 0.00\%  & -            & 0.00\%  & -        & 0.00\%  & -       & 0.00\%   & -      & 0.00\%  & -        & 0.00\%  & -     & 0.00\%   \\ \midrule
               \multirow{2}{*}{WTH}      & 0.356     & 0.355            & 0.353        & 0.352            & 0.354    & 0.353            & 0.342   & 0.340             & 0.341  & 0.342            & 0.364    & 0.364            & 0.376 & 0.382             \\
                                         & -         & -0.31\% & -            & -0.23\% & -        & -0.23\% & -       & -0.35\%  & -      & 0.21\%  & -        & -0.25\% & -     & 1.44\%   \\  \bottomrule
            \end{tabular}
         \end{small}
      \end{threeparttable}}
\end{table*}

From this section onward, all reported MSE values are averaged over the two forecast horizons $H=30$ and $H=60$. Experiment results are presented in Table~\ref{tab:baseline}. Across the seven backbones, adding the pooling mechanism lowers the MSE on $47$ of the $56$ backbone--dataset pairs, with an average reduction of $3.5\%$. The gain is largest for the TCN backbone, which drops from $0.344$ to $0.309$ on ECL ($-10.16\%$), from $0.981$ to $0.801$ on ETTm1 ($-18.40\%$), and from $0.542$ to $0.412$ on Exchange ($-23.90\%$). This substantial improvement can be attributed to the unique characteristics of TCN, which leverages both time and variable dimensions in its design, while models like PatchTST~\cite{patchtst} rely solely on the time dimension and gain comparatively little. On the Traffic dataset, however, the MSE is identical with and without \method (e.g., $0.478$ for TimeMixer), because the stream is dominated by frequent, low-magnitude fluctuations rather than the distinct, recurring concepts it is designed to capture. Crucially, this null result validates the robustness of the detection mechanism: in a regime of micro-fluctuations where macro-properties remain stable, the approach correctly withholds unnecessary evolution instead of over-adapting. Unlike hyper-sensitive detectors that might trigger false alarms, and thus catastrophic forgetting, it preserves the stable macroscopic model, and the neural network's gradient descent already suffices, so intervention is unnecessary.

\subsection{Comparison with Online Forecasting Models}

\begin{table*}[htbp]
   \centering
   \caption{Average MSE error of previous online forecasting methods; all previous methods and \method use TCN as the backbone. Percentages in parentheses represent the relative change compared to the TCN baseline. Best and second-best performances are highlighted. Full results are presented in Appendix Table~\ref{tab:full_online}.}
   \label{tab:online}
   \setlength{\tabcolsep}{20pt}
   \renewcommand{\arraystretch}{1}
   \resizebox{\linewidth}{!}{
      \begin{threeparttable}
         \begin{small}
            \begin{tabular}{ccccccccc}
               \toprule
               Data                     & ECL               & ETTh1             & ETTh2             & ETTm1             & ETTm2            & Exchange          & Traffic           & WTH              \\ \midrule
               TCN                      & \secres{0.344}    & 0.463             & 3.166             & \secres{0.981}    & \secres{0.271}   & \secres{0.542}    & \secres{0.691}    & \fires{0.376}    \\ \midrule
               \multirow{2}{*}{ER}      & 0.624             & 0.454             & 3.329             & 1.382             & 0.322            & 2.909             & 1.110             & 0.558            \\
                                        & 81.16\%  & -1.88\%  & 5.16\%   & 40.81\%  & 19.06\% & 437.19\% & 60.60\%  & 48.29\% \\ \midrule
               \multirow{2}{*}{DER++}   & 0.580             & \fires{0.407}     & \secres{3.073}    & 1.213             & 0.304            & 2.515             & 1.076             & 0.529            \\
                                        & 68.26\%  & -12.12\% & -2.94\%  & 23.61\%  & 12.38\% & 364.47\% & 55.67\%  & 40.69\% \\ \midrule
               \multirow{2}{*}{FSNet}   & 0.843             & 0.470             & 4.072             & 1.562             & 0.347            & 3.486             & 1.466             & 0.743            \\
                                        & 144.72\% & 1.51\%   & 28.61\%  & 59.17\%  & 28.15\% & 543.77\% & 112.05\% & 97.53\% \\ \midrule
               \multirow{2}{*}{OneNet}  & 0.425             & \secres{0.410}    & 3.402             & 1.093             & 0.285            & 1.130             & \fires{0.673}     & 0.414            \\
                                        & 23.49\%  & -11.47\% & 7.47\%   & 11.39\%  & 5.10\%  & 108.72\% & -2.65\%  & 10.12\% \\ \midrule
               \multirow{2}{*}{\method} & \fires{0.309}     & 0.459             & \fires{2.804}     & \fires{0.801}     & \fires{0.264}    & \fires{0.412}     & \secres{0.691}    & \secres{0.382}   \\
                                        & -10.16\% & -0.76\%  & -11.44\% & -18.40\% & -2.48\% & -23.90\% & 0.00\%   & 1.44\%  \\ \bottomrule
            \end{tabular}
         \end{small}
      \end{threeparttable}}
\end{table*}

Experiment results are presented in Table~\ref{tab:online}. The replay and parameter-updating baselines degrade sharply under delayed feedback. On Exchange, where the recurring drift is most pronounced, ER, DER++, FSNet, and OneNet reach $2.909$, $2.515$, $3.486$, and $1.130$, respectively, whereas \method stays at $0.412$, between $2.7\times$ and $8.5\times$ lower. These failures stem from structural limitations: replay methods~\cite{er, derpp} reuse stored instances but, as memory fills, earlier concepts are displaced and cannot be recalled; FSNet~\cite{fsnet} suffers forecasting bias from the delayed arrival of samples; and OneNet~\cite{onenet} only partially compensates by adapting to short-term information. To preserve the knowledge of different concepts comprehensively, the pooling mechanism assigns each concept to a dedicated forecaster, so recurring concepts are reactivated rather than relearned. On ETTm1, \method attains $0.801$ against $1.382$ for ER and $1.562$ for FSNet. Several baselines even fall behind the plain TCN backbone, such as FSNet's $3.486$ versus $0.542$ on Exchange, confirming that in delayed recurring-drift settings these methods offer no advantage over a single model.

\subsection{Comparison with Time Series Foundation Models}\label{sec:tsfm_analysis}

\begin{table*}[!t]
    \centering
    \caption{Comparison with zero-shot TSFMs. MSE is averaged over both horizons; lower is better. Size is the mean weight size in the corresponding evaluation setting. Full per-horizon results are presented in Appendix Tables~\ref{tab:tsfm_mse} and~\ref{tab:tsfm_size}.}
    \label{tab:tsfm_summary}
    \small
    \setlength{\tabcolsep}{15pt}
    \resizebox{\linewidth}{!}{
        \begin{tabular}{lcccccccc|c}
            \toprule
            Method               & ECL            & ETTh1          & ETTh2          & ETTm1          & ETTm2          & Exchange       & Traffic        & WTH            & Size (MB)     \\
            \midrule
            TCN+\method          & 0.309          & 0.459          & 2.804          & 0.801          & 0.264          & 0.413          & 0.691          & 0.382          & 74.09         \\
            TimesNet+\method     & 0.289          & 0.341          & 2.559          & 0.793          & 0.256          & 0.395          & 0.544          & 0.364          & 19.61         \\
            SegRNN+\method       & 0.275          & \textbf{0.300} & 2.584          & \textbf{0.716} & \textbf{0.228} & \textbf{0.293} & 0.811          & \textbf{0.342} & \textbf{0.43} \\
            DLinear+\method      & 0.299          & 0.307          & 2.631          & 0.752          & \textbf{0.228} & 0.378          & 0.627          & \textbf{0.341} & \textbf{0.33} \\
            PatchTST+\method     & 0.296          & 0.327          & 2.458          & 0.721          & 0.236          & 0.351          & 0.585          & 0.354          & 5.91          \\
            iTransformer+\method & \textbf{0.256} & 0.324          & 2.542          & 0.740          & 0.237          & 0.366          & \textbf{0.486} & 0.353          & 1.11          \\
            TimeMixer+\method    & 0.270          & 0.323          & \textbf{2.508} & 0.737          & 0.235          & 0.342          & \textbf{0.479} & 0.355          & 7.39          \\
            \midrule
            Chronos 2            & 0.355          & 0.325          & 2.598          & 0.876          & 0.418          & 0.298          & 0.614          & 0.460          & 455.79        \\
            Moirai 2.0           & 0.362          & 0.315          & 2.269          & 0.841          & 0.350          & 0.323          & 0.712          & 0.400          & 43.45         \\
            TimesFM 2.5          & 0.355          & 0.308          & 2.237          & 0.816          & 0.360          & 0.319          & 0.597          & 0.373          & 882.32        \\
            \bottomrule
        \end{tabular}}
\end{table*}

Table~\ref{tab:tsfm_summary} compares \method with Chronos 2~\cite{chronos}, Moirai 2.0~\cite{moirai}, and TimesFM 2.5~\cite{timesfm} under their zero-shot usage setting. The \method variants are trained online from the observed stream, while the TSFMs use fixed pretrained parameters without target-domain fine-tuning. On several recurring-drift settings the lightweight backbones are strongest, such as $0.256$ on ECL (iTransformer) and $0.228$ on ETTm2 (SegRNN), versus $0.355$ and $0.418$ for Chronos 2, whereas the TSFMs stay competitive on ETTh2, where TimesFM 2.5 reaches $2.237$ against $2.508$ for the best variant. The decisive gap is in model size: the smallest variant requires $0.33$ MB, roughly $1400\times$ less than Chronos 2 ($455.79$ MB) and $2700\times$ less than TimesFM 2.5 ($882.32$ MB), which matters when inference and adaptation must remain local. The comparison therefore separates two deployment regimes: broad zero-shot transfer from large pretrained models and continuous specialization to an evolving local stream.

\subsection{Forecaster Assignment Mechanisms}\label{sec:assignment_analysis}

\begin{table*}[htbp]
    \centering
    \caption{Forecaster assignment under recurring concept drift. MSE is averaged over both horizons and seven backbones; lower is better. Size is the mean weight size across the same settings. Full per-horizon results are presented in Appendix Tables~\ref{tab:moe_mse} and~\ref{tab:moe_size}.}
    \label{tab:moe_summary}
    \small
    \setlength{\tabcolsep}{15pt}
    \resizebox{\linewidth}{!}{
        \begin{tabular}{lcccccccc|c}
            \toprule
            Assignment                & ECL            & ETTh1          & ETTh2          & ETTm1          & ETTm2          & Exchange       & Traffic        & WTH            & Size (MB)     \\
            \midrule
            \method Nearest Retrieval & \textbf{0.285} & \textbf{0.336} & \textbf{2.584} & \textbf{0.751} & 0.240          & \textbf{0.355} & 0.606          & 0.356          & \textbf{3.11} \\
            Softmax Gating            & 0.304          & 0.359          & 2.749          & 0.819          & \textbf{0.238} & 0.417          & \textbf{0.557} & \textbf{0.346} & 6.16          \\
            Noisy Top-$K$ Gating      & 0.301          & 0.345          & 2.718          & 0.787          & 0.241          & 0.393          & 0.569          & 0.353          & 6.16          \\
            \bottomrule
        \end{tabular}}
\end{table*}

Table~\ref{tab:moe_summary} compares nearest retrieval with Softmax and Noisy Top-$K$ gating using the same seven forecasting backbones. Nearest retrieval achieves the lowest error on five of eight datasets, such as $0.285$ on ECL and $0.336$ on ETTh1, and the smallest model state ($3.11$ MB versus $6.16$ MB for both gating variants). Softmax gating is competitive on ETTm2, Traffic, and WTH, where \method rarely evolves new forecasters. Under clearer recurring shifts, however, soft routing updates multiple experts with non-zero weights, whereas nearest retrieval updates only the selected forecaster. This parameter isolation preserves inactive experts until their associated concepts recur.

\subsection{Visualization}

\begin{figure*}[!t]
    \begin{minipage}{0.32\linewidth}
        \centering
        \includegraphics[width = \linewidth]{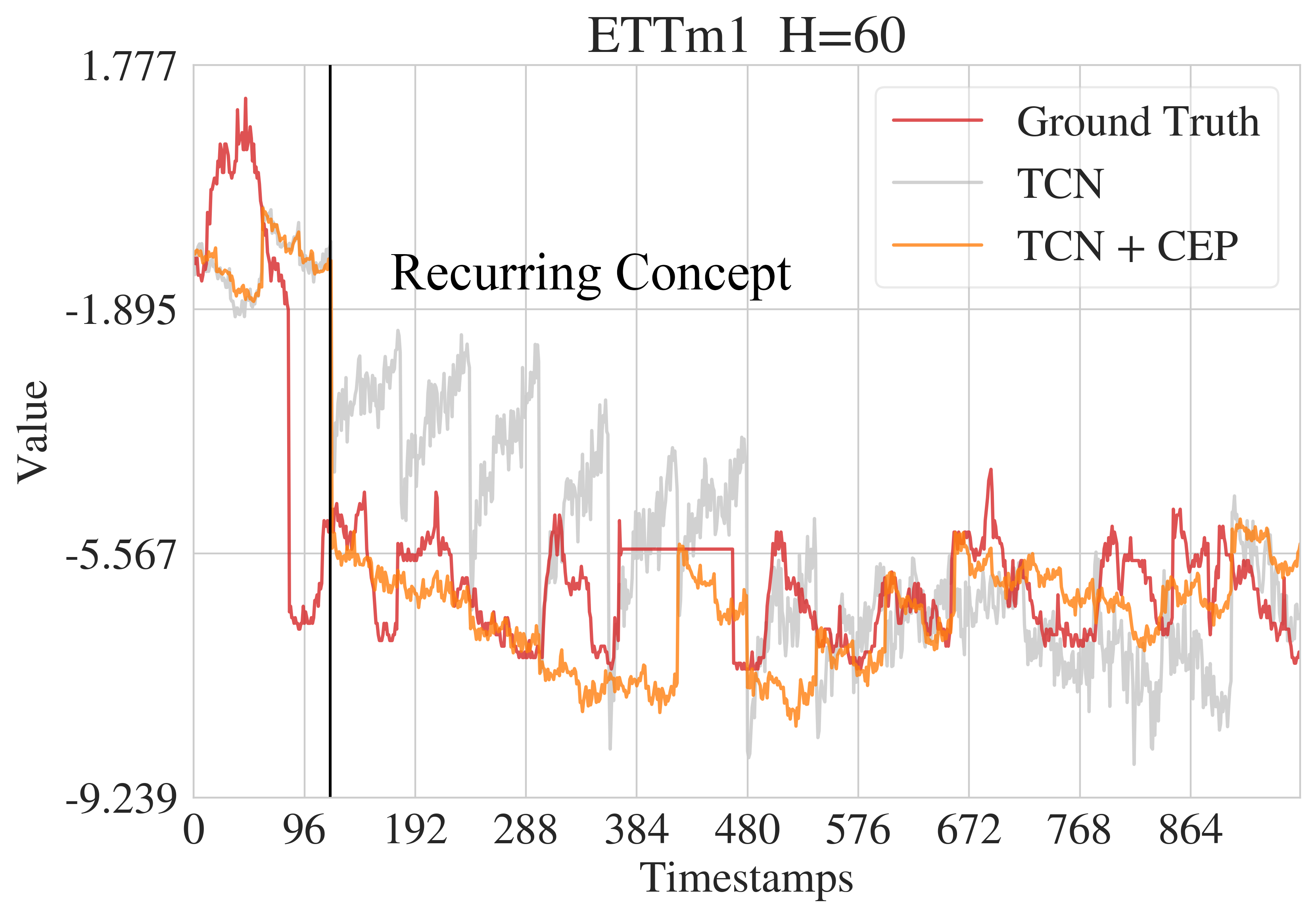}
        \caption{Forecasting values on ETTm1.}
        \label{fig:vis_value}
    \end{minipage}
    \hfill
    \begin{minipage}{0.32\linewidth}
        \centering
        \includegraphics[width = \linewidth]{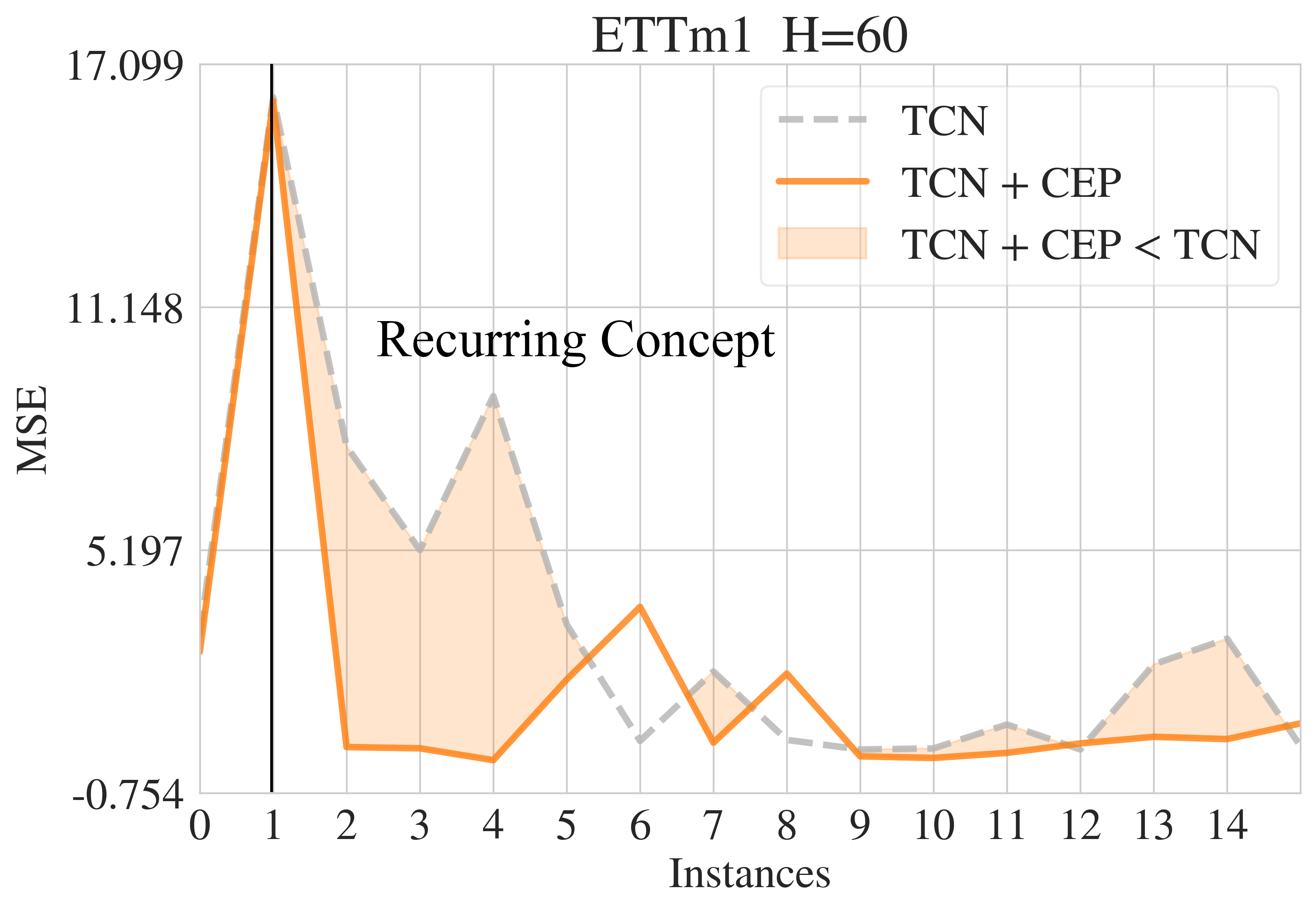}
        \caption{MSE on ETTm1.}
        \label{fig:vis_mse}
    \end{minipage}
    \hfill
    \begin{minipage}{0.32\linewidth}
        \centering
        \includegraphics[width = \linewidth]{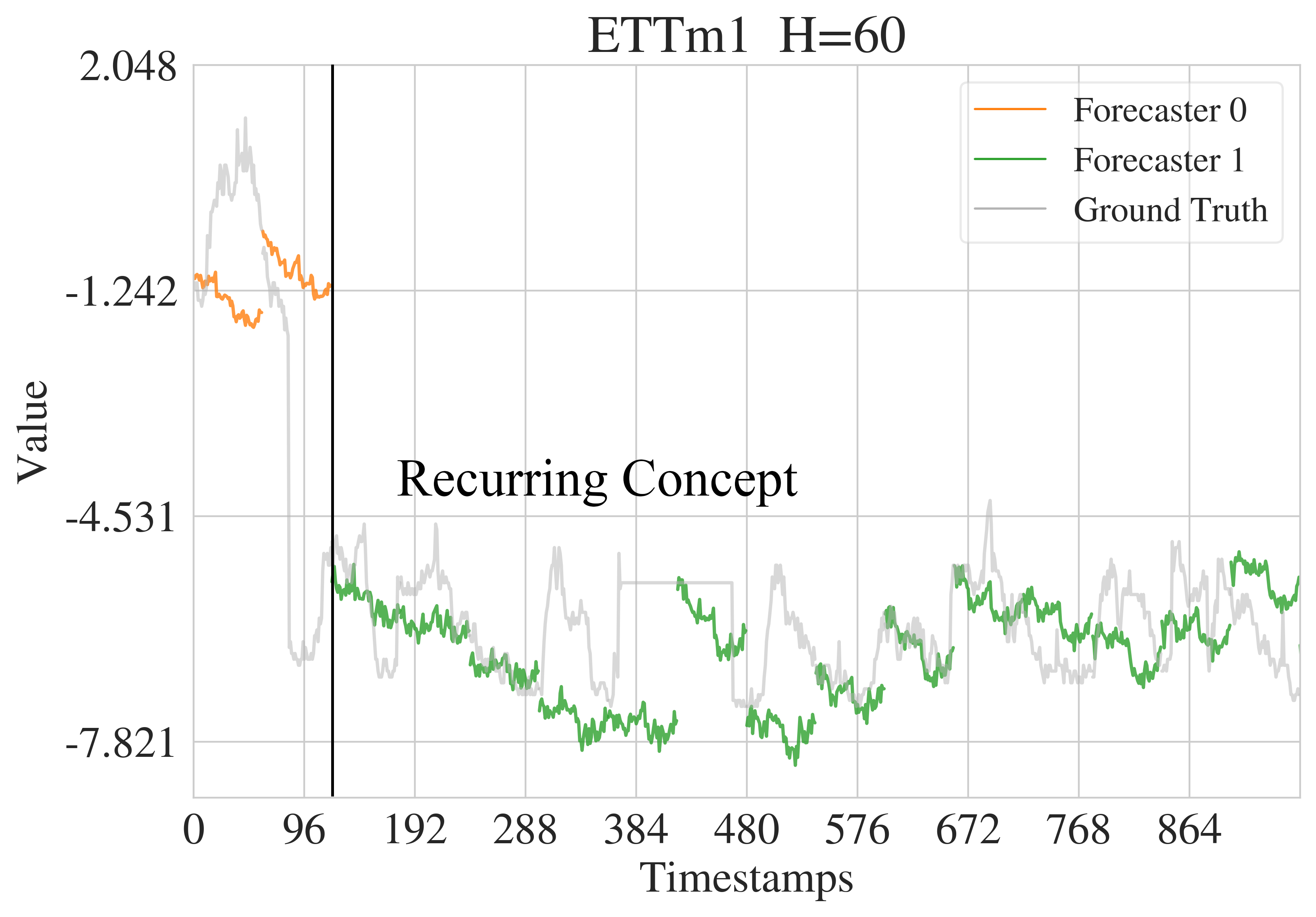}
        \caption{Forecaster assignment on ETTm1.}
        \label{fig:vis_traj_pred}
    \end{minipage}
\end{figure*}

Figures~\ref{fig:vis_value} and~\ref{fig:vis_mse} illustrate the forecasting results and MSE of both the naive model and \method. When the concept recurs during instances 2-14, the naive model exhibits a significant performance degradation, as evidenced by the surge in MSE. In contrast, \method maintains a stable and low MSE, demonstrating its effective adaptation to the recurring concept. Figure~\ref{fig:vis_traj_pred} provides insight into the internal mechanism of \method. The different colors represent distinct forecasters within the pool. The visualization confirms that \method promptly activates a specialized forecaster, Forecaster 1, that has learned the features of the recurring concept. This targeted activation is the reason for its superior performance, as it avoids catastrophic forgetting and leverages stored knowledge.

\begin{figure}[!h]
    \centering
    \subfigure[Lifecycle]{\label{fig:ettm1_lifecycle}
        \includegraphics[width = 0.95\linewidth]{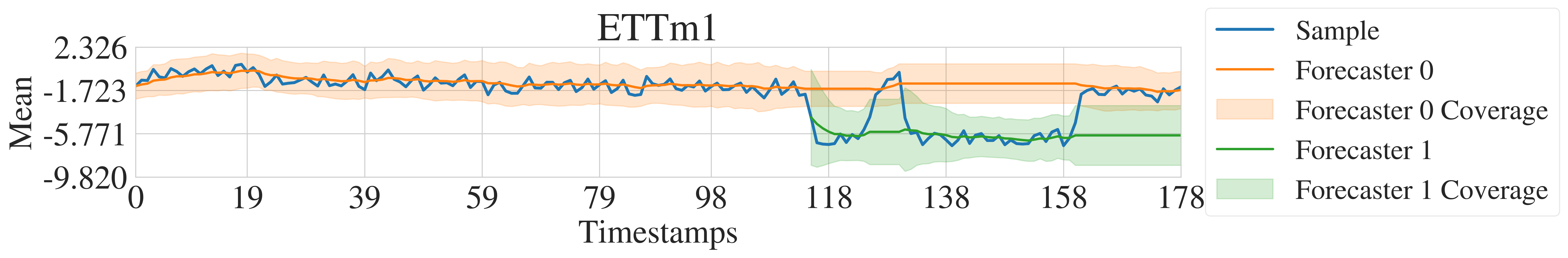}
    }
    \subfigure[Internal Prediction]{\label{fig:ettm1_internal}
        \includegraphics[width = 0.95\linewidth]{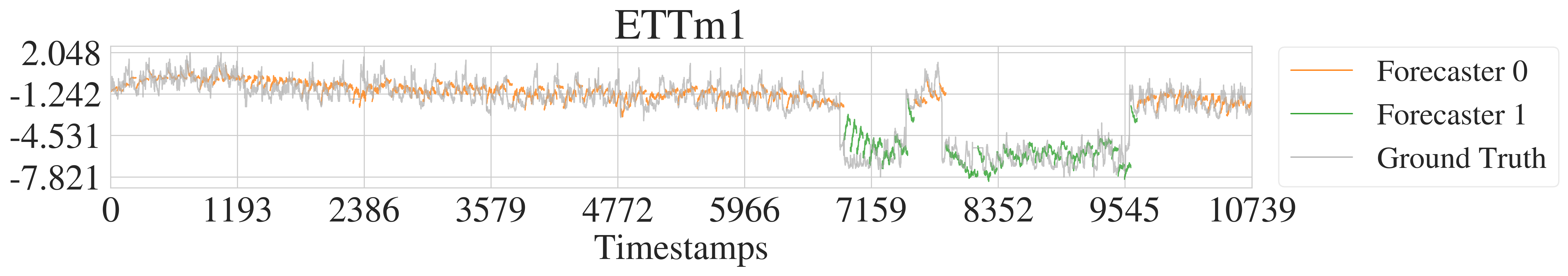}
    }
    \caption{Operational trajectory of \method on the ETTm1 dataset. (a) shows the lifecycle of forecaster genes as instances arrive, and (b) shows the forecaster each instance is assigned to over the full stream.}
    \label{fig:ettm1_traj}
\end{figure}

To make the runtime behavior of \method concrete, Figure~\ref{fig:ettm1_traj} traces the full lifecycle on the ETTm1 dataset, which is used throughout this section for consistency. In Figure~\ref{fig:ettm1_lifecycle}, the gene of the active forecaster moves as instances arrive, and a distinct forecaster is evolved only when the gene deviates enough to indicate a genuine shift rather than noise. Figure~\ref{fig:ettm1_internal} shows the resulting assignment over the entire stream: each instance is routed to a single specialized forecaster, so when a concept recurs, \method reactivates the forecaster that already learned it instead of retraining from scratch. This hand-off between forecasters, visible as the colored segments, is the direct mechanism by which \method avoids catastrophic forgetting while staying replay-free. The variation in the number of active forecasters over the full stream, across all datasets, is presented in Figure~\ref{fig:f_g_traj} in the appendix.

\subsection{Computational Complexity}

The time complexity per prediction step is $O(k)$, where $k$ represents the computational operations required by the base model for a single input, since \method activates only a single specialized forecaster from the pool for any given instance.

\begin{figure}[!h]
    \centering
    \subfigure[Time Complexity]{
        \label{fig:time_complexity}
        \begin{minipage}{0.95\linewidth}
            \centering
            \includegraphics[width = \linewidth]{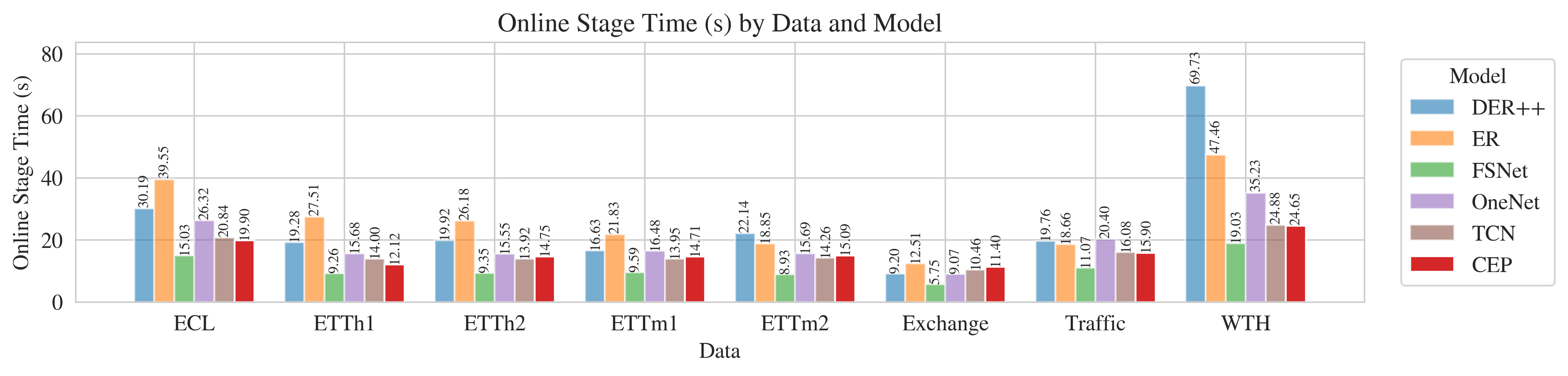}
        \end{minipage}
    }
    \subfigure[Space Complexity]{
        \label{fig:space_complexity}
        \begin{minipage}{0.95\linewidth}
            \centering
            \includegraphics[width = \linewidth]{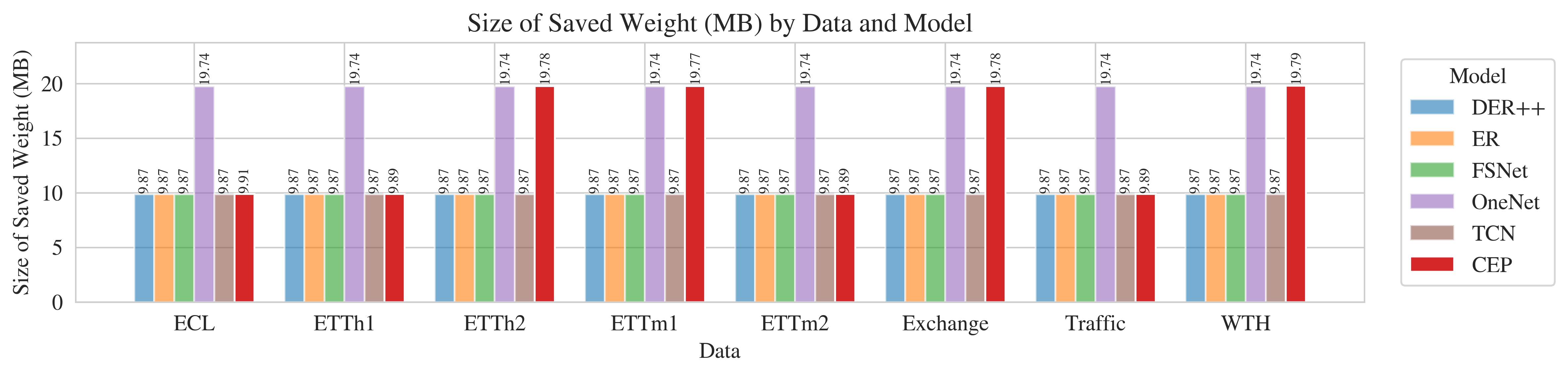}
        \end{minipage}
    }
    \caption{Computation Complexity of base TCN, online methods and \method}
    \label{fig:performance}
\end{figure}

The theoretical worst-case space complexity is $O(|\widetilde{\D}|\cdot k)$, where $|\widetilde{\D}|$ is the total number of unique concepts encountered. However, the Forecaster Elimination mechanism dynamically prunes the pool by removing forecasters corresponding to transient, noisy, or outdated concepts, preventing linear growth. Consequently, the actual memory footprint adapts to the number of active and relevant concepts in the current data regime, keeping the pool size bounded and the space complexity manageable in practice.

\subsection{Ablation Study}

\begin{table*}[!h]
    \centering
    \caption{MSE error of the ablation experiments. Results are the mean over backbone forecasters. Best and second-best performances are highlighted. Full results are presented in Appendix Table~\ref{tab:ablation}.}
    \label{tab:tcn_abl}
    \setlength{\tabcolsep}{12pt}
    \renewcommand{\arraystretch}{1}
    \resizebox{\linewidth}{!}{
        \begin{threeparttable}
            \begin{small}
                \begin{tabular}{cccccccccc|c}
                    \toprule
                    Dataset                           &  & ECL            & ETTh1          & ETTh2          & ETTm1         & ETTm2          & Exchange       & Traffic & WTH            & \fires{AVG}    \\ \midrule
                    \method                           &  & \fires{0.314}  & \secres{0.438} & \secres{2.799} & \fires{0.801} & \secres{0.256} & \secres{0.404} & 0.710   & 0.379          & \fires{0.762}  \\ \midrule
                    Base TCN (\textit{w/o} Evolution) &  & 0.347          & \fires{0.436}  & 3.106          & 0.991         & 0.261          & 0.524          & 0.710   & 0.375          & 0.844          \\ \midrule
                    \textit{w/o} Local Gene           &  & \secres{0.317} & 0.451          & 2.904          & \fires{0.801} & 0.257          & 0.524          & 0.710   & \secres{0.375} & 0.792          \\
                    \textit{w/o} Global Gene          &  & 0.321          & \secres{0.438} & 2.872          & 0.865         & \secres{0.256} & \fires{0.401}  & 0.710   & \fires{0.367}  & 0.779          \\
                    \textit{w/o} Elimination          &  & 0.335          & \secres{0.438} & \fires{2.797}  & \fires{0.801} & \fires{0.254}  & 0.406          & 0.710   & 0.379          & \secres{0.765} \\ \bottomrule
                \end{tabular}
            \end{small}
        \end{threeparttable}}
\end{table*}

To validate the efficacy of each component in \method, extensive ablation studies are conducted, with the results presented in Table~\ref{tab:tcn_abl}. The evolutionary mechanism is indispensable: disabling Evolution raises the average MSE from $0.762$ to $0.844$ ($+10.8\%$), the largest degradation among all ablations. The complete \method yields the best results, while removing the Elimination mechanism has the smallest effect, at $0.765$, within $0.4\%$ of the full model. This highlights the dual role of elimination: it not only maintains a manageable pool size but also discards outdated knowledge, ensuring continuous adaptation. Comprehensive ablation results and the full per-dataset experiment results are provided in Appendix~\ref{sec:app:results}.

\subsection{Hyperparameter Analysis}\label{sec:hyper_analysis}

\begin{figure}[h]
    \centering
    \subfigure[Mean Threshold $\tau_{\mu}$]{
        \includegraphics[width = 0.4\linewidth]{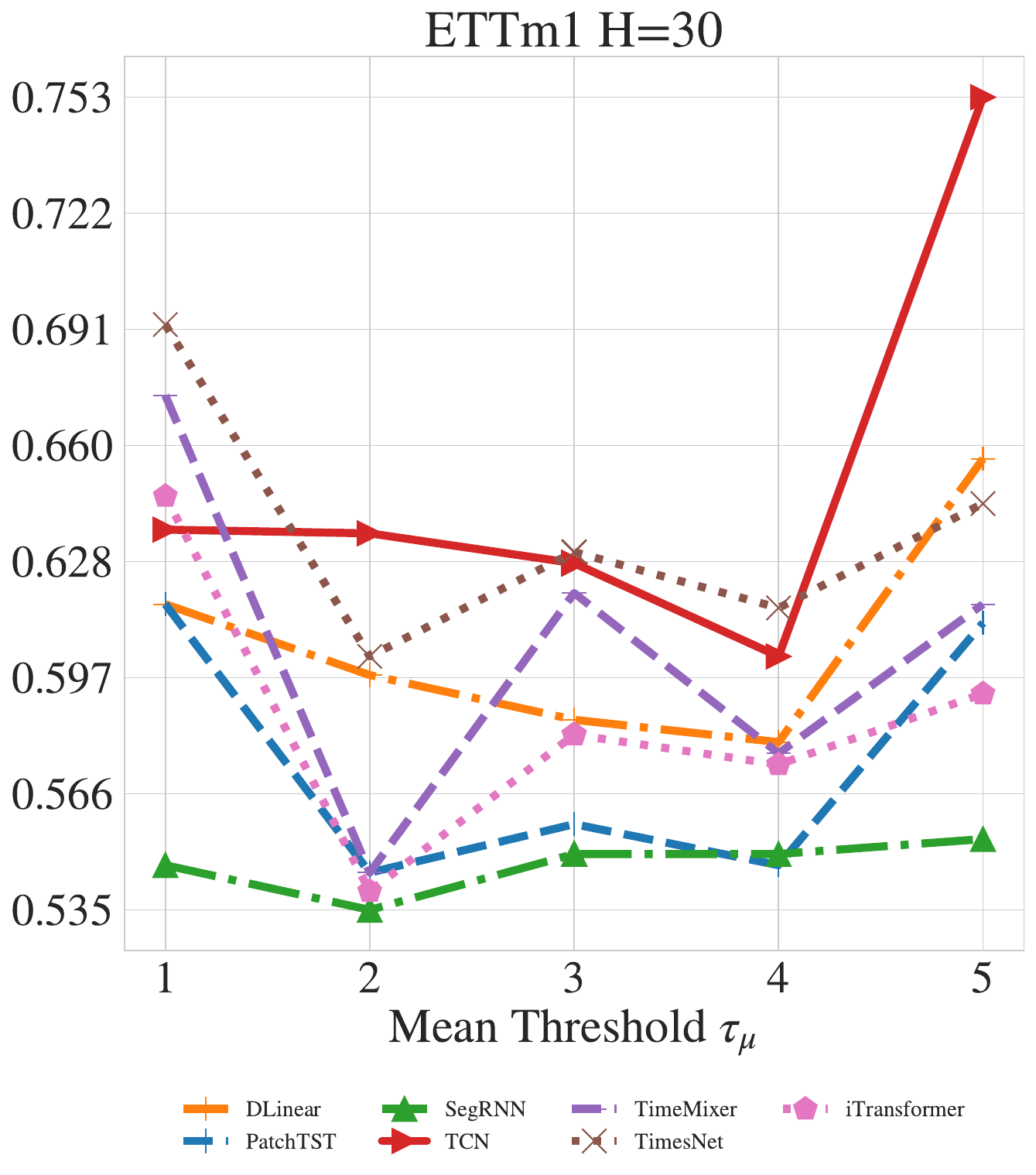}
        \label{fig:ETTm1_30_Mean_Threshold}
    }\hfill
    \subfigure[Safe time $\tau_{safe}$]{
        \includegraphics[width = 0.4\linewidth]{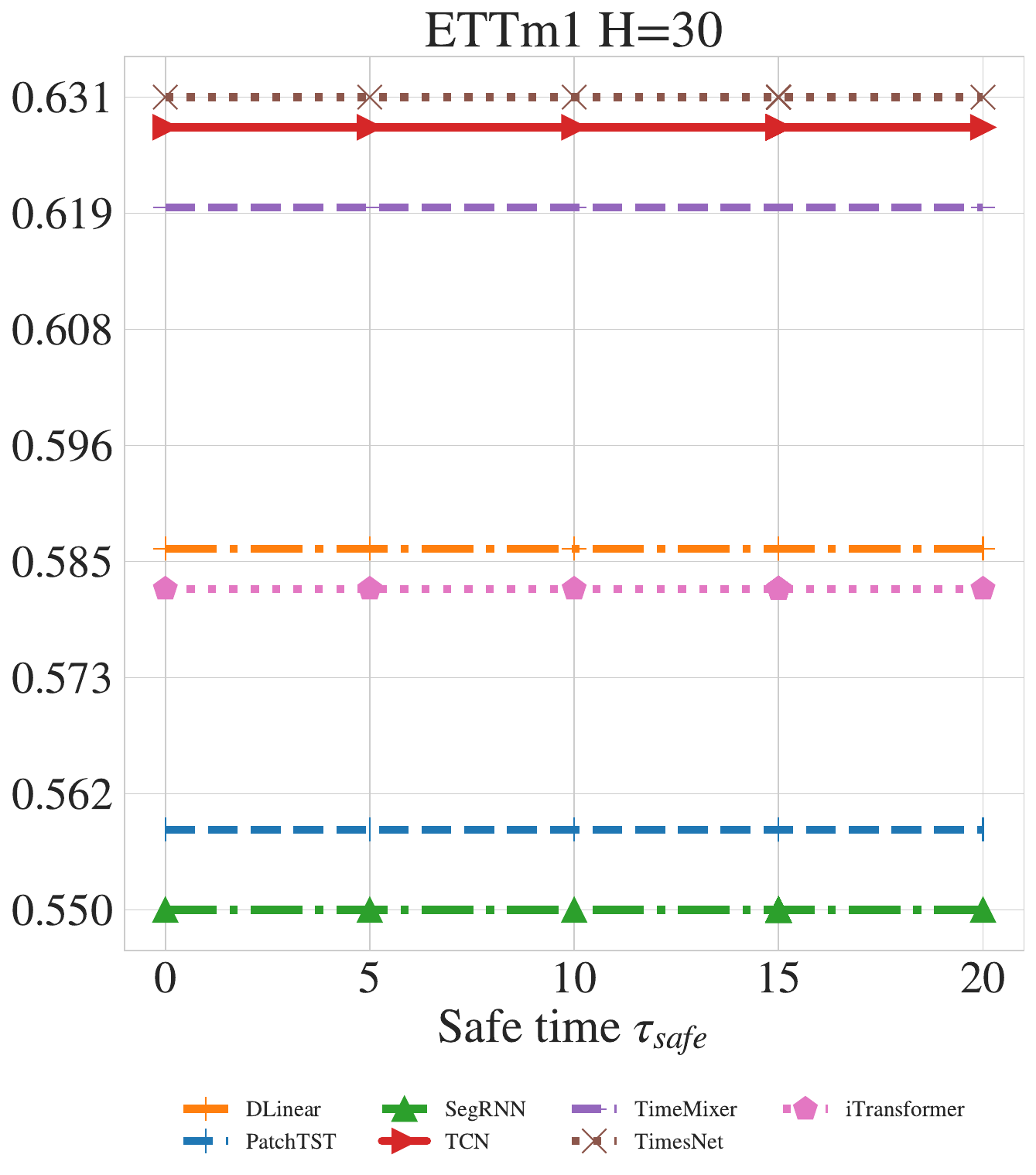}
        \label{fig:ETTm1_30_Safe_time}
    }\\
    \subfigure[Elimination Threshold $\tau_e$]{
        \includegraphics[width = 0.4\linewidth]{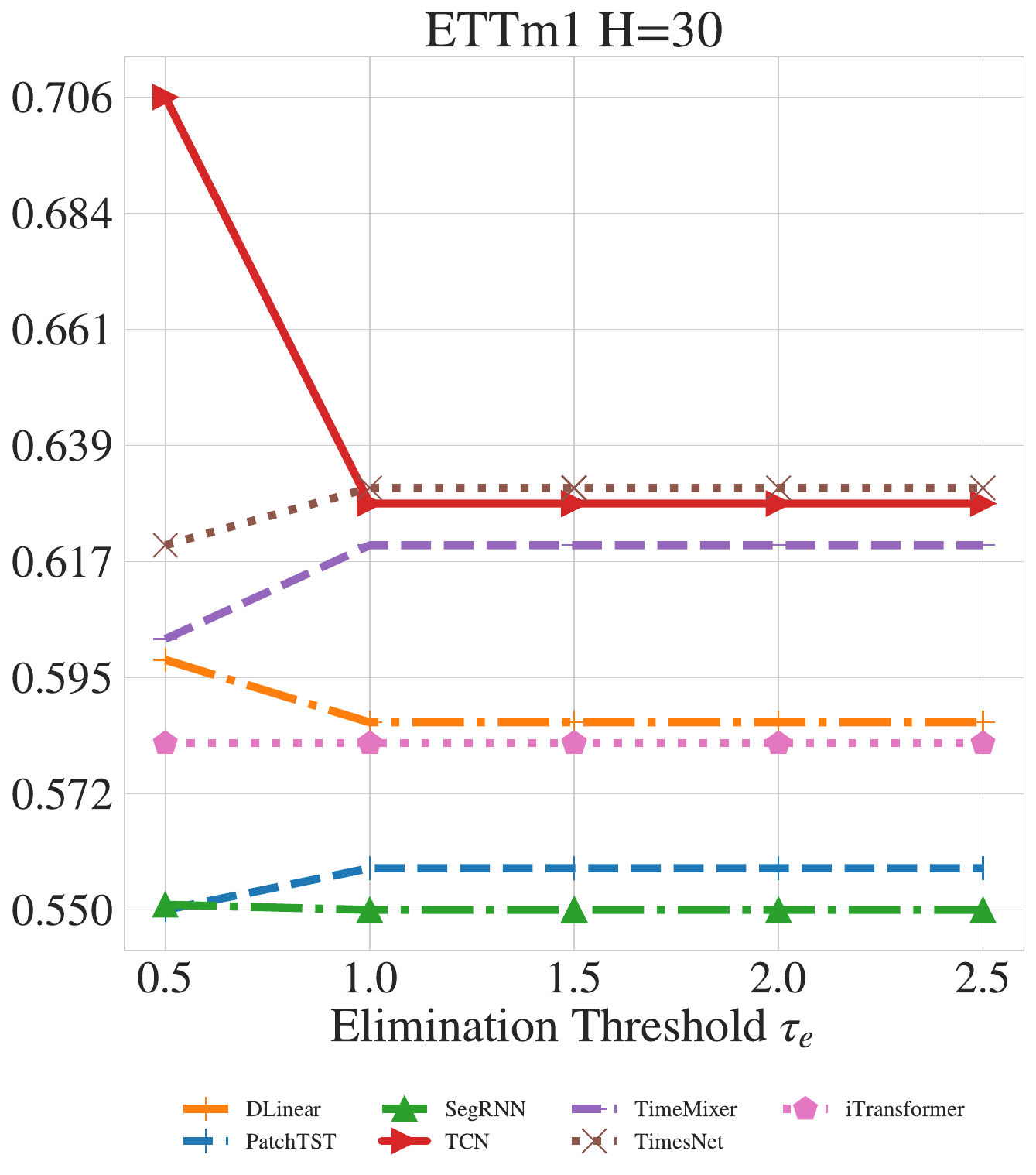}
        \label{fig:ETTm1_30_Elimination_Threshold}
    }\hfill
    \subfigure[Gene Ratio $\tau_{g}$]{
        \includegraphics[width = 0.4\linewidth]{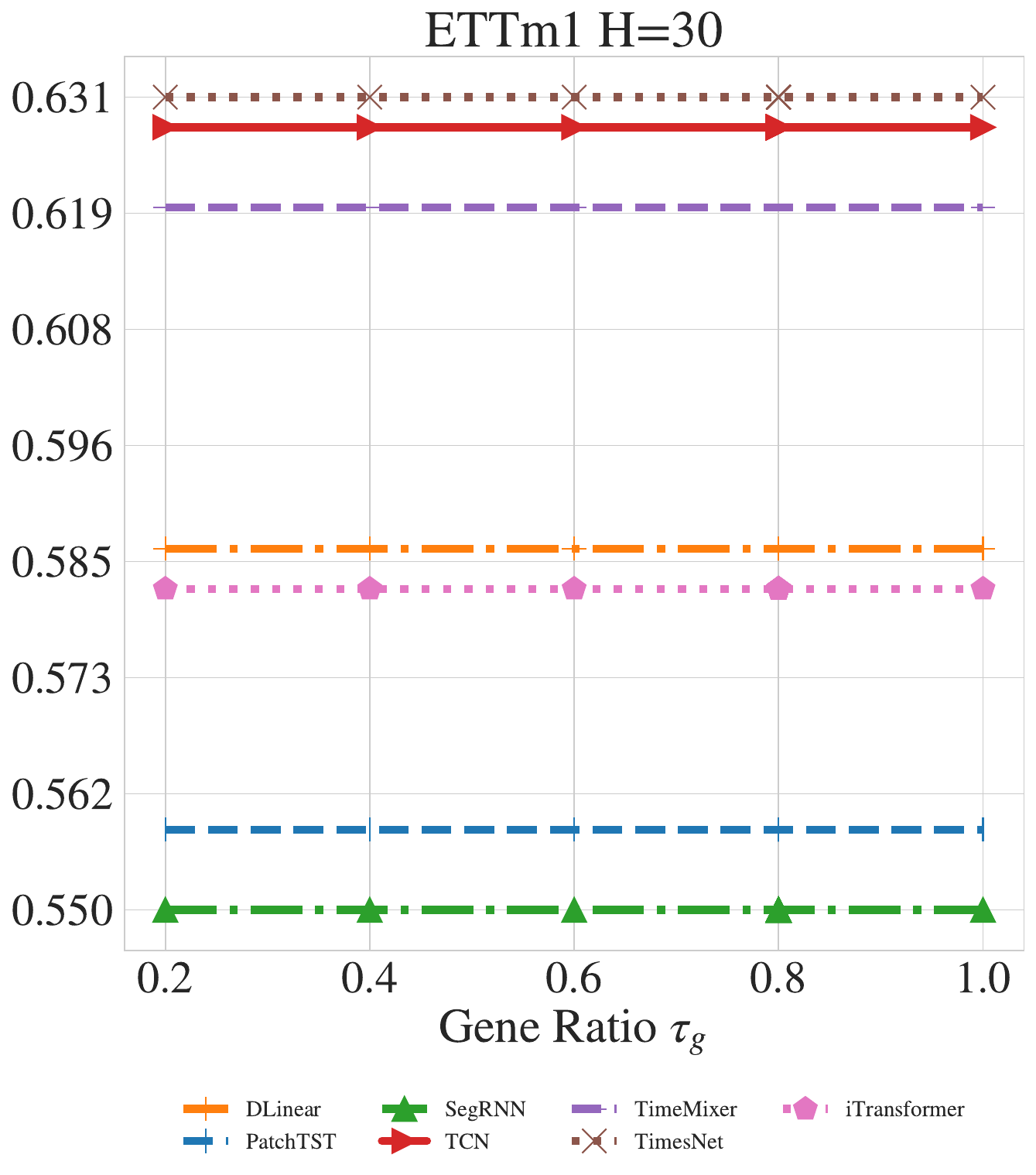}
        \label{fig:ETTm1_30_Gene_Ratio}
    }\\
    \subfigure[Adjust Ratio $\tau_{lr}$]{
        \includegraphics[width = 0.4\linewidth]{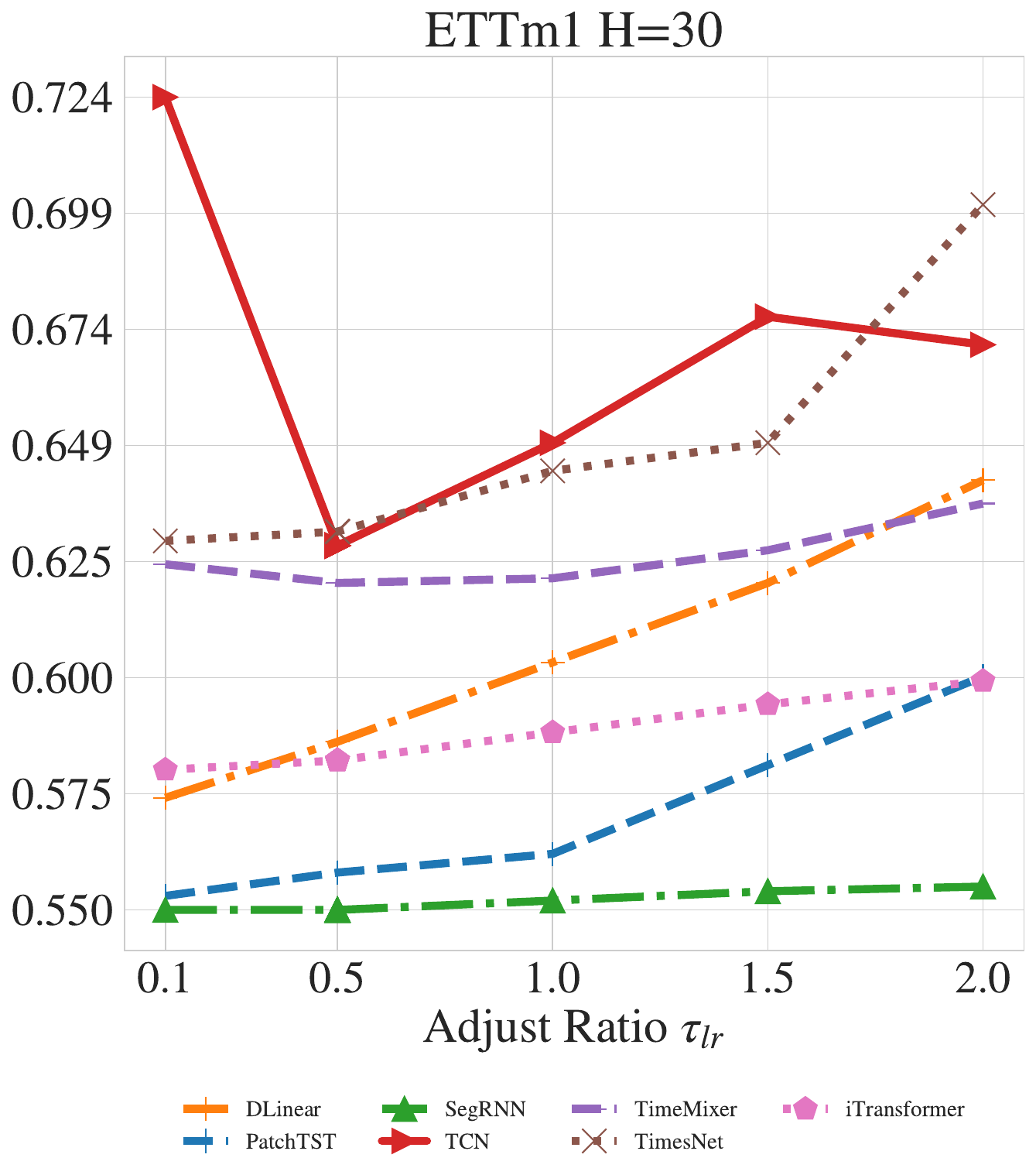}
        \label{fig:ETTm1_30_Adjust_Ratio}
    }\hfill
    \subfigure[Local Ratio $\tau_{l}$]{
        \includegraphics[width = 0.4\linewidth]{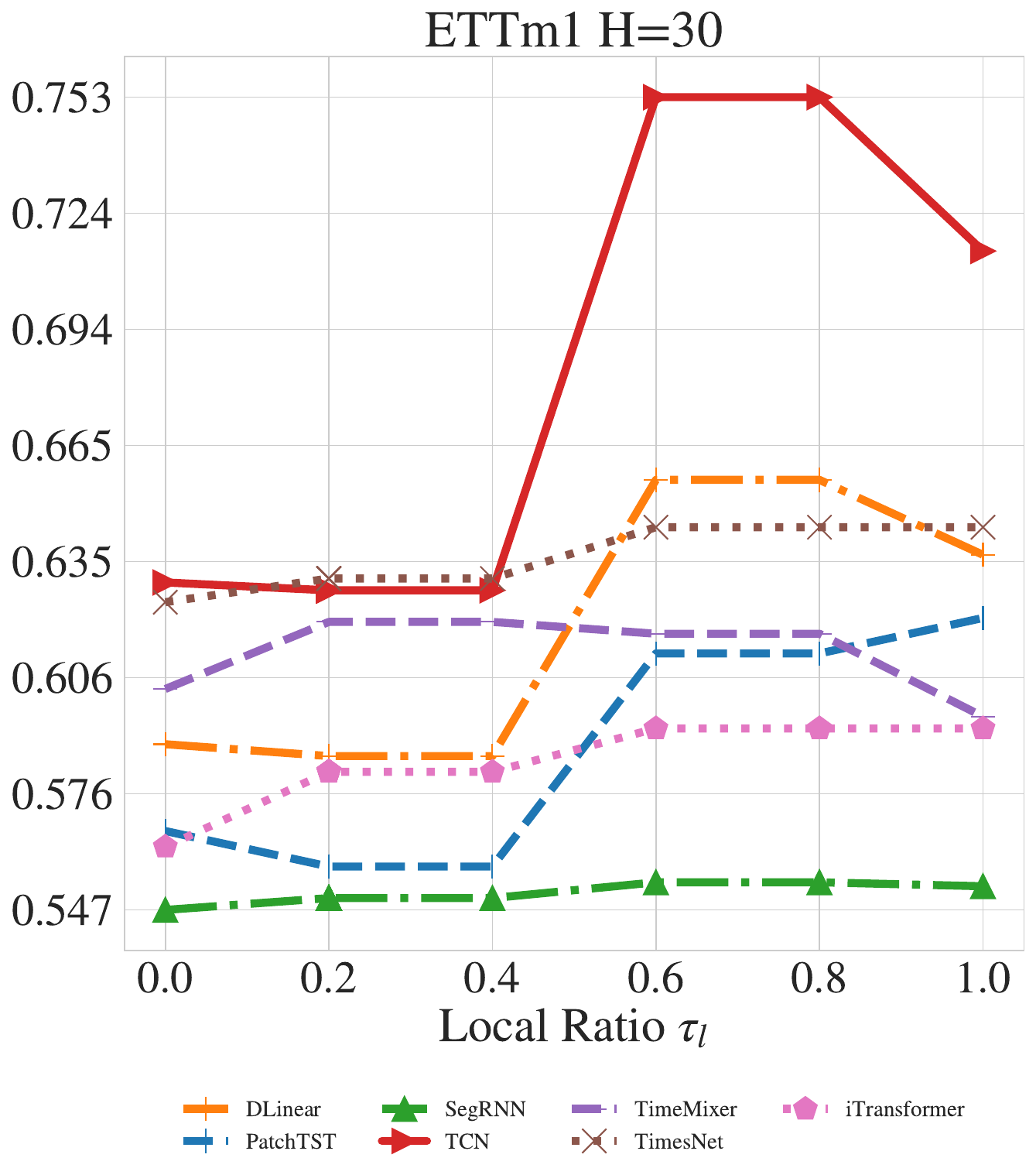}
        \label{fig:ETTm1_30_Local_Ratio}
    }
    \caption{Results of Hyperparameter Sensitivity Analysis}
    \label{fig:Sen}
\end{figure}

To examine how \method behaves as its configuration varies, each hyperparameter was swept independently on the ETTm1 dataset with horizon $\mathbf{H}=30$, holding the others at their defaults; the resulting curves are in Figure~\ref{fig:Sen}. The Safe time in Figure~\ref{fig:ETTm1_30_Safe_time} and the Gene Ratio in Figure~\ref{fig:ETTm1_30_Gene_Ratio} are both entirely flat across their ranges, and the Elimination Threshold in Figure~\ref{fig:ETTm1_30_Elimination_Threshold} is flat once above a small threshold, so none of these requires tuning. The remaining three show a clear low-error region: the Mean Threshold in Figure~\ref{fig:ETTm1_30_Mean_Threshold} follows a U-shaped curve minimized in $[2,4]$, the Adjust Ratio in Figure~\ref{fig:ETTm1_30_Adjust_Ratio} is lowest at the left end and degrades only gradually, and the Local Ratio in Figure~\ref{fig:ETTm1_30_Local_Ratio} stays flat up to about $0.4$ before degrading. Within these regions the errors are tightly clustered, so the framework tolerates moderate deviation from its operating point.

\begin{table}[!ht]
    \centering
    \caption{Practical guide for tuning \method's hyperparameters.}
    \label{tab:hyper_guide}
    \setlength{\tabcolsep}{4pt}
    \renewcommand{\arraystretch}{1.15}
    \begin{footnotesize}
        \begin{tabular}{@{}p{3.4cm} p{4.8cm}@{}}
            \toprule
            \textbf{Parameter} & \textbf{Tuning guidance}                                                                                                   \\
            \midrule
            $\tau_{\mu}$: mean threshold, how much the mean must shift to trigger evolution.
                               & Noisy or gradually drifting data: raise to 3.0--5.0. Abrupt, clear shifts: lower to 2.0--3.0.                              \\
            \midrule
            $\tau_{g}$: gene ratio, the weight of short-term local statistics in the gene.
                               & Insensitive in practice: the default 0.8 works well across datasets.                                                       \\
            \midrule
            $\tau_{e}$: elimination threshold, how long an idle forecaster is retained.
                               & Concepts recurring over long intervals (e.g., annual seasonality): raise. Memory-constrained or transient concepts: lower. \\
            \midrule
            $\tau_{safe}$: safety period, instances a new forecaster must see before evolving again.
                               & Volatile or spiky data: lengthen. Cleaner data: shorten.                                                                   \\
            \midrule
            $\tau_{lr}$: learning-rate adjustment ratio for a newly evolved forecaster.
                               & Keep small (0.1--0.5) for stable convergence; larger values risk unstable training.                                        \\
            \midrule
            $\tau_{l}$: local gene ratio, how fast the local gene adapts (EMA decay).
                               & Rapidly changing environments: raise to 0.5--0.8. Smoother trends: lower to 0.1--0.3.                                      \\
            \bottomrule
        \end{tabular}
    \end{footnotesize}
\end{table}

The defaults in Table~\ref{tab:default_hypara} therefore sit at the center of these basins rather than at their edges, away from the regimes where performance degrades, so they transfer across datasets without per-dataset retuning. Table~\ref{tab:hyper_guide} summarizes each parameter's effect and the tuning direction for adapting to a specific dataset.

\section{Conclusion}\label{sec:conclusion}

In this paper, the critical challenge of recurring concept drift in online time series forecasting is addressed, where models often suffer from catastrophic forgetting. The Continuous Evolution Pool (\method), an evolutionary framework that maintains a diverse pool of specialized forecasters to preserve learned concepts, is introduced. By evolving new forecasters from the most relevant existing ones and eliminating outdated knowledge, \method effectively adapts to concept shifts. Experiments demonstrate that \method significantly enhances long-horizon forecasting performance, overcoming the limitations of current online methods that struggle with knowledge retention. Future work will explore richer gene representations, potentially incorporating higher-order moments or spectral features to address more complex drifts, and extend this framework to handle asynchronous shifts in multivariate time series.

\bibliographystyle{IEEEtran}
\bibliography{main}

\appendices

\newpage
\section{Additional Results}\label{sec:app:results}

This appendix collects the full experimental results that are summarized in the main text: the complete per-dataset results for the base forecasters and the online methods, the full ablation study, the per-horizon results of the assignment mechanisms and the TSFM comparison, and the visualizations of the forecasting results and the forecaster gene trajectories across all datasets.

\begin{table*}[!ht]
    \centering
    \caption{Full MSE errors of different forecasters with and without \method. Percentages in parentheses show the relative change after adding \method.}
    \label{tab:full_baseline}
    \setlength{\tabcolsep}{2pt}
    \renewcommand{\arraystretch}{1}
    \resizebox{\linewidth}{!}{    
        \begin{threeparttable}    
            \begin{small}    
                \begin{tabular}{ccccccccccccccccc}
                    \toprule
                    \multicolumn{17}{c}{\textbf{MSE}}                                                                                                                                                                                                                                                                                                                                                                             \\ \midrule
                    \multirow{2}{*}{Data} & \multicolumn{2}{c}{ECL} & \multicolumn{2}{c}{ETTh1} & \multicolumn{2}{c}{ETTh2} & \multicolumn{2}{c}{ETTm1} & \multicolumn{2}{c}{ETTm2} & \multicolumn{2}{c}{Exchange} & \multicolumn{2}{c}{Traffic} & \multicolumn{2}{c}{WTH}                                                                                                                                                        \\ \cmidrule{2-17}
                                          & 30                      & 60                        & 30                        & 60                        & 30                        & 60                           & 30                          & 60                      & 30               & 60               & 30                & 60                & 30              & 60              & 30              & 60               \\ \midrule
                    TimeMixer             & 0.266                   & 0.314                     & 0.295                     & 0.376                     & 2.112                     & 3.149                        & 0.577                       & 0.933                   & 0.219            & \secres{0.255}   & 0.245             & 0.466             & \fires{0.433}   & \fires{0.524}   & 0.313           & 0.399            \\
                    +\method  & 0.244                   & 0.297                     & 0.274                     & 0.371                     & 2.000                     & \secres{3.016}               & 0.579                       & 0.895                   & 0.216            & \fires{0.254}    & 0.233             & 0.450             & \fires{0.433}   & \fires{0.524}   & 0.313           & 0.396            \\
                                          & -8.34\%        & -5.17\%          & -7.13\%          & -1.38\%          & -5.29\%          & -4.21\%             & 0.38\%             & -4.10\%        & -1.19\% & -0.39\% & -5.13\%  & -3.43\%  & 0.00\% & 0.00\% & 0.00\% & -0.55\% \\ \midrule
                    iTransformer          & 0.254                   & \secres{0.296}            & 0.280                     & 0.371                     & 1.974                     & 3.159                        & 0.606                       & 0.915                   & 0.212            & 0.267            & 0.274             & 0.495             & \secres{0.437}  & \secres{0.534}  & 0.311           & 0.395            \\
                    +\method  & \fires{0.229}           & \fires{0.282}             & 0.275                     & 0.372                     & 1.968                     & 3.116                        & 0.594                       & 0.886                   & 0.210            & 0.265            & 0.249             & 0.484             & \secres{0.437}  & \secres{0.534}  & 0.311           & 0.394            \\
                                          & -9.62\%        & -4.72\%          & -2.00\%          & 0.11\%           & -0.27\%          & -1.34\%             & -1.98\%            & -3.08\%        & -1.22\% & -0.45\% & -9.33\%  & -2.18\%  & 0.00\% & 0.00\% & 0.00\% & -0.40\% \\ \midrule
                    PatchTST              & 0.279                   & 0.340                     & 0.284                     & 0.380                     & 2.028                     & 3.160                        & 0.607                       & 0.911                   & 0.209            & 0.265            & 0.263             & 0.479             & 0.582           & 0.589           & 0.310           & 0.398            \\
                    +\method  & 0.265                   & 0.328                     & 0.276                     & 0.377                     & 1.922                     & \fires{2.995}                & 0.567                       & \fires{0.875}           & 0.208            & 0.264            & 0.234             & 0.467             & 0.582           & 0.589           & 0.310           & 0.397            \\
                                          & -5.01\%        & -3.70\%          & -2.81\%          & -0.95\%          & -5.26\%          & -5.22\%             & -6.55\%            & -3.97\%        & -0.67\% & -0.45\% & -11.25\% & -2.51\%  & 0.00\% & 0.00\% & 0.00\% & -0.40\% \\ \midrule
                    DLinear               & 0.275                   & 0.340                     & 0.289                     & 0.385                     & 2.070                     & 3.366                        & 0.656                       & 1.042                   & 0.200            & 0.271            & 0.292             & 0.523             & 0.614           & 0.641           & \fires{0.295}   & 0.388            \\
                    +\method  & 0.264                   & 0.335                     & 0.259                     & 0.356                     & 2.052                     & 3.209                        & 0.586                       & 0.917                   & \fires{0.196}    & 0.259            & 0.261             & 0.494             & 0.614           & 0.641           & \fires{0.295}   & 0.386            \\
                                          & -4.07\%        & -1.41\%          & -10.38\%         & -7.43\%          & -0.88\%          & -4.65\%             & -10.69\%           & -12.05\%       & -2.00\% & -4.36\% & -10.69\% & -5.62\%  & 0.00\% & 0.00\% & 0.00\% & -0.62\% \\ \midrule
                    SegRNN                & 0.242                   & 0.314                     & \secres{0.250}            & \secres{0.356}            & \secres{1.864}            & 3.391                        & \secres{0.554}              & 0.883                   & 0.201            & 0.263            & \secres{0.193}    & \secres{0.397}    & 0.693           & 0.930           & \secres{0.298}  & \fires{0.384}    \\
                    +\method  & \secres{0.239}          & 0.312                     & \fires{0.248}             & \fires{0.353}             & \fires{1.840}             & 3.328                        & \fires{0.551}               & \secres{0.881}          & \secres{0.197}   & 0.259            & \fires{0.192}     & \fires{0.395}     & 0.693           & 0.930           & \secres{0.298}  & \secres{0.385}   \\
                                          & -1.08\%        & -0.57\%          & -0.80\%          & -0.73\%          & -1.26\%          & -1.88\%             & -0.58\%            & -0.16\%        & -1.89\% & -1.82\% & -0.52\%  & -0.50\%  & 0.00\% & 0.00\% & 0.00\% & 0.36\%  \\ \midrule
                    TimesNet              & 0.281                   & 0.330                     & 0.304                     & 0.398                     & 2.141                     & 3.212                        & 0.692                       & 0.981                   & 0.237            & 0.279            & 0.297             & 0.546             & 0.492           & 0.597           & 0.323           & 0.406            \\
                    +\method  & 0.259                   & 0.319                     & 0.291                     & 0.390                     & 2.054                     & 3.065                        & 0.639                       & 0.947                   & 0.234            & 0.277            & 0.259             & 0.530             & 0.492           & 0.597           & 0.323           & 0.404            \\
                                          & -7.77\%        & -3.27\%          & -4.08\%          & -2.11\%          & -4.04\%          & -4.59\%             & -7.60\%            & -3.53\%        & -1.35\% & -0.79\% & -13.05\% & -3.07\%  & 0.00\% & 0.00\% & 0.00\% & -0.49\% \\ \midrule
                    TCN                   & 0.295                   & 0.393                     & 0.371                     & 0.555                     & 2.493                     & 3.838                        & 0.761                       & 1.201                   & 0.221            & 0.320            & 0.389             & 0.694             & 0.668           & 0.714           & 0.314           & 0.439            \\
                    +\method  & 0.261                   & 0.358                     & 0.392                     & 0.526                     & 2.164                     & 3.443                        & 0.638                       & 0.964                   & 0.219            & 0.309            & 0.272             & 0.553             & 0.668           & 0.714           & 0.314           & 0.450            \\
                                          & -11.71\%       & -9.00\%          & 5.88\%           & -5.19\%          & -13.20\%         & -10.29\%            & -16.21\%           & -19.79\%       & -0.81\% & -3.62\% & -30.25\% & -20.33\% & 0.00\% & 0.00\% & 0.00\% & 2.46\%  \\ \midrule
                    \multicolumn{17}{c}{\textbf{Std}}                                                                                                                                                                                                                                                                                                                                                                             \\ \midrule
                    \multirow{2}{*}{Data} & \multicolumn{2}{c}{ECL} & \multicolumn{2}{c}{ETTh1} & \multicolumn{2}{c}{ETTh2} & \multicolumn{2}{c}{ETTm1} & \multicolumn{2}{c}{ETTm2} & \multicolumn{2}{c}{Exchange} & \multicolumn{2}{c}{Traffic} & \multicolumn{2}{c}{WTH}                                                                                                                                                        \\ \cmidrule{2-17}
                                          & 30                      & 60                        & 30                        & 60                        & 30                        & 60                           & 30                          & 60                      & 30               & 60               & 30                & 60                & 30              & 60              & 30              & 60               \\ \midrule
                    TimeMixer             & 0.006                   & 0.003                     & 0.016                     & 0.006                     & 0.060                     & 0.032                        & 0.027                       & 0.019                   & 0.020            & 0.018            & 0.010             & 0.013             & 0.018           & 0.033           & 0.002           & 0.003            \\
                    +\method  & 0.007                   & 0.003                     & 0.005                     & 0.005                     & 0.013                     & 0.006                        & 0.025                       & 0.010                   & 0.022            & 0.018            & 0.011             & 0.014             & 0.018           & 0.033           & 0.002           & 0.003            \\ \midrule
                    iTransformer          & 0.004                   & 0.003                     & 0.002                     & 0.004                     & 0.014                     & 0.027                        & 0.010                       & 0.012                   & 0.002            & 0.003            & 0.010             & 0.008             & 0.012           & 0.009           & 0.002           & 0.001            \\
                    +\method  & 0.004                   & 0.001                     & 0.002                     & 0.004                     & 0.009                     & 0.036                        & 0.012                       & 0.020                   & 0.002            & 0.003            & 0.003             & 0.009             & 0.012           & 0.009           & 0.002           & 0.002            \\ \midrule
                    PatchTST              & 0.004                   & 0.002                     & 0.001                     & 0.005                     & 0.018                     & 0.035                        & 0.014                       & 0.012                   & 0.005            & 0.007            & 0.002             & 0.011             & 0.006           & 0.004           & 0.001           & 0.001            \\
                    +\method  & 0.002                   & 0.004                     & 0.004                     & 0.004                     & 0.015                     & 0.019                        & 0.007                       & 0.008                   & 0.005            & 0.007            & 0.006             & 0.012             & 0.006           & 0.004           & 0.001           & 0.001            \\ \midrule
                    DLinear               & 0.000                   & 0.001                     & 0.001                     & 0.002                     & 0.002                     & 0.002                        & 0.002                       & 0.001                   & 0.001            & 0.006            & 0.012             & 0.013             & 0.001           & 0.000           & 0.000           & 0.000            \\
                    +\method  & 0.001                   & 0.000                     & 0.000                     & 0.001                     & 0.003                     & 0.003                        & 0.002                       & 0.001                   & 0.001            & 0.006            & 0.010             & 0.013             & 0.001           & 0.000           & 0.000           & 0.000            \\ \midrule
                    SegRNN                & 0.003                   & 0.003                     & 0.001                     & 0.002                     & 0.006                     & 0.011                        & 0.001                       & 0.004                   & 0.003            & 0.002            & 0.001             & 0.000             & 0.013           & 0.026           & 0.002           & 0.001            \\
                    +\method  & 0.003                   & 0.003                     & 0.001                     & 0.002                     & 0.004                     & 0.006                        & 0.002                       & 0.003                   & 0.003            & 0.002            & 0.001             & 0.000             & 0.013           & 0.026           & 0.002           & 0.002            \\ \midrule
                    TimesNet              & 0.012                   & 0.008                     & 0.010                     & 0.020                     & 0.020                     & 0.042                        & 0.062                       & 0.045                   & 0.004            & 0.009            & 0.023             & 0.039             & 0.028           & 0.036           & 0.005           & 0.003            \\
                    +\method  & 0.014                   & 0.011                     & 0.009                     & 0.014                     & 0.059                     & 0.066                        & 0.022                       & 0.048                   & 0.003            & 0.008            & 0.019             & 0.052             & 0.028           & 0.036           & 0.006           & 0.003            \\ \midrule
                    TCN                   & 0.009                   & 0.007                     & 0.013                     & 0.058                     & 0.053                     & 0.107                        & 0.006                       & 0.052                   & 0.021            & 0.012            & 0.019             & 0.016             & 0.070           & 0.006           & 0.002           & 0.003            \\
                    +\method  & 0.010                   & 0.005                     & 0.015                     & 0.044                     & 0.021                     & 0.083                        & 0.006                       & 0.031                   & 0.019            & 0.011            & 0.009             & 0.018             & 0.070           & 0.006           & 0.002           & 0.004            \\ \bottomrule
                \end{tabular}
            \end{small}
        \end{threeparttable}}
\end{table*}

\begin{table*}[htbp]
    \centering
    \caption{Full MSE error of previous online forecasting methods. Best and second-best performances are highlighted.}
    \label{tab:full_online}
    \setlength{\tabcolsep}{10pt}
    \renewcommand{\arraystretch}{1.2}
    \resizebox{\linewidth}{!}{    
        \begin{threeparttable}    
            \begin{small}    
                \begin{tabular}{ccccccccccccccccc}
                    \toprule
                    \multicolumn{17}{c}{\textbf{MSE}}                                                                                                                                                                                                                                                                                                                                                                     \\ \midrule
                    \multirow{2}{*}{Data}        & \multicolumn{2}{c}{ECL} & \multicolumn{2}{c}{ETTh1} & \multicolumn{2}{c}{ETTh2} & \multicolumn{2}{c}{ETTm1} & \multicolumn{2}{c}{ETTm2} & \multicolumn{2}{c}{Exchange} & \multicolumn{2}{c}{Traffic} & \multicolumn{2}{c}{WTH}                                                                                                                                         \\ \cmidrule{2-17}
                                                 & 30                      & 60                        & 30                        & 60                        & 30                        & 60                           & 30                          & 60                      & 30             & 60             & 30             & 60             & 30             & 60             & 30             & 60             \\ \midrule
                    TCN                 & \secres{0.295}          & \secres{0.393}            & 0.371                     & 0.555                     & 2.493                     & 3.838                        & \secres{0.761}              & \secres{1.201}          & \secres{0.221} & \secres{0.320} & \secres{0.389} & \secres{0.694} & \fires{0.668}  & \secres{0.714} & \fires{0.314}  & \fires{0.439}  \\
                    ER                           & 0.563                   & 0.685                     & 0.358                     & 0.551                     & 2.624                     & 4.035                        & 0.901                       & 1.862                   & 0.246          & 0.399          & 2.268          & 3.550          & 1.359          & 0.862          & 0.524          & 0.592          \\
                    DER++                        & 0.538                   & 0.621                     & \fires{0.332}             & \secres{0.481}            & \secres{2.339}            & \secres{3.806}               & 0.804                       & 1.622                   & 0.235          & 0.373          & 1.568          & 3.462          & 1.325          & 0.827          & 0.486          & 0.573          \\
                    FSNet                        & 0.863                   & 0.822                     & 0.392                     & 0.547                     & 3.118                     & 5.025                        & 1.111                       & 2.013                   & 0.275          & 0.419          & 3.093          & 3.879          & 1.618          & 1.314          & 0.641          & 0.846          \\
                    OneNet                       & 0.385                   & 0.466                     & \secres{0.344}            & \fires{0.475}             & 2.540                     & 4.265                        & 0.949                       & 1.237                   & 0.237          & 0.332          & 0.899          & 1.361          & \secres{0.705} & \fires{0.640}  & \secres{0.360} & 0.469          \\ \midrule
                    \method & \fires{0.261}           & \fires{0.358}             & 0.392                     & 0.526                     & \fires{2.164}             & \fires{3.443}                & \fires{0.638}               & \fires{0.964}           & \fires{0.219}  & \fires{0.309}  & \fires{0.272}  & \fires{0.553}  & \fires{0.668}  & \secres{0.714} & \fires{0.314}  & \secres{0.450} \\ \midrule
                    \multicolumn{17}{c}{\textbf{Std}}                                                                                                                                                                                                                                                                                                                                                                     \\ \midrule
                    \multirow{2}{*}{Data}        & \multicolumn{2}{c}{ECL} & \multicolumn{2}{c}{ETTh1} & \multicolumn{2}{c}{ETTh2} & \multicolumn{2}{c}{ETTm1} & \multicolumn{2}{c}{ETTm2} & \multicolumn{2}{c}{Exchange} & \multicolumn{2}{c}{Traffic} & \multicolumn{2}{c}{WTH}                                                                                                                                         \\ \cmidrule{2-17}
                                                 & 30                      & 60                        & 30                        & 60                        & 30                        & 60                           & 30                          & 60                      & 30             & 60             & 30             & 60             & 30             & 60             & 30             & 60             \\ \midrule
                    TCN                          & 0.010                   & 0.005                     & 0.015                     & 0.044                     & 0.021                     & 0.083                        & 0.006                       & 0.031                   & 0.019          & 0.011          & 0.009          & 0.018          & 0.070          & 0.006          & 0.002          & 0.004          \\
                    ER                           & 0.006                   & 0.050                     & 0.002                     & 0.026                     & 0.116                     & 0.040                        & 0.016                       & 0.058                   & 0.006          & 0.005          & 0.243          & 0.684          & 0.283          & 0.048          & 0.014          & 0.012          \\
                    DER++                        & 0.015                   & 0.055                     & 0.007                     & 0.034                     & 0.126                     & 0.070                        & 0.013                       & 0.068                   & 0.007          & 0.007          & 0.847          & 0.364          & 0.278          & 0.048          & 0.014          & 0.012          \\
                    FSNet                        & 0.108                   & 0.044                     & 0.014                     & 0.027                     & 0.042                     & 0.088                        & 0.008                       & 0.060                   & 0.019          & 0.012          & 0.335          & 0.508          & 0.093          & 0.147          & 0.055          & 0.042          \\
                    OneNet                       & 0.011                   & 0.036                     & 0.013                     & 0.016                     & 0.068                     & 0.254                        & 0.034                       & 0.099                   & 0.013          & 0.027          & 0.179          & 0.151          & 0.029          & 0.018          & 0.011          & 0.014          \\
                    \method                          & 0.009                   & 0.007                     & 0.013                     & 0.058                     & 0.053                     & 0.107                        & 0.006                       & 0.052                   & 0.021          & 0.012          & 0.019          & 0.016          & 0.070          & 0.006          & 0.002          & 0.003          \\ \bottomrule
                \end{tabular}
            \end{small}
        \end{threeparttable}}
\end{table*}

\begin{table*}[htbp]
    \centering
    \caption{Full ablation study results. MSE is averaged over backbone forecasters. Best and second-best performances are highlighted.}
    \label{tab:ablation}
    \setlength{\tabcolsep}{5pt}
    \renewcommand{\arraystretch}{1.2}
    \resizebox{\linewidth}{!}{    
        \begin{threeparttable}    
            \begin{small}    
                \begin{tabular}{cccccccccccccccccccccc|cc}
                    \toprule
                    \multirow{2}{*}{Evolution}                       & \multirow{2}{*}{Inspiration} & \multirow{2}{*}{\begin{tabular}[c]{@{}c@{}}Gradient\\ Abandonment\end{tabular}} & \multirow{2}{*}{Elimination} & \multirow{2}{*}{\begin{tabular}[c]{@{}c@{}}Local\\ Gene\end{tabular}} & \multirow{2}{*}{\begin{tabular}[c]{@{}c@{}}Global\\ Gene\end{tabular}} & \multicolumn{2}{c}{ECL} & \multicolumn{2}{c}{ETTh1} & \multicolumn{2}{c}{ETTh2} & \multicolumn{2}{c}{ETTm1} & \multicolumn{2}{c}{ETTm2} & \multicolumn{2}{c}{Exchange} & \multicolumn{2}{c}{Traffic} & \multicolumn{2}{c}{WTH} & \multicolumn{1}{c}{\multirow{2}{*}{\textbf{AVG}}} & \multirow{2}{*}{\textbf{Rank}}                                                                                                                                               \\ \cmidrule{7-22}
                                                                     &                              &                                                                                 &                              &                                                                       &                                                                        & 30                      & 60                        & 30                        & 60                        & 30                        & 60                           & 30                          & 60                      & 30                                                & 60                             & 30             & 60             & 30                 & 60    & 30             & 60             & \multicolumn{1}{c}{} &                     \\ \midrule
                    \multicolumn{6}{c}{TCN}                          & 0.295                        & 0.398                                                                           & 0.359                        & 0.512                                                                 & 2.480                                                                  & 3.732                   & 0.753                     & 1.228                     & 0.208                     & 0.313                     & 0.361                        & 0.687                       & 0.696                   & 0.724                                             & 0.313                          & 0.437          & 0.844          & 25                                                                                                        \\ \midrule
                    \Checkmark                                       & \XSolidBrush                 & \XSolidBrush                                                                    & \Checkmark                   & \XSolidBrush                                                          & \Checkmark                                                             & 0.275                   & 0.377                     & 0.332                     & 0.477                     & 2.339                     & 3.845                        & \secres{0.627}              & \fires{0.929}           & 0.219                                             & 0.309                          & 0.361          & 0.687          & 0.696              & 0.724 & 0.313          & 0.437          & 0.809                & 24                  \\
                    \Checkmark                                       & \XSolidBrush                 & \XSolidBrush                                                                    & \XSolidBrush                 & \XSolidBrush                                                          & \Checkmark                                                             & 0.294                   & 0.392                     & 0.332                     & \secres{0.446}            & 2.339                     & 3.814                        & \secres{0.627}              & \fires{0.929}           & 0.219                                             & 0.309                          & 0.361          & 0.687          & 0.696              & 0.724 & 0.313          & 0.437          & 0.807                & 23                  \\
                    \Checkmark                                       & \XSolidBrush                 & \XSolidBrush                                                                    & \Checkmark                   & \Checkmark                                                            & \XSolidBrush                                                           & 0.271                   & 0.373                     & \fires{0.330}             & \fires{0.443}             & 2.368                     & 3.958                        & \secres{0.627}              & 0.988                   & 0.207                                             & 0.309                          & 0.288          & 0.579          & 0.696              & 0.724 & 0.312          & 0.415          & 0.806                & 22                  \\
                    \Checkmark                                       & \XSolidBrush                 & \Checkmark                                                                      & \Checkmark                   & \XSolidBrush                                                          & \Checkmark                                                             & 0.275                   & 0.379                     & 0.348                     & 0.480                     & 2.353                     & 3.688                        & 0.650                       & \secres{0.935}          & 0.210                                             & \secres{0.305}                 & 0.361          & 0.687          & 0.696              & 0.724 & 0.313          & 0.437          & 0.803                & 21                  \\
                    \Checkmark                                       & \XSolidBrush                 & \XSolidBrush                                                                    & \XSolidBrush                 & \Checkmark                                                            & \XSolidBrush                                                           & 0.281                   & 0.383                     & \fires{0.330}             & \fires{0.443}             & 2.323                     & 3.886                        & \secres{0.627}              & 0.988                   & \fires{0.204}                                     & 0.309                          & 0.291          & 0.567          & 0.696              & 0.724 & 0.307          & \fires{0.410}  & 0.798                & 20                  \\
                    \Checkmark                                       & \XSolidBrush                 & \Checkmark                                                                      & \XSolidBrush                 & \XSolidBrush                                                          & \Checkmark                                                             & 0.296                   & 0.408                     & 0.348                     & 0.462                     & 2.353                     & 3.565                        & 0.650                       & \secres{0.935}          & 0.217                                             & \secres{0.305}                 & 0.361          & 0.687          & 0.696              & 0.724 & 0.313          & 0.437          & 0.797                & 19                  \\
                    \Checkmark                                       & \XSolidBrush                 & \XSolidBrush                                                                    & \XSolidBrush                 & \Checkmark                                                            & \Checkmark                                                             & 0.289                   & 0.378                     & \fires{0.330}             & \fires{0.443}             & 2.393                     & 3.815                        & \secres{0.627}              & \fires{0.929}           & 0.213                                             & 0.309                          & 0.298          & \secres{0.525} & 0.696              & 0.724 & 0.313          & 0.434          & 0.795                & 18                  \\
                    \Checkmark                                       & \XSolidBrush                 & \XSolidBrush                                                                    & \Checkmark                   & \Checkmark                                                            & \Checkmark                                                             & 0.274                   & 0.377                     & \fires{0.330}             & \fires{0.443}             & 2.366                     & 3.832                        & \secres{0.627}              & \fires{0.929}           & 0.207                                             & 0.309                          & 0.292          & \secres{0.525} & 0.696              & 0.724 & 0.313          & 0.434          & 0.792                & 17                  \\
                    \Checkmark                                       & \Checkmark                   & \Checkmark                                                                      & \Checkmark                   & \XSolidBrush                                                          & \Checkmark                                                             & 0.270                   & 0.363                     & 0.397                     & 0.505                     & 2.204                     & 3.604                        & 0.628                       & 0.973                   & 0.212                                             & \fires{0.302}                  & 0.361          & 0.687          & 0.696              & 0.724 & 0.313          & 0.437          & 0.792                & 16                  \\
                    \Checkmark                                       & \Checkmark                   & \XSolidBrush                                                                    & \Checkmark                   & \Checkmark                                                            & \XSolidBrush                                                           & 0.268                   & 0.370                     & 0.379                     & 0.490                     & 2.314                     & 3.641                        & \fires{0.596}               & 1.059                   & 0.208                                             & 0.307                          & 0.267          & 0.564          & 0.696              & 0.724 & 0.310          & 0.422          & 0.788                & 15                  \\
                    \Checkmark                                       & \Checkmark                   & \XSolidBrush                                                                    & \Checkmark                   & \XSolidBrush                                                          & \Checkmark                                                             & 0.271                   & \fires{0.358}             & 0.373                     & 0.493                     & \fires{2.160}             & 3.651                        & \fires{0.596}               & 0.936                   & 0.215                                             & 0.307                          & 0.361          & 0.687          & 0.696              & 0.724 & 0.313          & 0.437          & 0.786                & 14                  \\
                    \Checkmark                                       & \XSolidBrush                 & \Checkmark                                                                      & \XSolidBrush                 & \Checkmark                                                            & \XSolidBrush                                                           & 0.281                   & 0.392                     & \secres{0.331}            & 0.447                     & 2.311                     & 3.665                        & 0.650                       & 1.026                   & \secres{0.205}                                    & \secres{0.305}                 & 0.291          & 0.534          & 0.696              & 0.724 & 0.307          & \secres{0.411} & 0.786                & 13                  \\
                    \Checkmark                                       & \Checkmark                   & \XSolidBrush                                                                    & \XSolidBrush                 & \XSolidBrush                                                          & \Checkmark                                                             & 0.297                   & 0.394                     & 0.373                     & 0.489                     & \fires{2.160}             & 3.550                        & \fires{0.596}               & 0.936                   & 0.208                                             & 0.307                          & 0.361          & 0.687          & 0.696              & 0.724 & 0.313          & 0.437          & 0.783                & 12                  \\
                    \Checkmark                                       & \XSolidBrush                 & \Checkmark                                                                      & \Checkmark                   & \Checkmark                                                            & \XSolidBrush                                                           & 0.269                   & 0.378                     & \secres{0.331}            & 0.447                     & 2.353                     & 3.577                        & 0.650                       & 1.026                   & 0.207                                             & \secres{0.305}                 & 0.287          & 0.545          & 0.696              & 0.724 & 0.313          & 0.416          & 0.783                & 11                  \\
                    \Checkmark                                       & \Checkmark                   & \Checkmark                                                                      & \XSolidBrush                 & \XSolidBrush                                                          & \Checkmark                                                             & 0.298                   & 0.412                     & 0.397                     & 0.506                     & 2.204                     & \fires{3.368}                & 0.628                       & 0.973                   & \secres{0.205}                                    & \fires{0.302}                  & 0.361          & 0.687          & 0.696              & 0.724 & 0.313          & 0.437          & 0.782                & 10                  \\
                    \Checkmark                                       & \XSolidBrush                 & \Checkmark                                                                      & \XSolidBrush                 & \Checkmark                                                            & \Checkmark                                                             & 0.286                   & 0.387                     & \secres{0.331}            & 0.447                     & 2.334                     & 3.651                        & 0.650                       & \secres{0.935}          & 0.207                                             & \secres{0.305}                 & 0.288          & \fires{0.518}  & 0.696              & 0.724 & 0.313          & 0.438          & 0.782                & 9                   \\
                    \Checkmark                                       & \Checkmark                   & \Checkmark                                                                      & \Checkmark                   & \Checkmark                                                            & \XSolidBrush                                                           & \fires{0.264}           & 0.378                     & 0.380                     & 0.495                     & 2.313                     & 3.431                        & 0.628                       & 1.102                   & 0.209                                             & \fires{0.302}                  & 0.269          & 0.533          & 0.696              & 0.724 & 0.310          & 0.424          & 0.779                & 8                   \\
                    \Checkmark                                       & \Checkmark                   & \XSolidBrush                                                                    & \XSolidBrush                 & \Checkmark                                                            & \XSolidBrush                                                           & 0.278                   & 0.371                     & 0.379                     & 0.490                     & 2.255                     & 3.526                        & \fires{0.596}               & 1.059                   & \secres{0.205}                                    & 0.307                          & 0.272          & 0.561          & 0.696              & 0.724 & \fires{0.302}  & 0.414          & 0.777                & 7                   \\
                    \Checkmark                                       & \Checkmark                   & \Checkmark                                                                      & \XSolidBrush                 & \Checkmark                                                            & \XSolidBrush                                                           & 0.277                   & 0.383                     & 0.380                     & 0.495                     & 2.255                     & 3.425                        & 0.628                       & 1.102                   & 0.206                                             & \fires{0.302}                  & 0.274          & 0.526          & 0.696              & 0.724 & \secres{0.303} & 0.414          & 0.774                & 6                   \\
                    \Checkmark                                       & \XSolidBrush                 & \Checkmark                                                                      & \Checkmark                   & \Checkmark                                                            & \Checkmark                                                             & 0.271                   & 0.380                     & \secres{0.331}            & 0.447                     & 2.318                     & 3.570                        & 0.650                       & \secres{0.935}          & 0.207                                             & \secres{0.305}                 & 0.283          & \fires{0.518}  & 0.696              & 0.724 & 0.313          & 0.438          & 0.774                & 5                   \\
                    \Checkmark                                       & \Checkmark                   & \XSolidBrush                                                                    & \Checkmark                   & \Checkmark                                                            & \Checkmark                                                             & 0.273                   & \secres{0.359}            & 0.379                     & 0.490                     & \secres{2.165}            & 3.592                        & \fires{0.596}               & 0.936                   & 0.208                                             & 0.307                          & \secres{0.266} & 0.542          & 0.696              & 0.724 & 0.313          & 0.440          & 0.768                & 4                   \\
                    \Checkmark                                       & \Checkmark                   & \XSolidBrush                                                                    & \XSolidBrush                 & \Checkmark                                                            & \Checkmark                                                             & 0.286                   & 0.377                     & 0.379                     & 0.490                     & 2.191                     & 3.531                        & \fires{0.596}               & 0.936                   & 0.208                                             & 0.307                          & 0.269          & 0.542          & 0.696              & 0.724 & 0.313          & 0.440          & 0.768                & 3                   \\
                    \Checkmark                                       & \Checkmark                   & \Checkmark                                                                      & \XSolidBrush                 & \Checkmark                                                            & \Checkmark                                                             & 0.281                   & 0.389                     & 0.380                     & 0.495                     & 2.187                     & \secres{3.406}               & 0.628                       & 0.973                   & 0.206                                             & \fires{0.302}                  & \secres{0.266} & 0.546          & 0.696              & 0.724 & 0.313          & 0.445          & \secres{0.765}       & \secres{2} \\ \midrule
                    \multicolumn{6}{c}{\method} & \secres{0.265}               & 0.363                                                                           & 0.380                        & 0.495                                                                 & 2.174                                                                  & 3.424                   & 0.628                     & 0.973                     & 0.209                     & \fires{0.302}             & \fires{0.262}                & 0.546                       & 0.696                   & 0.724                                             & 0.313                          & 0.445          & \fires{0.762}  & \fires{1}                                                                                        \\ \bottomrule
                \end{tabular}
            \end{small}
        \end{threeparttable}}
\end{table*}

\begin{table*}[!h]
    \centering
    \caption{Full MSE error of comparison of the assignment mechanisms. The \fires{best} performances are highlighted.}
    \label{tab:moe_mse}
    \setlength{\tabcolsep}{2pt}
    \renewcommand{\arraystretch}{1.2}
    \resizebox{\linewidth}{!}{    
        \begin{threeparttable}    
            \begin{small}    
                \begin{tabular}{c|c|cccccccccccccccc|c}
                    \toprule
                    \multicolumn{2}{c}{\multirow{2}{*}{Data}} & \multicolumn{2}{c}{ECL} & \multicolumn{2}{c}{ETTh1} & \multicolumn{2}{c}{ETTh2} & \multicolumn{2}{c}{ETTm1} & \multicolumn{2}{c}{ETTm2} & \multicolumn{2}{c}{Exchange} & \multicolumn{2}{c}{Traffic} & \multicolumn{2}{c}{WTH} & \multirow{2}{*}{Count}                                                                                                                                              \\ \cmidrule{3-18}
                    \multicolumn{2}{c}{}                      & 30                      & 60                        & 30                        & 60                        & 30                        & 60                           & 30                          & 60                      & 30                     & 60            & 30            & 60            & 30            & 60            & 30            & 60            &                            \\ \midrule
                    \multirow{3}{*}{TCN}                      & Gene        & \fires{0.265}             & \fires{0.363}             & \fires{0.380}             & 0.495                     & \fires{2.174}                & \fires{3.424}               & \fires{0.628}           & \fires{0.973}          & 0.209         & \fires{0.302} & \fires{0.262} & \fires{0.546} & 0.696         & 0.724         & 0.313         & 0.445         & \fires{10} \\
                                                              & Softmax                 & 0.276                     & 0.424                     & 0.404                     & 0.550                     & 2.684                        & 4.278                       & 0.919                   & 1.327                  & 0.207         & 0.325         & 0.408         & 0.700         & 0.612         & \fires{0.636} & \fires{0.307} & \fires{0.417} & 3          \\
                                                              & Noisy Top-K             & 0.285                     & 0.393                     & 0.397                     & \fires{0.460}             & 2.541                        & 3.745                       & 0.804                   & 1.228                  & \fires{0.206} & 0.337         & 0.329         & 0.654         & \fires{0.583} & 0.681         & 0.316         & \fires{0.417} & 4          \\ \midrule
                    \multirow{3}{*}{TimesNet}                 & Gene        & \fires{0.259}             & 0.311                     & \fires{0.279}             & 0.387                     & \fires{2.009}                & 3.140                       & \fires{0.631}           & 0.938                  & 0.236         & \fires{0.269} & \fires{0.256} & \fires{0.460} & \fires{0.524} & 0.619         & 0.325         & 0.409         & \fires{8}  \\
                                                              & Softmax                 & 0.279                     & \fires{0.300}             & 0.288                     & 0.396                     & 2.261                        & \fires{3.066}               & 0.760                   & 0.952                  & 0.238         & 0.284         & 0.353         & 0.585         & 0.543         & 0.576         & \fires{0.314} & \fires{0.397} & 4          \\
                                                              & Noisy Top-K             & 0.276                     & 0.322                     & 0.295                     & \fires{0.371}             & 2.016                        & 3.224                       & 0.641                   & \fires{0.924}          & \fires{0.223} & 0.285         & 0.308         & 0.520         & 0.531         & \fires{0.542} & 0.329         & 0.412         & 4          \\ \midrule
                    \multirow{3}{*}{SegRNN}                   & Gene        & \fires{0.238}             & 0.310                     & \fires{0.248}             & \fires{0.351}             & \fires{1.843}                & \fires{3.321}               & 0.550                   & 0.880                  & 0.196         & 0.260         & \fires{0.192} & \fires{0.395} & 0.670         & 0.899         & 0.297         & 0.387         & 7          \\
                                                              & Softmax                 & 0.240                     & \fires{0.303}             & 0.254                     & \fires{0.351}             & 1.927                        & 3.401                       & \fires{0.523}           & \fires{0.785}          & \fires{0.186} & \fires{0.227} & 0.194         & 0.396         & \fires{0.630} & \fires{0.804} & \fires{0.287} & \fires{0.376} & \fires{10} \\
                                                              & Noisy Top-K             & 0.243                     & 0.311                     & 0.255                     & 0.372                     & 1.894                        & 3.454                       & 0.528                   & 0.813                  & 0.194         & 0.253         & 0.194         & 0.394         & 0.653         & 0.813         & 0.292         & 0.382         & 0          \\ \midrule
                    \multirow{3}{*}{DLinear}                  & Gene        & \fires{0.264}             & \fires{0.334}             & \fires{0.259}             & \fires{0.356}             & \fires{2.051}                & \fires{3.208}               & \fires{0.586}           & \fires{0.916}          & \fires{0.198} & 0.269         & \fires{0.272} & \fires{0.489} & 0.614         & 0.641         & 0.295         & 0.386         & \fires{11} \\
                                                              & Softmax                 & 0.281                     & 0.344                     & 0.292                     & 0.415                     & 2.072                        & 3.396                       & 0.740                   & 0.961                  & 0.203         & \fires{0.261} & 0.301         & 0.589         & 0.522         & \fires{0.543} & \fires{0.291} & \fires{0.374} & 4          \\
                                                              & Noisy Top-K             & 0.278                     & 0.362                     & 0.285                     & 0.380                     & 2.315                        & 3.427                       & 0.707                   & 0.960                  & 0.203         & 0.276         & 0.295         & 0.542         & \fires{0.519} & 0.554         & 0.295         & 0.379         & 1          \\ \midrule
                    \multirow{3}{*}{PatchTST}                 & Gene        & \fires{0.264}             & 0.324                     & \fires{0.270}             & \fires{0.379}             & \fires{1.915}                & \fires{3.017}               & \fires{0.558}           & \fires{0.881}          & 0.216         & 0.274         & \fires{0.234} & \fires{0.457} & 0.587         & 0.588         & 0.311         & 0.396         & \fires{9}  \\
                                                              & Softmax                 & 0.288                     & 0.345                     & 0.311                     & 0.380                     & 2.074                        & 3.113                       & 0.627                   & 0.884                  & \fires{0.197} & \fires{0.263} & 0.256         & 0.528         & \fires{0.569} & \fires{0.564} & \fires{0.303} & \fires{0.389} & 6          \\
                                                              & Noisy Top-K             & 0.289                     & \fires{0.322}             & 0.291                     & 0.386                     & 2.096                        & 3.052                       & 0.588                   & 0.899                  & 0.209         & 0.265         & 0.269         & 0.472         & 0.583         & 0.612         & 0.307         & 0.399         & 1          \\ \midrule
                    \multirow{3}{*}{iTransformer}             & Gene        & \fires{0.231}             & \fires{0.283}             & 0.274                     & 0.379                     & 1.962                        & \fires{3.083}               & \fires{0.582}           & 0.881                  & 0.206         & \fires{0.265} & \fires{0.250} & \fires{0.474} & 0.449         & 0.524         & 0.314         & 0.395         & \fires{7}  \\
                                                              & Softmax                 & 0.252                     & 0.303                     & 0.275                     & 0.375                     & \fires{1.941}                & 3.057                       & 0.608                   & 0.883                  & 0.202         & 0.272         & 0.280         & 0.492         & 0.428         & \fires{0.469} & \fires{0.301} & 0.386         & 3          \\
                                                              & Noisy Top-K             & 0.245                     & 0.304                     & \fires{0.272}             & \fires{0.373}             & 1.980                        & 3.236                       & 0.595                   & \fires{0.844}          & \fires{0.199} & 0.279         & 0.264         & 0.520         & \fires{0.421} & 0.470         & 0.306         & \fires{0.394} & 6          \\ \midrule
                    \multirow{3}{*}{TimeMixer}                & Gene        & \fires{0.242}             & \fires{0.299}             & \fires{0.276}             & \fires{0.371}             & 2.007                        & \fires{3.021}               & 0.620                   & 0.889                  & 0.223         & 0.241         & 0.251         & \fires{0.431} & 0.456         & 0.495         & 0.315         & 0.393         & 6          \\
                                                              & Softmax                 & 0.270                     & 0.351                     & 0.317                     & 0.413                     & 2.168                        & 3.054                       & 0.613                   & \fires{0.883}          & 0.226         & \fires{0.240} & \fires{0.240} & 0.518         & \fires{0.410} & \fires{0.490} & \fires{0.309} & \fires{0.391} & \fires{7}  \\
                                                              & Noisy Top-K             & 0.269                     & 0.309                     & 0.283                     & 0.406                     & \fires{1.984}                & 3.095                       & \fires{0.587}           & 0.907                  & \fires{0.199} & 0.245         & 0.256         & 0.488         & 0.439         & 0.571         & 0.318         & 0.401         & 3          \\ \bottomrule
                \end{tabular}
            \end{small}
        \end{threeparttable}}
\end{table*}

\begin{table*}[!h]
    \centering
    \caption{Full weight sizes (MB) of assignment mechanisms. The \fires{smallest} sizes are highlighted.}
    \label{tab:moe_size}
    \setlength{\tabcolsep}{2pt}
    \renewcommand{\arraystretch}{1.2}
    \resizebox{\linewidth}{!}{    
        \begin{threeparttable}    
            \begin{small}    
                \begin{tabular}{c|c|cccccccccccccccc|c}
                    \toprule
                    \multicolumn{2}{c}{\multirow{2}{*}{Data}} & \multicolumn{2}{c}{ECL} & \multicolumn{2}{c}{ETTh1} & \multicolumn{2}{c}{ETTh2} & \multicolumn{2}{c}{ETTm1} & \multicolumn{2}{c}{ETTm2} & \multicolumn{2}{c}{Exchange} & \multicolumn{2}{c}{Traffic} & \multicolumn{2}{c}{WTH} & \multirow{2}{*}{Count}                                                                                                                                         \\ \cmidrule{3-18}
                    \multicolumn{2}{c}{}                      & 30                      & 60                        & 30                        & 60                        & 30                        & 60                           & 30                          & 60                      & 30                     & 60           & 30           & 60            & 30            & 60           & 30           & 60           &                            \\ \midrule
                    \multirow{3}{*}{TCN}                      & Gene        & \fires{9.89}              & \fires{9.91}              & \fires{9.87}              & \fires{9.89}              & \fires{19.72}                & \fires{19.78}               & \fires{19.71}           & \fires{19.77}          & \fires{9.87} & \fires{9.89} & 29.56         & \fires{19.78} & \fires{9.87} & \fires{9.89} & \fires{9.89} & \fires{19.79} & \fires{15} \\
                                                              & Softmax                 & 29.50                     & 29.61                     & 29.50                     & 29.61                     & 29.50                        & 29.61                       & 29.50                   & 29.61                  & 29.50        & 29.61        & \fires{29.50} & 29.61         & 29.50        & 29.61        & 29.50        & 29.61         & 1          \\
                                                              & Noisy Top-K             & 29.50                     & 29.61                     & 29.50                     & 29.61                     & 29.50                        & 29.61                       & 29.50                   & 29.61                  & 29.50        & 29.61        & \fires{29.50} & 29.61         & 29.50        & 29.61        & 29.50        & 29.61         & 1          \\ \midrule
                    \multirow{3}{*}{TimesNet}                 & Gene        & \fires{2.64}              & \fires{2.63}              & \fires{2.62}              & \fires{2.62}              & \fires{5.22}                 & \fires{5.22}                & \fires{5.21}            & \fires{5.21}           & \fires{2.62} & \fires{2.62} & 7.80          & \fires{5.21}  & \fires{2.62} & \fires{2.62} & \fires{2.65} & \fires{5.23}  & \fires{15} \\
                                                              & Softmax                 & 7.77                      & 7.79                      & 7.77                      & 7.79                      & 7.77                         & 7.79                        & 7.77                    & 7.79                   & 7.77         & 7.79         & \fires{7.77}  & 7.79          & 7.77         & 7.79         & 7.77         & 7.79          & 1          \\
                                                              & Noisy Top-K             & 7.77                      & 7.80                      & 7.77                      & 7.80                      & 7.77                         & 7.80                        & 7.77                    & 7.80                   & 7.77         & 7.80         & \fires{7.77}  & 7.80          & 7.77         & 7.80         & 7.77         & 7.80          & 1          \\ \midrule
                    \multirow{3}{*}{SegRNN}                   & Gene        & \fires{0.09}              & \fires{0.06}              & \fires{0.07}              & \fires{0.06}              & \fires{0.11}                 & \fires{0.10}                & \fires{0.10}            & \fires{0.09}           & \fires{0.07} & \fires{0.06} & 0.14          & \fires{0.09}  & \fires{0.07} & \fires{0.05} & \fires{0.09} & \fires{0.11}  & \fires{15} \\
                                                              & Softmax                 & 0.12                      & 0.12                      & 0.12                      & 0.12                      & 0.12                         & 0.12                        & 0.12                    & 0.12                   & 0.12         & 0.12         & \fires{0.12}  & 0.12          & 0.12         & 0.12         & 0.12         & 0.12          & 1          \\
                                                              & Noisy Top-K             & 0.12                      & 0.12                      & 0.12                      & 0.12                      & 0.12                         & 0.12                        & 0.12                    & 0.12                   & 0.12         & 0.12         & \fires{0.12}  & 0.12          & 0.12         & 0.12         & 0.12         & 0.12          & 1          \\ \midrule
                    \multirow{3}{*}{DLinear}                  & Gene        & \fires{0.07}              & \fires{0.06}              & \fires{0.05}              & \fires{0.05}              & \fires{0.07}                 & \fires{0.09}                & \fires{0.07}            & \fires{0.08}           & \fires{0.05} & \fires{0.05} & 0.08          & \fires{0.08}  & \fires{0.05} & \fires{0.05} & \fires{0.08} & \fires{0.10}  & \fires{15} \\
                                                              & Softmax                 & 0.06                      & 0.10                      & 0.06                      & 0.10                      & 0.06                         & 0.10                        & 0.06                    & 0.10                   & 0.06         & 0.10         & \fires{0.06}  & 0.10          & 0.06         & 0.10         & 0.06         & 0.10          & 1          \\
                                                              & Noisy Top-K             & 0.06                      & 0.10                      & 0.06                      & 0.10                      & 0.06                         & 0.10                        & 0.06                    & 0.10                   & 0.06         & 0.10         & \fires{0.06}  & 0.10          & 0.06         & 0.10         & 0.06         & 0.10          & 1          \\ \midrule
                    \multirow{3}{*}{PatchTST}                 & Gene        & \fires{0.80}              & \fires{0.81}              & \fires{0.79}              & \fires{0.80}              & \fires{1.55}                 & \fires{1.59}                & \fires{1.54}            & \fires{1.58}           & \fires{0.79} & \fires{0.80} & 2.29          & \fires{1.58}  & \fires{0.79} & \fires{0.80} & \fires{0.81} & \fires{1.60}  & \fires{15} \\
                                                              & Softmax                 & 2.27                      & 2.35                      & 2.27                      & 2.35                      & 2.27                         & 2.35                        & 2.27                    & 2.35                   & 2.27         & 2.35         & \fires{2.27}  & 2.35          & 2.27         & 2.35         & 2.27         & 2.35          & 1          \\
                                                              & Noisy Top-K             & 2.27                      & 2.35                      & 2.27                      & 2.35                      & 2.27                         & 2.35                        & 2.27                    & 2.35                   & 2.27         & 2.35         & \fires{2.27}  & 2.35          & 2.27         & 2.35         & 2.27         & 2.35          & 1          \\ \midrule
                    \multirow{3}{*}{iTransformer}             & Gene        & \fires{0.17}              & \fires{0.16}              & \fires{0.16}              & \fires{0.15}              & \fires{0.29}                 & \fires{0.29}                & \fires{0.28}            & \fires{0.28}           & \fires{0.16} & \fires{0.15} & 0.41          & \fires{0.28}  & \fires{0.16} & \fires{0.15} & \fires{0.18} & \fires{0.30}  & \fires{15} \\
                                                              & Softmax                 & 0.38                      & 0.39                      & 0.38                      & 0.39                      & 0.38                         & 0.39                        & 0.38                    & 0.39                   & 0.38         & 0.39         & \fires{0.38}  & 0.39          & 0.38         & 0.39         & 0.38         & 0.39          & 1          \\
                                                              & Noisy Top-K             & 0.38                      & 0.39                      & 0.38                      & 0.39                      & 0.38                         & 0.39                        & 0.38                    & 0.39                   & 0.38         & 0.39         & \fires{0.38}  & 0.39          & 0.38         & 0.39         & 0.38         & 0.39          & 1          \\ \midrule
                    \multirow{3}{*}{TimeMixer}                & Gene        & \fires{1.00}              & \fires{1.00}              & \fires{0.99}              & \fires{1.00}              & \fires{1.95}                 & \fires{1.98}                & \fires{1.94}            & \fires{1.97}           & \fires{0.99} & \fires{0.99} & 2.89          & \fires{1.97}  & \fires{0.99} & \fires{0.99} & \fires{1.01} & \fires{1.99}  & \fires{15} \\
                                                              & Softmax                 & 2.87                      & 2.93                      & 2.87                      & 2.93                      & 2.87                         & 2.93                        & 2.87                    & 2.93                   & 2.87         & 2.93         & \fires{2.87}  & 2.93          & 2.87         & 2.93         & 2.87         & 2.93          & 1          \\
                                                              & Noisy Top-K             & 2.87                      & 2.93                      & 2.87                      & 2.93                      & 2.87                         & 2.93                        & 2.87                    & 2.93                   & 2.87         & 2.93         & \fires{2.87}  & 2.93          & 2.87         & 2.93         & 2.87         & 2.93          & 1          \\ \bottomrule
                \end{tabular}
                
            \end{small}
        \end{threeparttable}}
\end{table*}

\begin{table*}[htbp]
    \centering
    \caption{Full MSE error of comparison of \method and TSFM. The \fires{best} performances are highlighted.}
    \label{tab:tsfm_mse}
    \setlength{\tabcolsep}{2pt}
    \renewcommand{\arraystretch}{1.2}
    \resizebox{\linewidth}{!}{    
        \begin{threeparttable}    
            \begin{small}    
                \begin{tabular}{ccccccccccccccccc}
                    \toprule
                    \multirow{2}{*}{Data}            & \multicolumn{2}{c}{ECL} & \multicolumn{2}{c}{ETTh1} & \multicolumn{2}{c}{ETTh2} & \multicolumn{2}{c}{ETTm1} & \multicolumn{2}{c}{ETTm2} & \multicolumn{2}{c}{Exchange} & \multicolumn{2}{c}{Traffic} & \multicolumn{2}{c}{WTH}                                                                                                                                         \\ \cmidrule{2-17}
                                                     & 30                      & 60                        & 30                        & 60                        & 30                        & 60                           & 30                          & 60                      & 30             & 60             & 30             & 60             & 30             & 60             & 30             & 60             \\ \midrule
                    TCN+\method          & 0.261                   & 0.358                     & 0.392                     & 0.526                     & 2.164                     & 3.443                        & 0.638                       & 0.964                   & 0.219          & 0.309          & 0.272          & 0.553          & 0.668          & 0.714          & 0.314          & 0.450          \\
                    TimesNet+\method     & 0.259                   & 0.319                     & 0.291                     & 0.390                     & 2.054                     & 3.065                        & 0.639                       & 0.947                   & 0.234          & 0.277          & 0.259          & 0.530          & 0.492          & 0.597          & 0.323          & 0.404          \\
                    SegRNN+\method       & \secres{0.239}          & 0.312                     & \fires{0.248}             & \fires{0.353}             & 1.840                     & 3.328                        & \fires{0.551}               & \secres{0.881}          & \secres{0.197} & \secres{0.259} & \fires{0.192}  & \secres{0.395} & 0.693          & 0.930          & \secres{0.298} & \fires{0.385}  \\
                    DLinear+\method      & 0.264                   & 0.335                     & 0.259                     & \secres{0.356}            & 2.052                     & 3.209                        & 0.586                       & 0.917                   & \fires{0.196}  & 0.259          & 0.261          & 0.494          & 0.614          & 0.641          & \fires{0.295}  & \secres{0.386} \\
                    PatchTST+\method     & 0.265                   & 0.328                     & 0.276                     & 0.377                     & 1.922                     & 2.995                        & \secres{0.567}              & \fires{0.875}           & 0.208          & 0.264          & 0.234          & 0.467          & 0.582          & 0.589          & 0.310          & 0.397          \\
                    iTransformer+\method & \fires{0.229}           & \fires{0.282}             & 0.275                     & 0.372                     & 1.968                     & 3.116                        & 0.594                       & 0.886                   & 0.210          & 0.265          & 0.249          & 0.484          & \secres{0.437} & \secres{0.534} & 0.311          & 0.394          \\
                    TimeMixer+\method    & 0.244                   & \secres{0.297}            & 0.274                     & 0.371                     & 2.000                     & 3.016                        & 0.579                       & 0.895                   & 0.216          & \fires{0.254}  & 0.233          & 0.450          & \fires{0.433}  & \fires{0.524}  & 0.313          & 0.396          \\ \midrule
                    Chronos 2                        & 0.315                   & 0.394                     & 0.250                     & 0.399                     & 1.909                     & 3.287                        & 0.665                       & 1.087                   & 0.311          & 0.525          & 0.208          & \fires{0.389}  & 0.559          & 0.668          & 0.367          & 0.553          \\
                    Moirai 2.0                       & 0.316                   & 0.409                     & 0.259                     & 0.370                     & \secres{1.769}            & \fires{2.769}                & 0.635                       & 1.046                   & 0.282          & 0.419          & \secres{0.204} & 0.441          & 0.625          & 0.798          & 0.332          & 0.469          \\
                    TimesFM 2.5                      & 0.316                   & 0.394                     & \secres{0.249}            & 0.367                     & \fires{1.684}             & \secres{2.790}               & 0.624                       & 1.008                   & 0.297          & 0.424          & 0.206          & 0.432          & 0.557          & 0.637          & 0.318          & 0.429          \\ \bottomrule
                \end{tabular}
            \end{small}
        \end{threeparttable}}
\end{table*}

\begin{table*}[htbp]
    \centering
    \caption{Full weight sizes (MB) of \method and TSFM.}
    \label{tab:tsfm_size}
    \setlength{\tabcolsep}{2pt}
    \renewcommand{\arraystretch}{1.2}
    \resizebox{\linewidth}{!}{    
        \begin{threeparttable}    
            \begin{small}    
                \begin{tabular}{ccccccccccccccccc|c}
                    \toprule
                    \multirow{2}{*}{Data}            & \multicolumn{2}{c}{ECL} & \multicolumn{2}{c}{ETTh1} & \multicolumn{2}{c}{ETTh2} & \multicolumn{2}{c}{ETTm1} & \multicolumn{2}{c}{ETTm2} & \multicolumn{2}{c}{Exchange} & \multicolumn{2}{c}{Traffic} & \multicolumn{2}{c}{WTH} & \multirow{2}{*}{Rank}                                                                     \\ \cmidrule{2-17}
                                                     & 30                      & 60                        & 30                        & 60                        & 30                        & 60                           & 30                          & 60                      & 30                    & 60     & 30     & 60     & 30     & 60     & 30     & 60     &    \\ \midrule
                    DLinear+\method      & 0.33                    & 0.29                      & 0.25                      & 0.25                      & 0.36                      & 0.45                         & 0.33                        & 0.40                    & 0.24                  & 0.25   & 0.41   & 0.40   & 0.24   & 0.25   & 0.38   & 0.50   & 1  \\
                    SegRNN+\method       & 0.43                    & 0.32                      & 0.35                      & 0.28                      & 0.56                      & 0.50                         & 0.52                        & 0.46                    & 0.34                  & 0.28   & 0.70   & 0.46   & 0.34   & 0.27   & 0.47   & 0.56   & 2  \\
                    iTransformer+\method & 0.87                    & 0.79                      & 0.79                      & 0.75                      & 1.45                      & 1.43                         & 1.41                        & 1.39                    & 0.79                  & 0.74   & 2.03   & 1.38   & 0.79   & 0.74   & 0.92   & 1.49   & 3  \\
                    PatchTST+\method     & 4.02                    & 4.04                      & 3.94                      & 4.01                      & 7.74                      & 7.95                         & 7.71                        & 7.90                    & 3.94                  & 4.00   & 11.47  & 7.89   & 3.94   & 4.00   & 4.07   & 8.00   & 4  \\
                    TimeMixer+\method    & 5.02                    & 5.01                      & 4.94                      & 4.98                      & 9.73                      & 9.89                         & 9.70                        & 9.84                    & 4.93                  & 4.97   & 14.45  & 9.83   & 4.93   & 4.97   & 5.06   & 9.94   & 5  \\
                    TimesNet+\method     & 13.20                   & 13.13                     & 13.12                     & 13.09                     & 26.10                     & 26.12                        & 26.07                       & 26.07                   & 13.12                 & 13.09  & 39.01  & 26.06  & 13.12  & 13.09  & 13.25  & 26.17  & 6  \\
                    TCN+\method          & 49.46                   & 49.54                     & 49.34                     & 49.46                     & 98.60                     & 98.92                        & 98.57                       & 98.87                   & 49.34                 & 49.45  & 147.79 & 98.88  & 49.33  & 49.45  & 49.47  & 98.97  & 8  \\ \midrule
                    Moirai 2.0                       & 43.45                   & 43.45                     & 43.45                     & 43.45                     & 43.45                     & 43.45                        & 43.45                       & 43.45                   & 43.45                 & 43.45  & 43.45  & 43.45  & 43.45  & 43.45  & 43.45  & 43.45  & 7  \\
                    Chronos 2                        & 455.79                  & 455.79                    & 455.79                    & 455.79                    & 455.79                    & 455.79                       & 455.79                      & 455.79                  & 455.79                & 455.79 & 455.79 & 455.79 & 455.79 & 455.79 & 455.79 & 455.79 & 9  \\
                    TimesFM 2.5                      & 882.32                  & 882.32                    & 882.32                    & 882.32                    & 882.32                    & 882.32                       & 882.32                      & 882.32                  & 882.32                & 882.32 & 882.32 & 882.32 & 882.32 & 882.32 & 882.32 & 882.32 & 10 \\ \bottomrule
                \end{tabular}
            \end{small}
        \end{threeparttable}}
\end{table*}

\begin{figure*}[!t]
    \centering
    \subfigure[ECL, $\textbf{H}=30$]{
        \includegraphics[width = 0.48\linewidth]{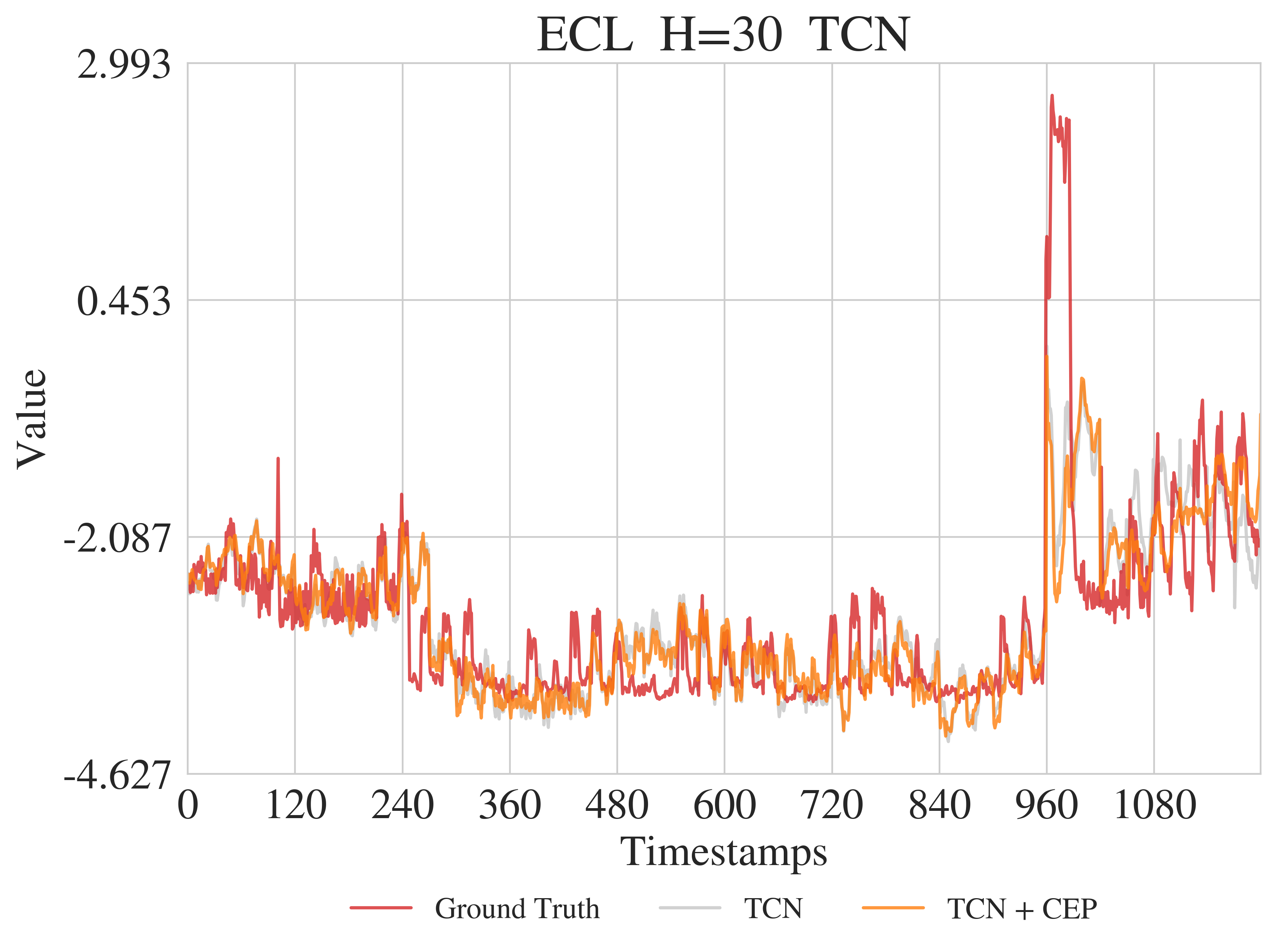}
    }\hfill
    \subfigure[ECL, $\textbf{H}=60$]{
        \includegraphics[width = 0.48\linewidth]{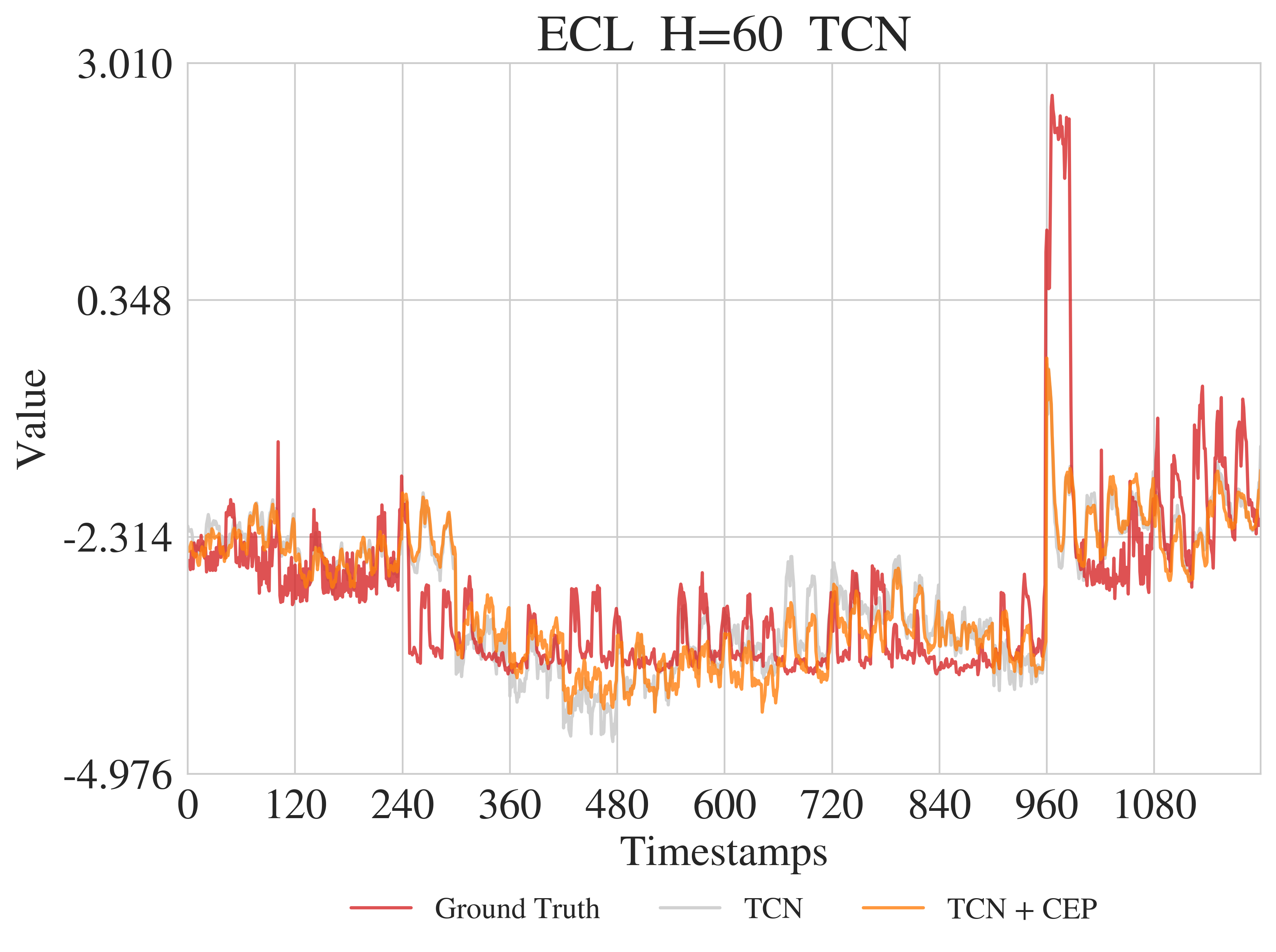}
    }\\[2pt]
    \subfigure[ETTh1, $\textbf{H}=30$]{
        \includegraphics[width = 0.48\linewidth]{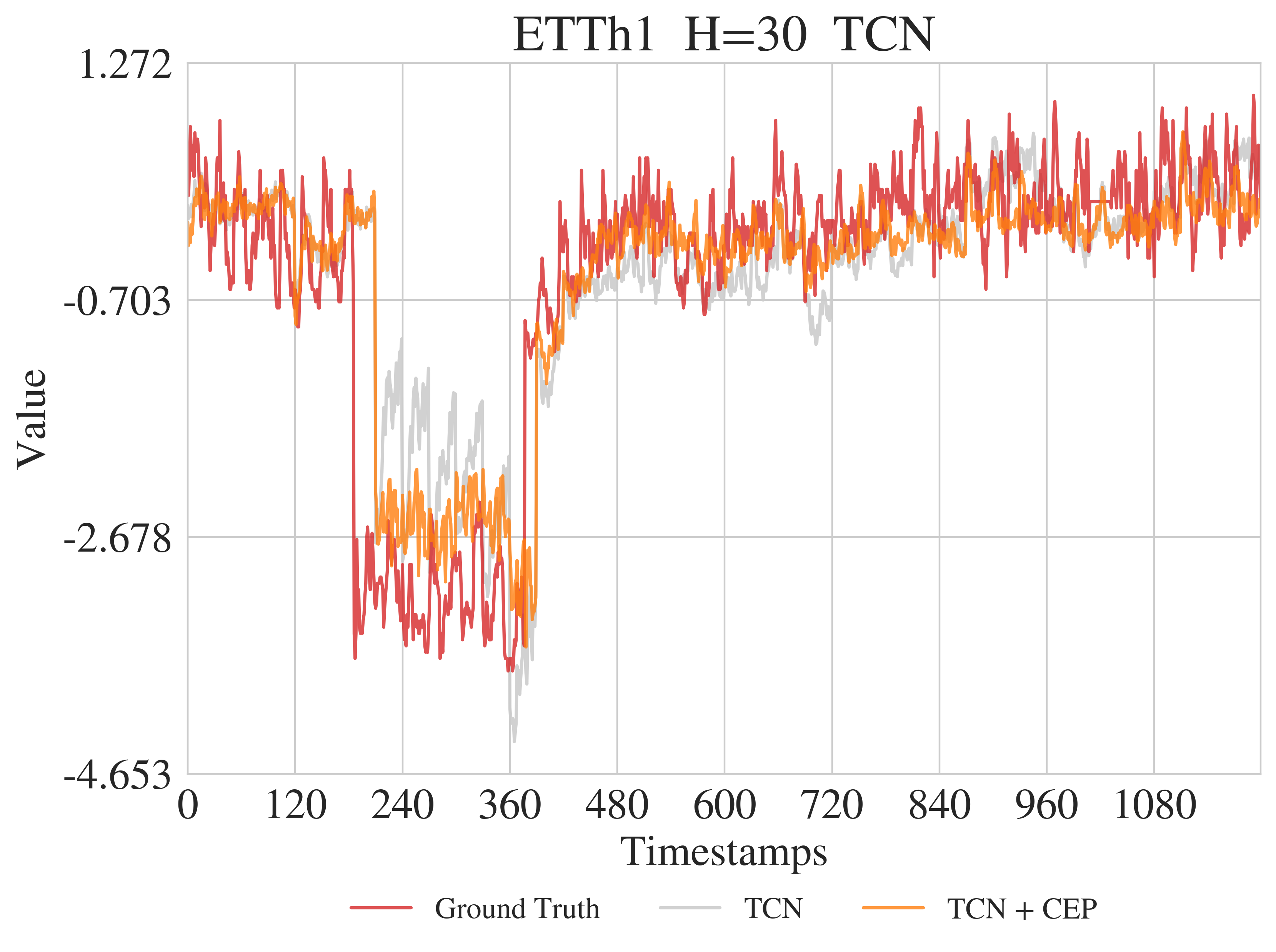}
    }\hfill
    \subfigure[ETTh1, $\textbf{H}=60$]{
        \includegraphics[width = 0.48\linewidth]{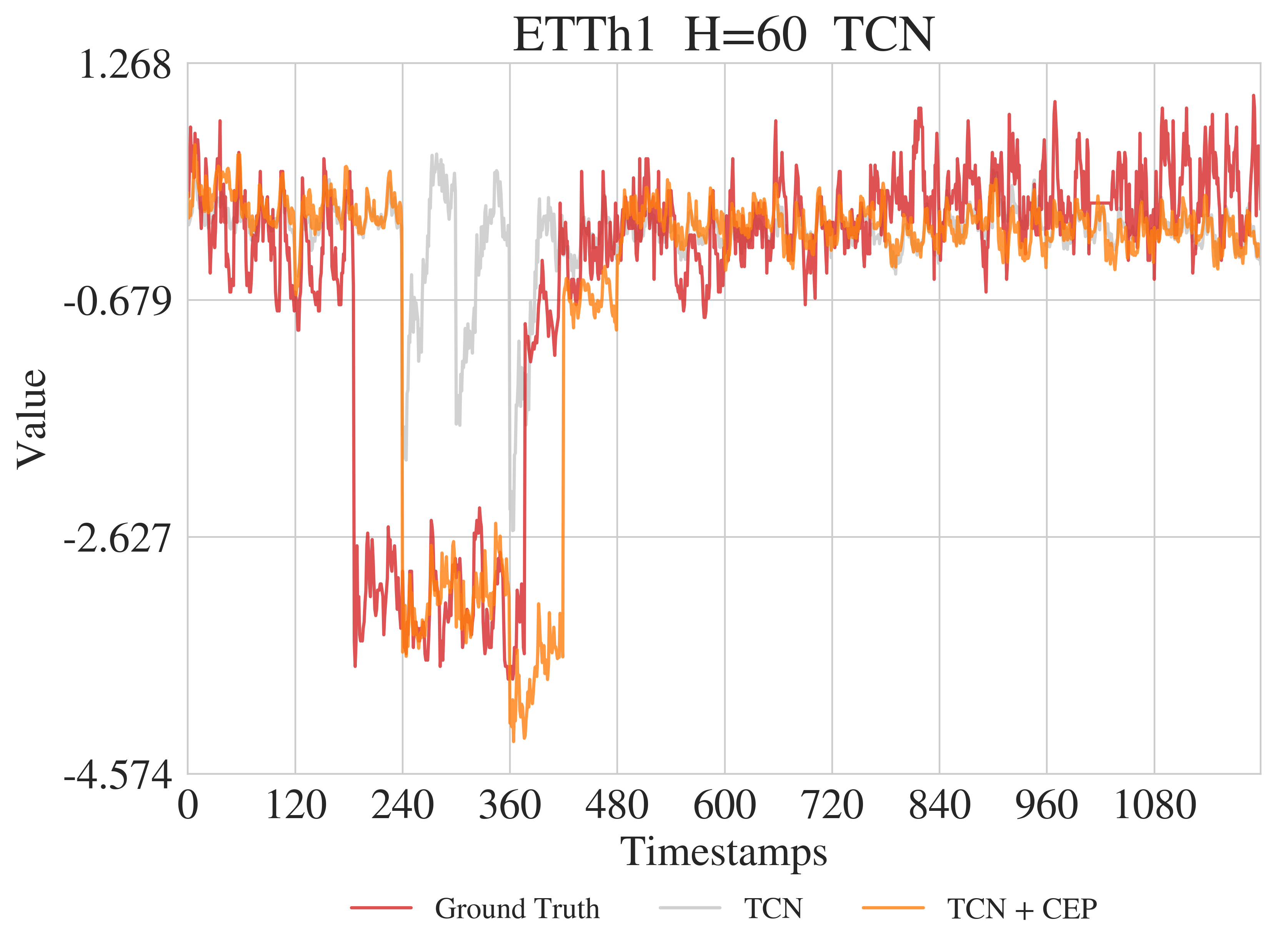}
    }\\[2pt]
    \subfigure[ETTh2, $\textbf{H}=30$]{
        \includegraphics[width = 0.48\linewidth]{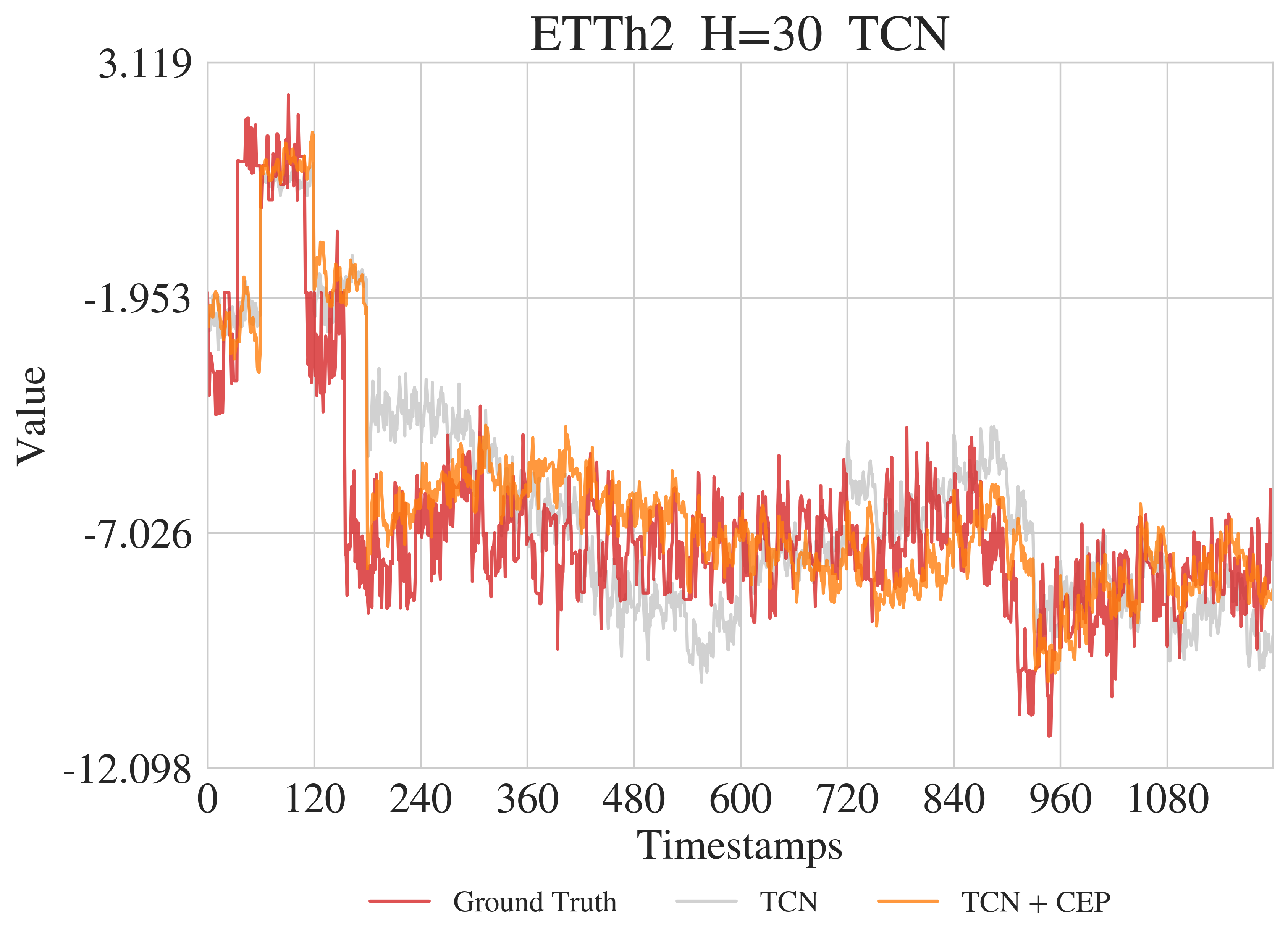}
    }\hfill
    \subfigure[ETTh2, $\textbf{H}=60$]{
        \includegraphics[width = 0.48\linewidth]{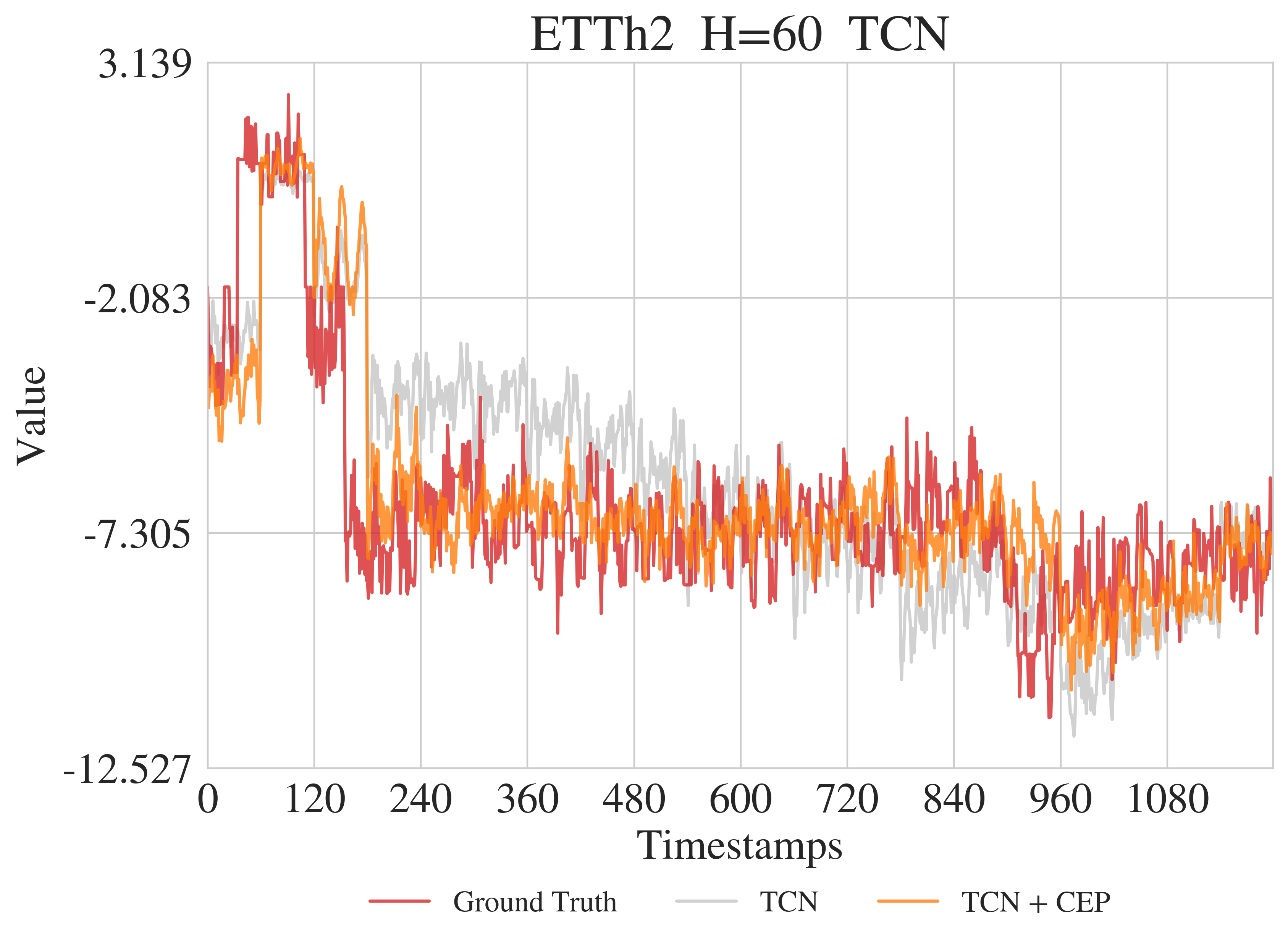}
    }
    \caption{Visualization of results by base forecasters vs. forecasters with \method across all datasets}
    \label{fig:f_r}
\end{figure*}

\clearpage
\begin{figure*}[!t]
    \addtocounter{figure}{-1}
    \setcounter{subfigure}{6}
    \centering
    \subfigure[ETTm1, $\textbf{H}=30$]{
        \includegraphics[width = 0.48\linewidth]{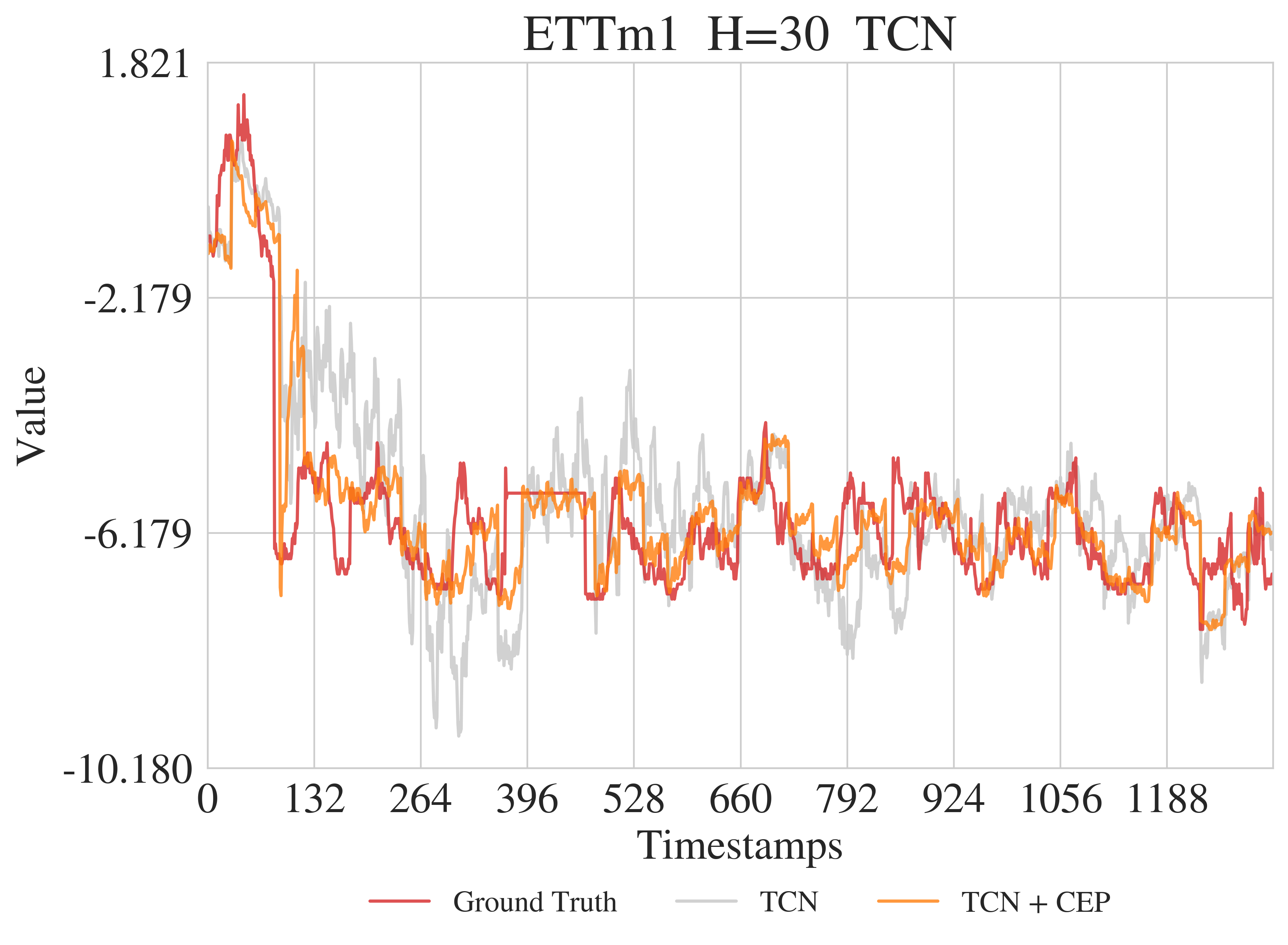}
    }\hfill
    \subfigure[ETTm1, $\textbf{H}=60$]{
        \includegraphics[width = 0.48\linewidth]{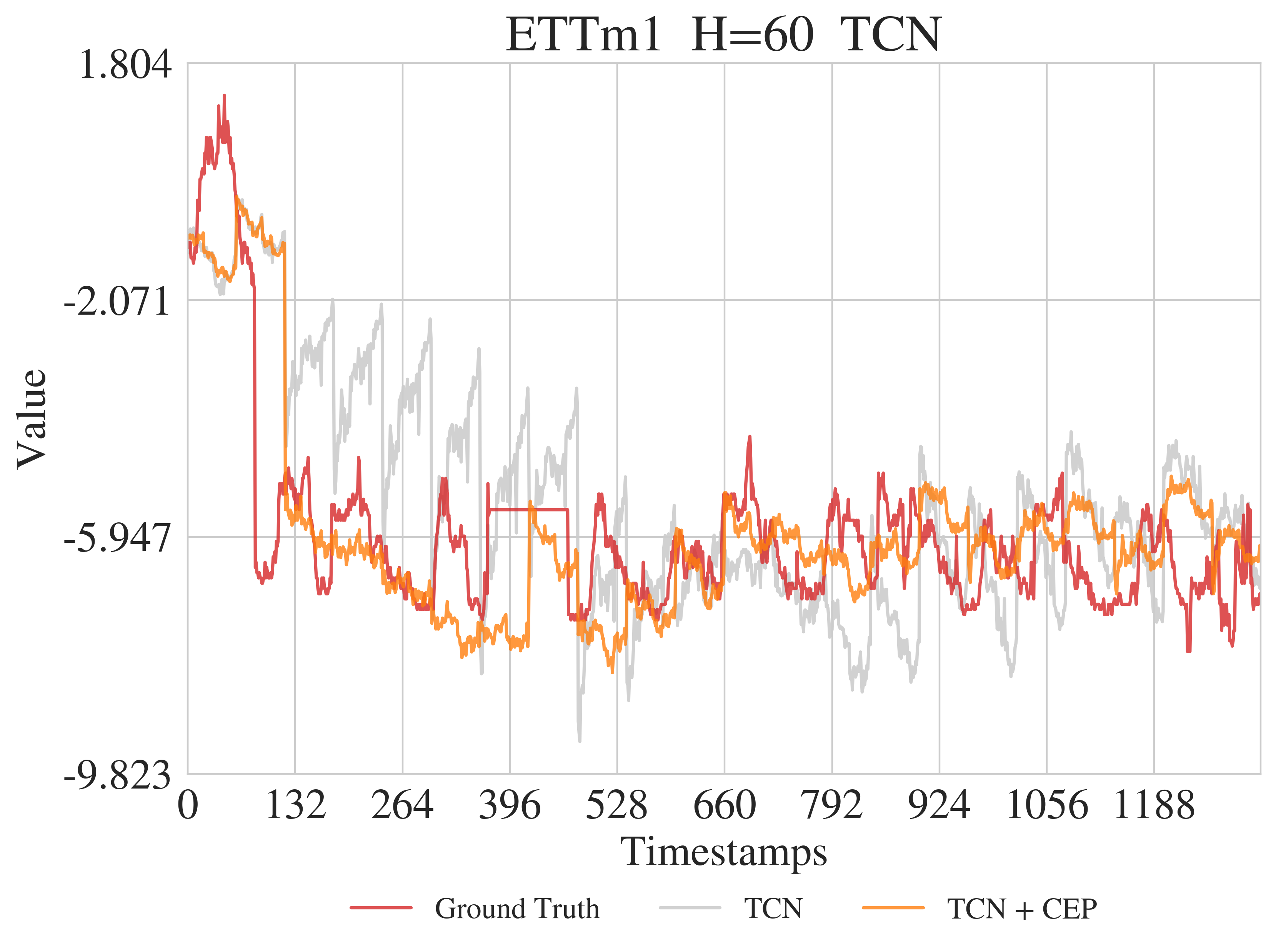}
    }\\[2pt]
    \subfigure[ETTm2, $\textbf{H}=30$]{
        \includegraphics[width = 0.48\linewidth]{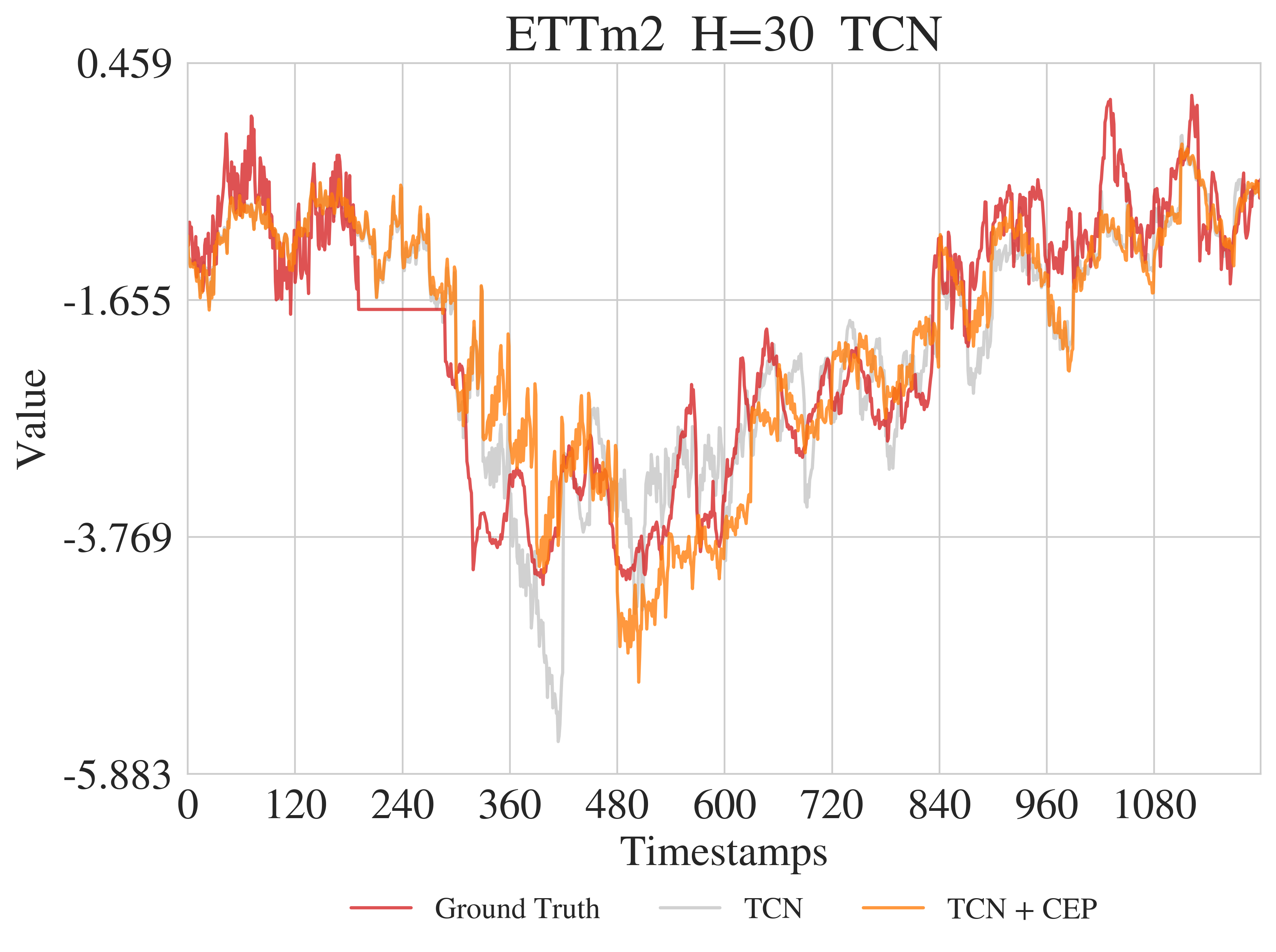}
    }\hfill
    \subfigure[ETTm2, $\textbf{H}=60$]{
        \includegraphics[width = 0.48\linewidth]{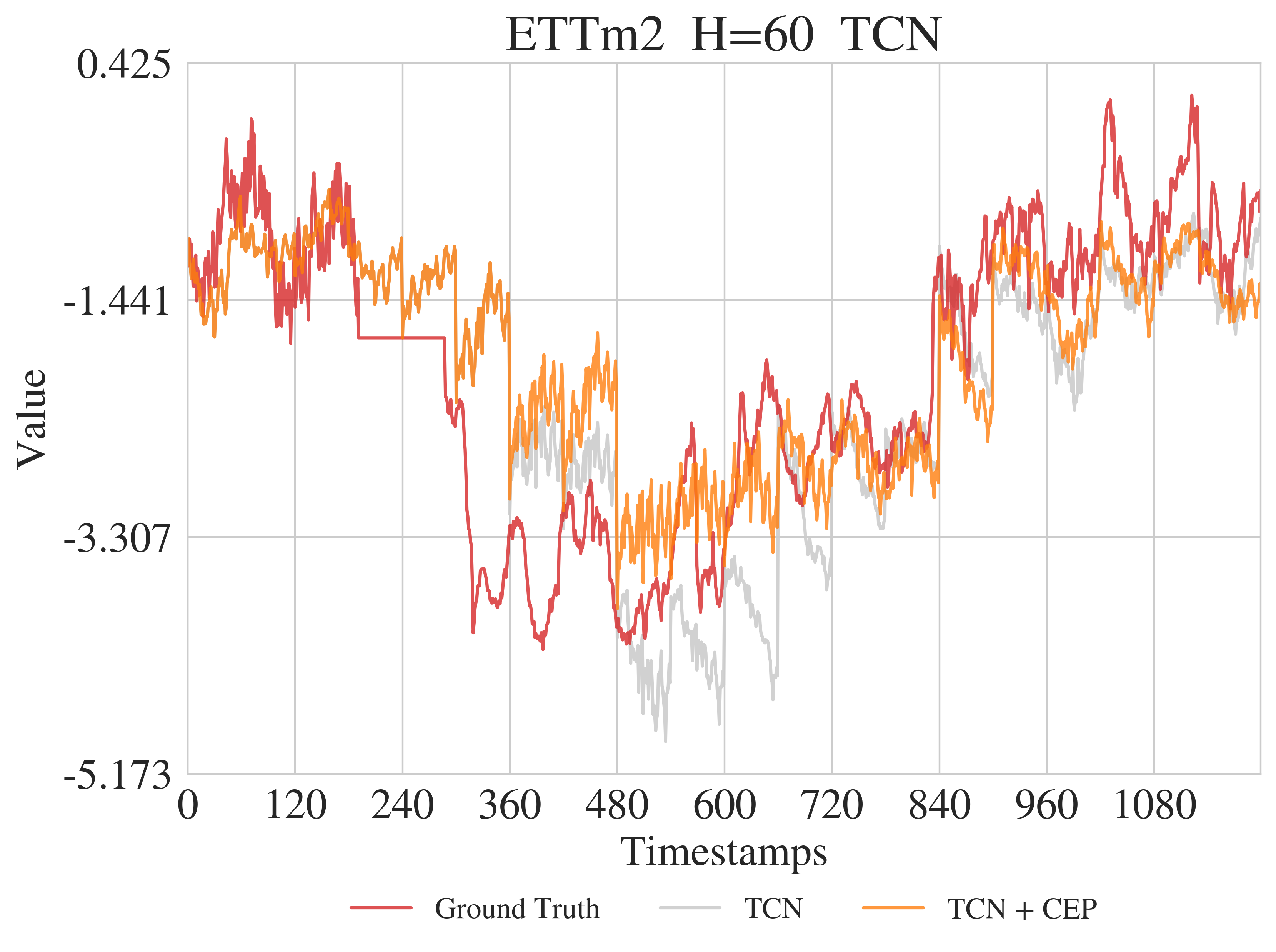}
    }\\[2pt]
    \subfigure[Exchange, $\textbf{H}=30$]{
        \includegraphics[width = 0.48\linewidth]{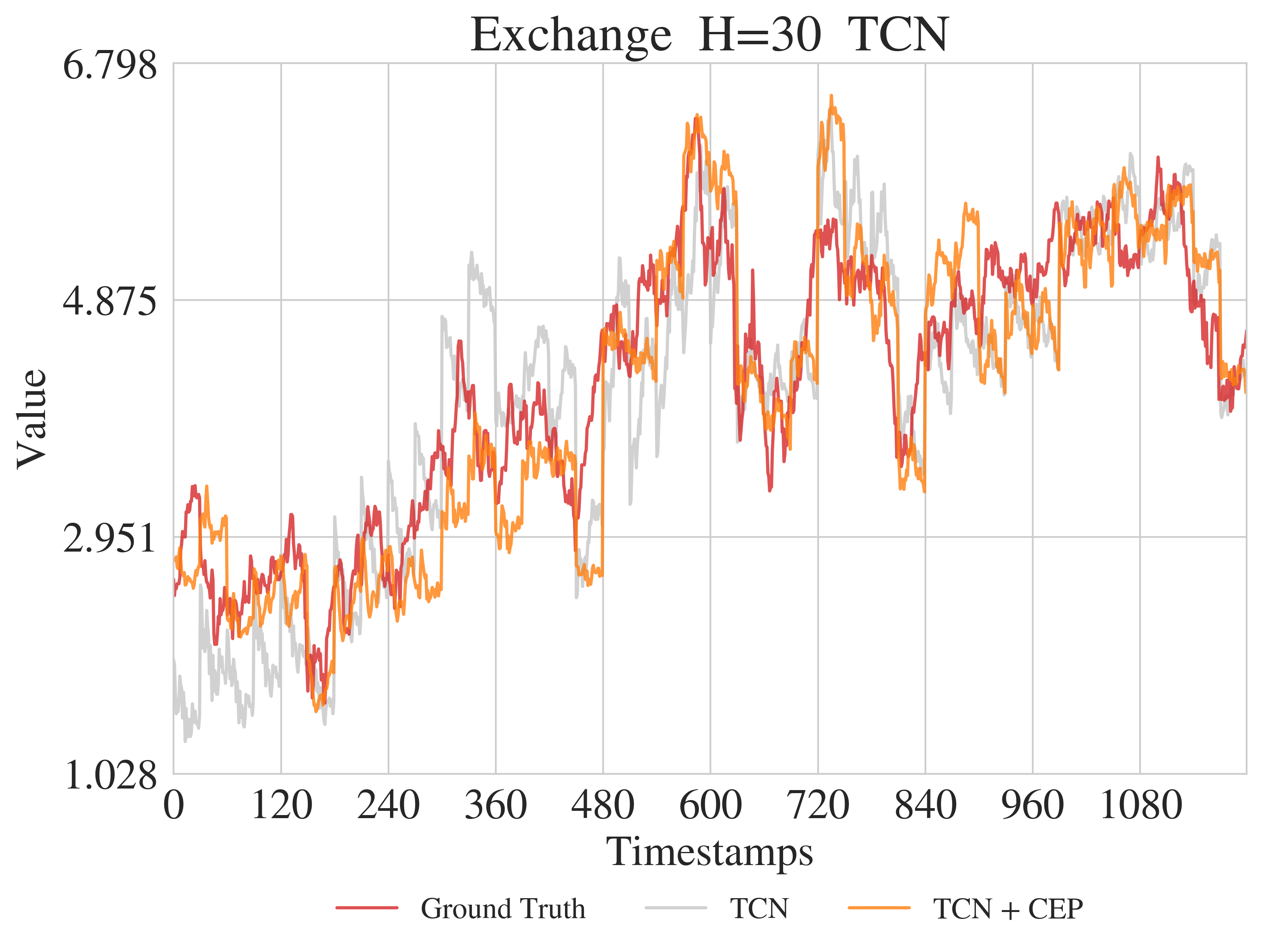}
    }\hfill
    \subfigure[Exchange, $\textbf{H}=60$]{
        \includegraphics[width = 0.48\linewidth]{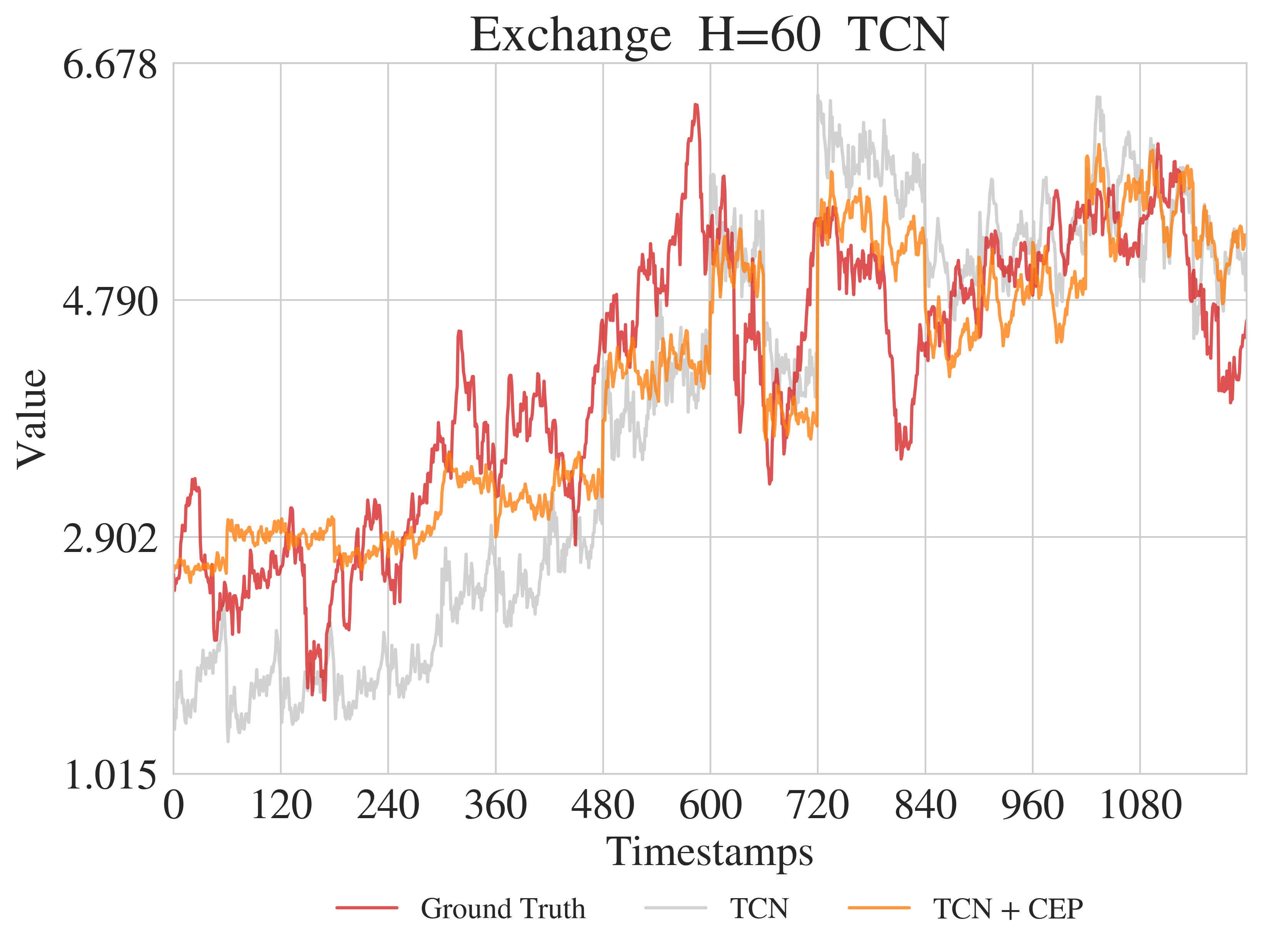}
    }
    \caption[]{Visualization of results by base forecasters vs. forecasters with \method across all datasets (continued)}
\end{figure*}

\clearpage
\begin{figure*}[!t]
    \addtocounter{figure}{-1}
    \setcounter{subfigure}{12}
    \centering
    \subfigure[Traffic, $\textbf{H}=30$]{
        \includegraphics[width = 0.48\linewidth]{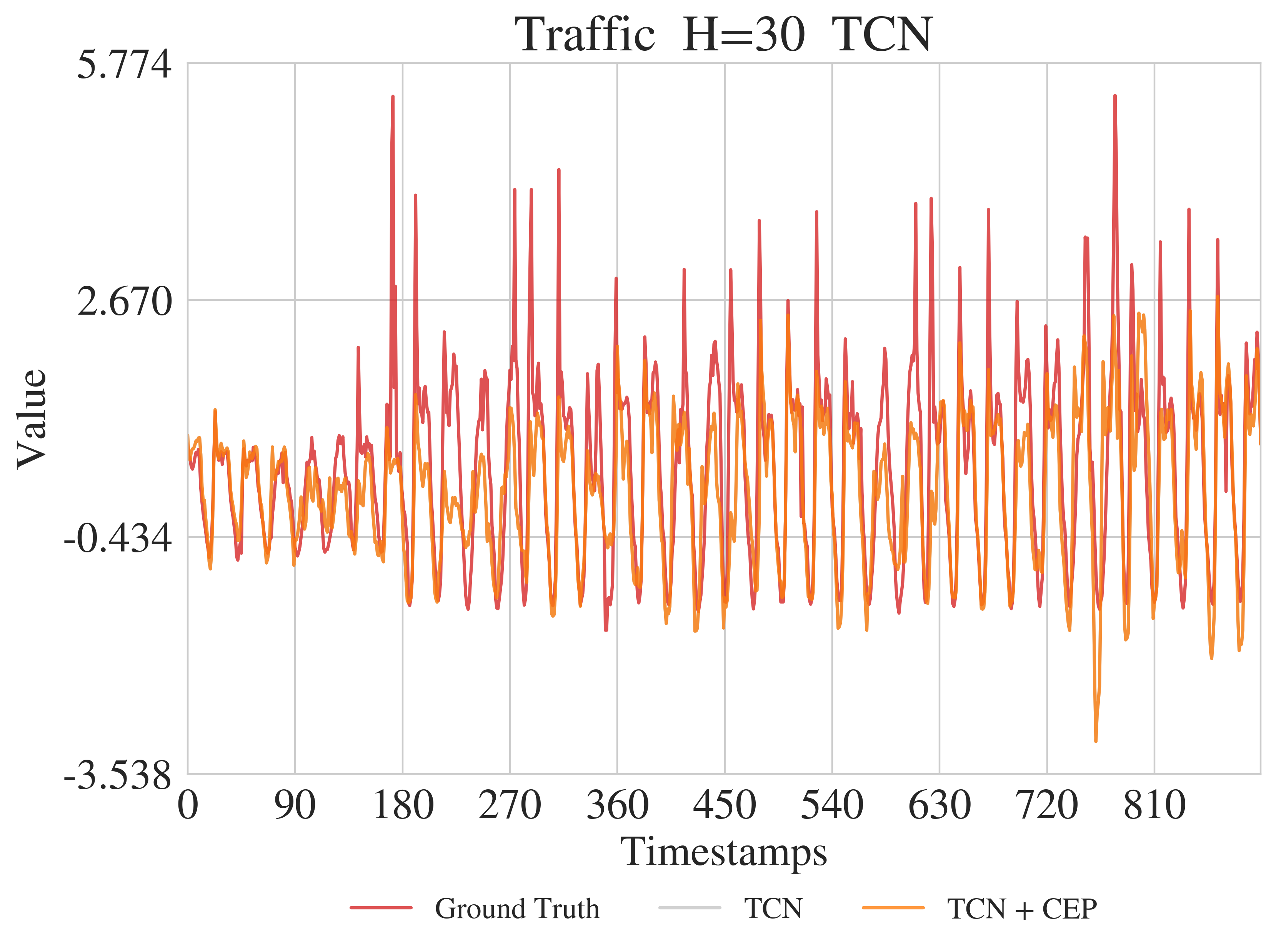}
    }\hfill
    \subfigure[Traffic, $\textbf{H}=60$]{
        \includegraphics[width = 0.48\linewidth]{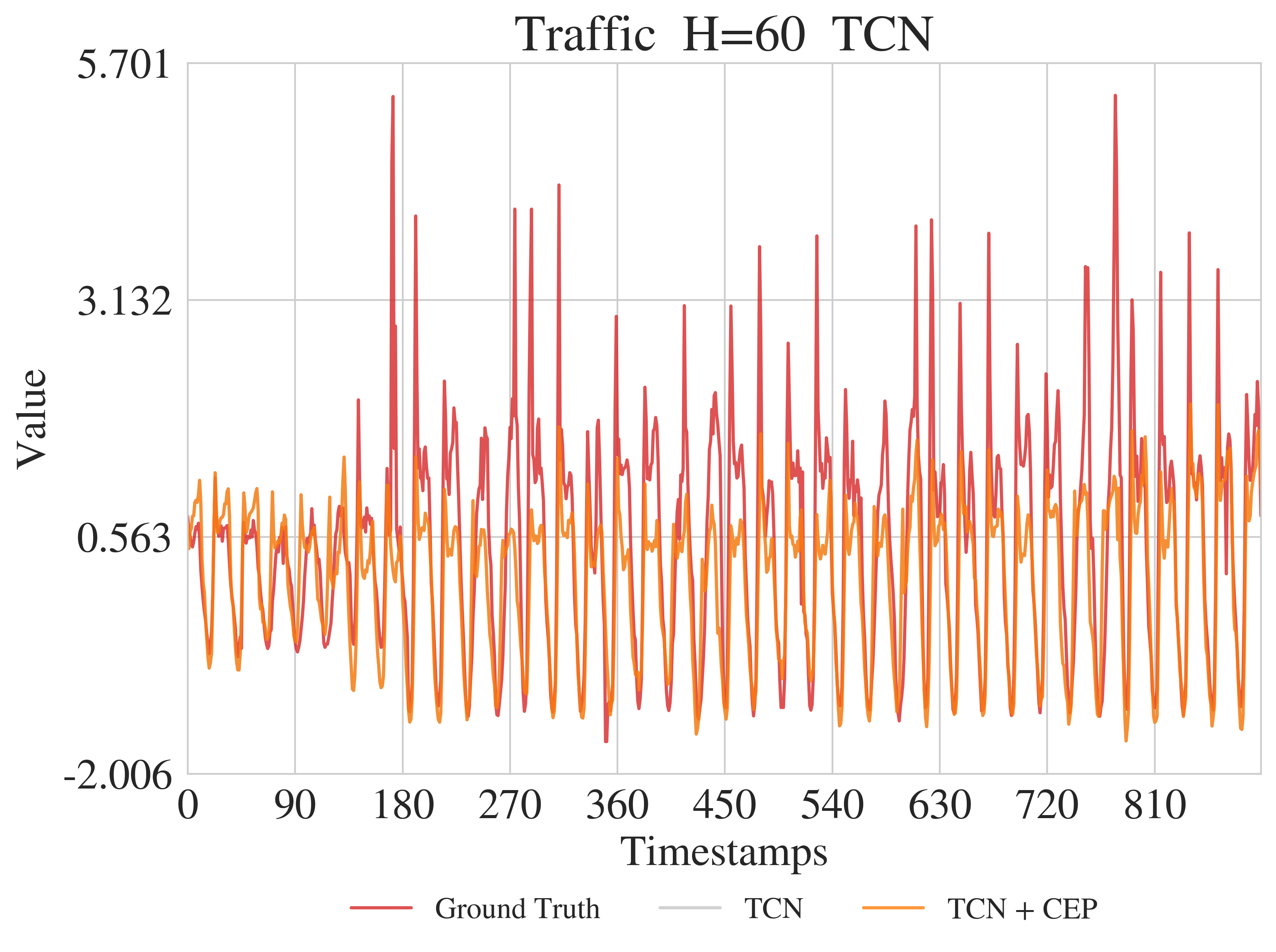}
    }\\[2pt]
    \subfigure[WTH, $\textbf{H}=30$]{
        \includegraphics[width = 0.48\linewidth]{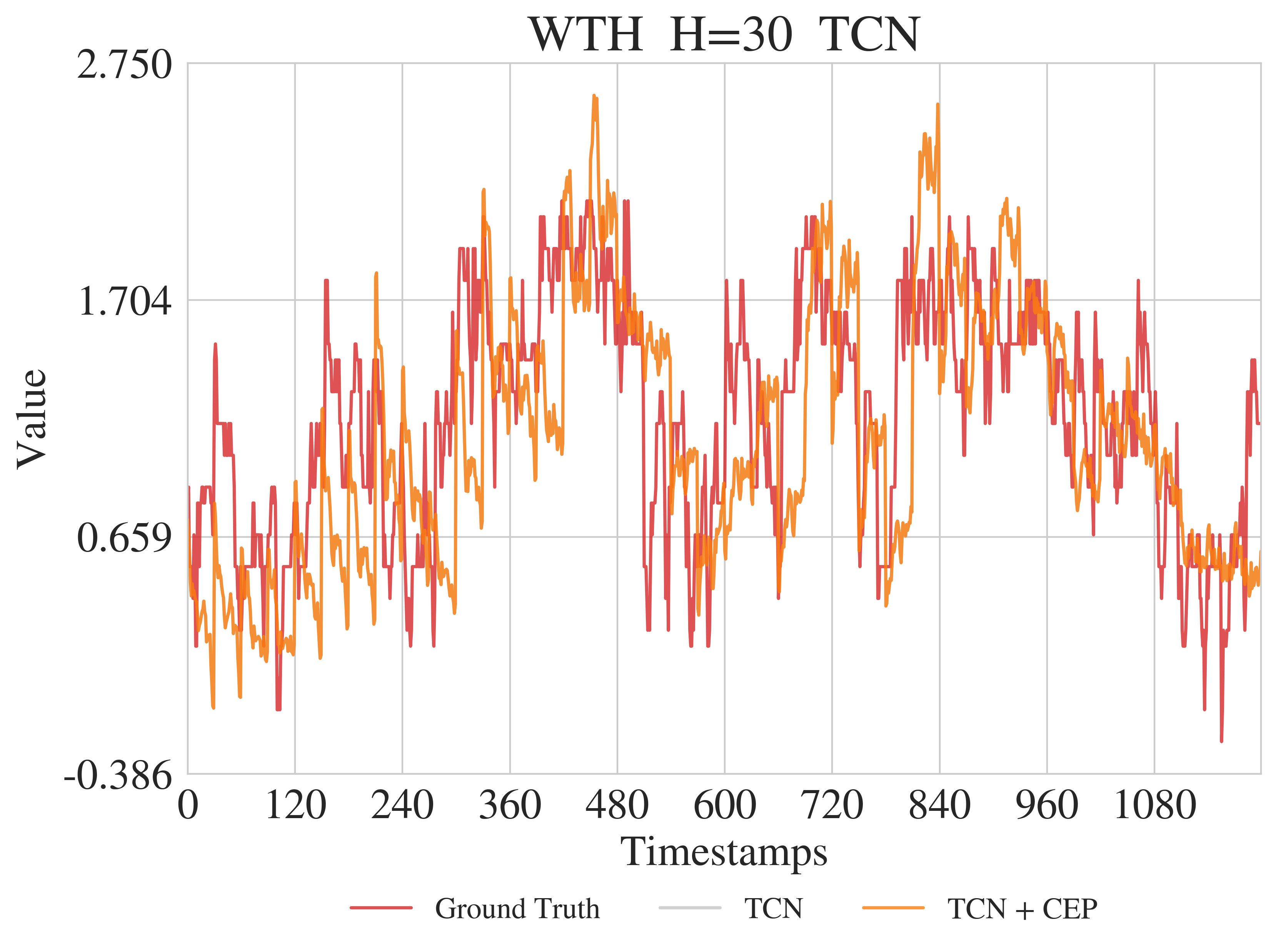}
    }\hfill
    \subfigure[WTH, $\textbf{H}=60$]{
        \includegraphics[width = 0.48\linewidth]{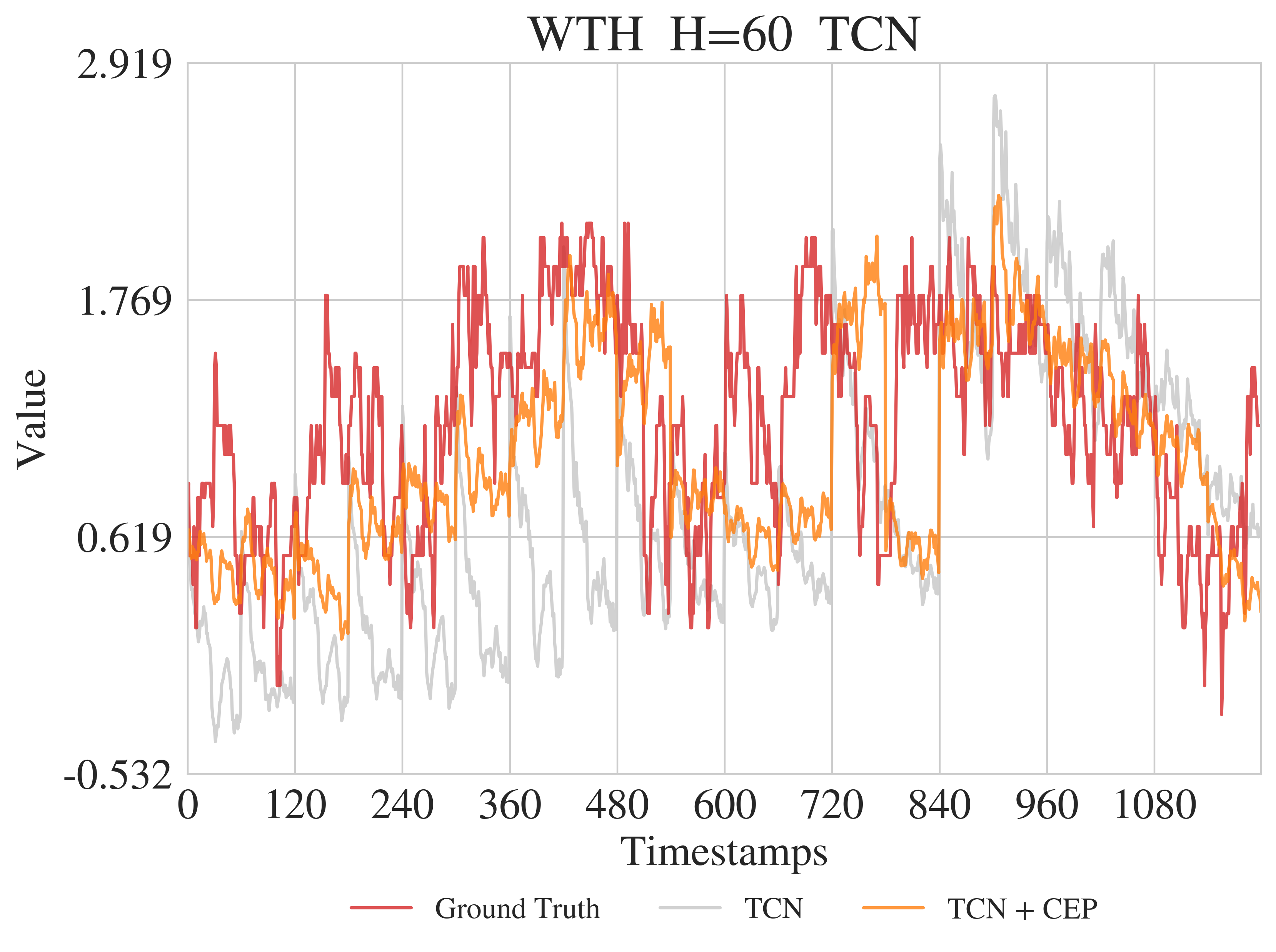}
    }
    \caption[]{Visualization of results by base forecasters vs. forecasters with \method across all datasets (continued)}
\end{figure*}

\begin{figure*}[!t]
    \centering
    \subfigure[ECL, Lifecycle]{                    
        \includegraphics[width = 0.48\linewidth]{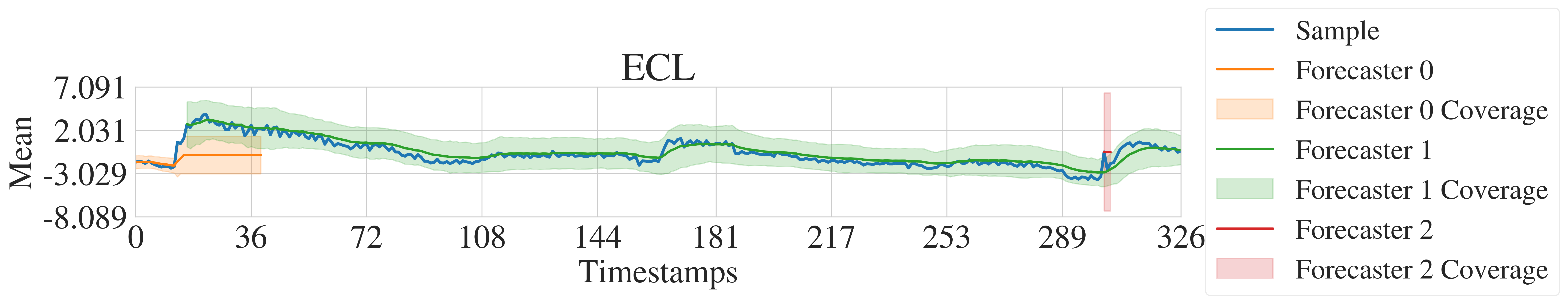}
    }\hfill
    \subfigure[ECL, Internal Prediction]{                    
        \includegraphics[width = 0.48\linewidth]{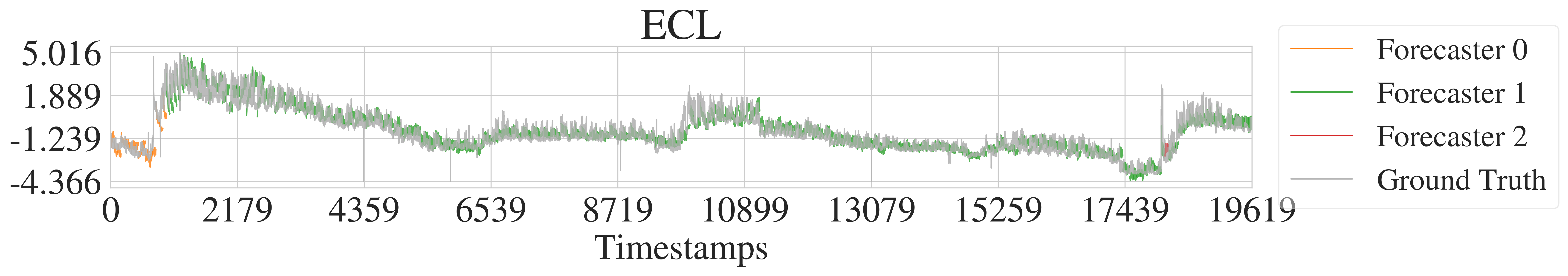}
    }\\[2pt]
    \subfigure[ETTh1, Lifecycle]{                    
        \includegraphics[width = 0.48\linewidth]{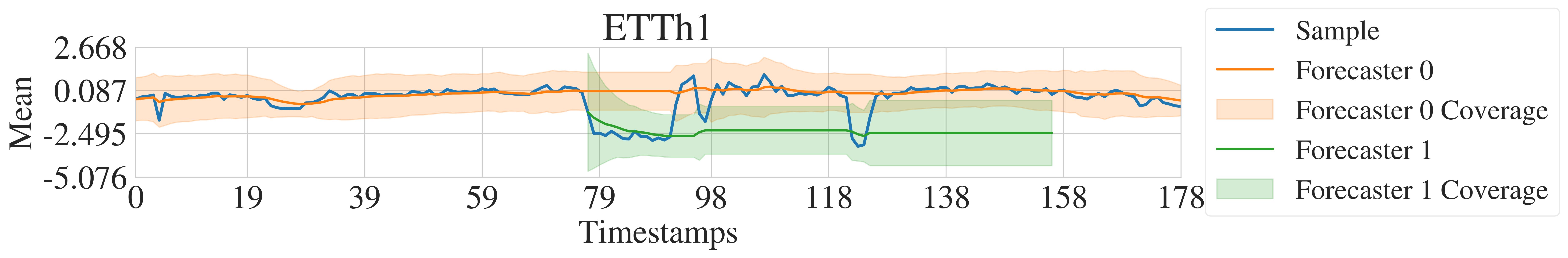}
    }\hfill
    \subfigure[ETTh1, Internal Prediction]{                    
        \includegraphics[width = 0.48\linewidth]{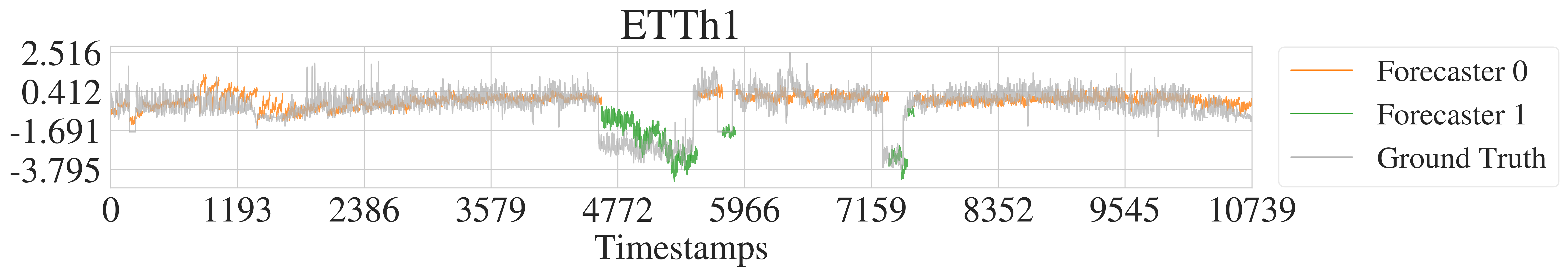}
    }\\[2pt]
    \subfigure[ETTh2, Lifecycle]{                    
        \includegraphics[width = 0.48\linewidth]{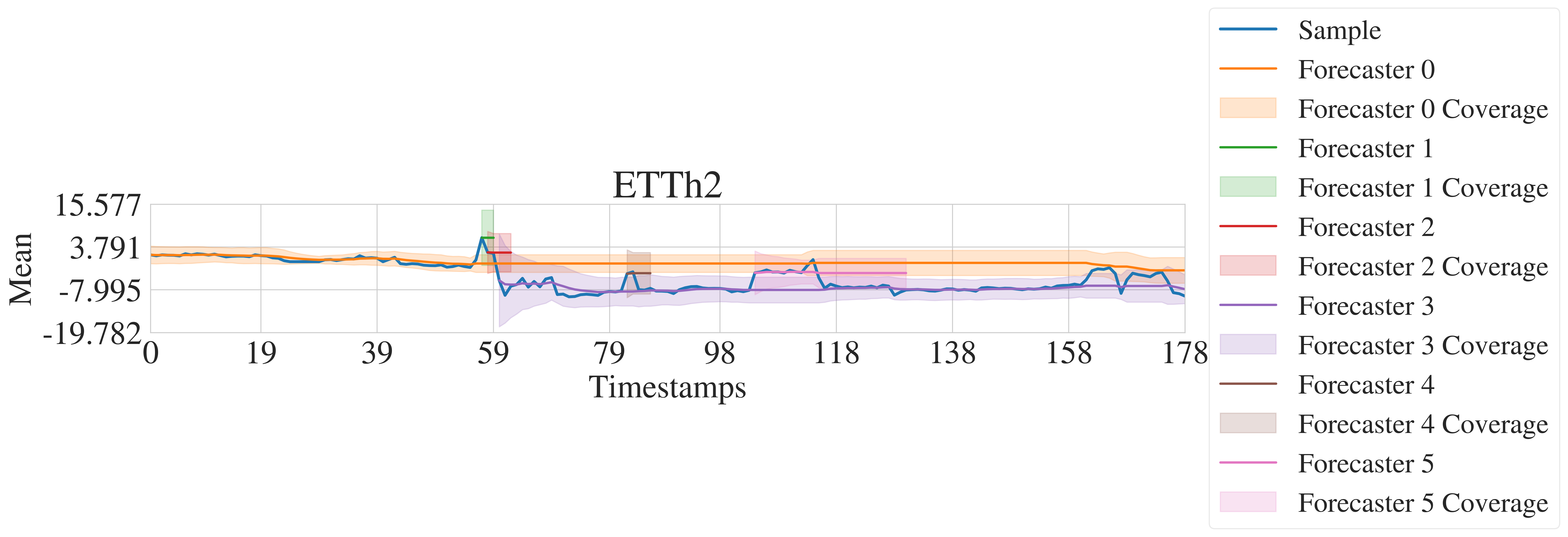}
    }\hfill
    \subfigure[ETTh2, Internal Prediction]{                    
        \includegraphics[width = 0.48\linewidth]{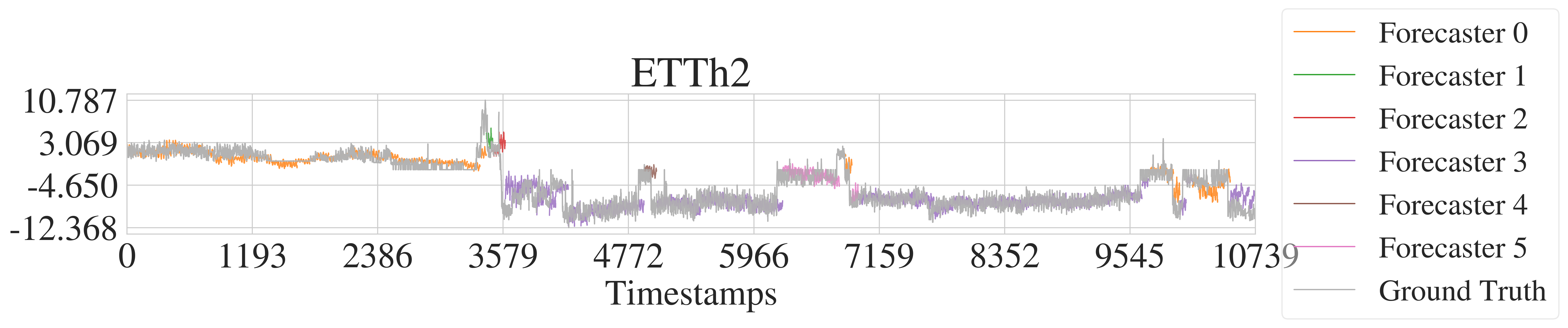}
    }\hfill
    \subfigure[ETTm1, Lifecycle]{                    
        \includegraphics[width = 0.48\linewidth]{./life/ETTm1_60.png}
    }\hfill
    \subfigure[ETTm1, Internal Prediction]{                    
        \includegraphics[width = 0.48\linewidth]{./traj/ETTm1_60.png}
    }\\[2pt]
    \subfigure[ETTm2, Lifecycle]{                    
        \includegraphics[width = 0.48\linewidth]{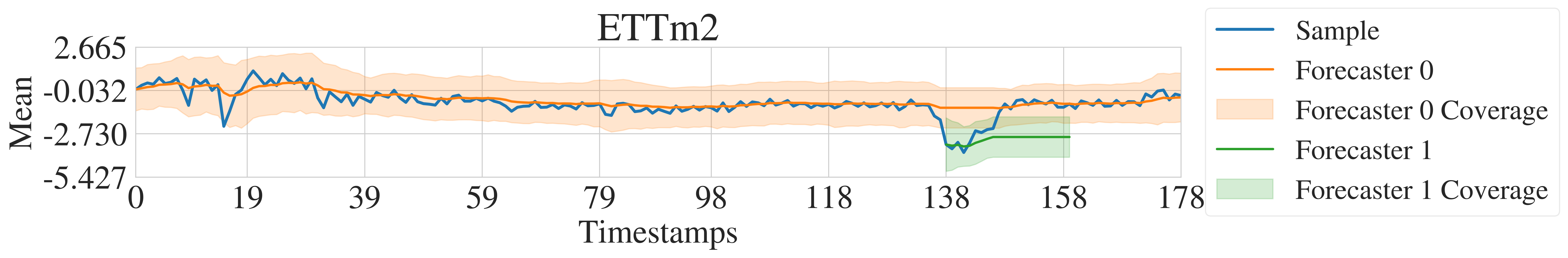}
    }\hfill
    \subfigure[ETTm2, Internal Prediction]{                    
        \includegraphics[width = 0.48\linewidth]{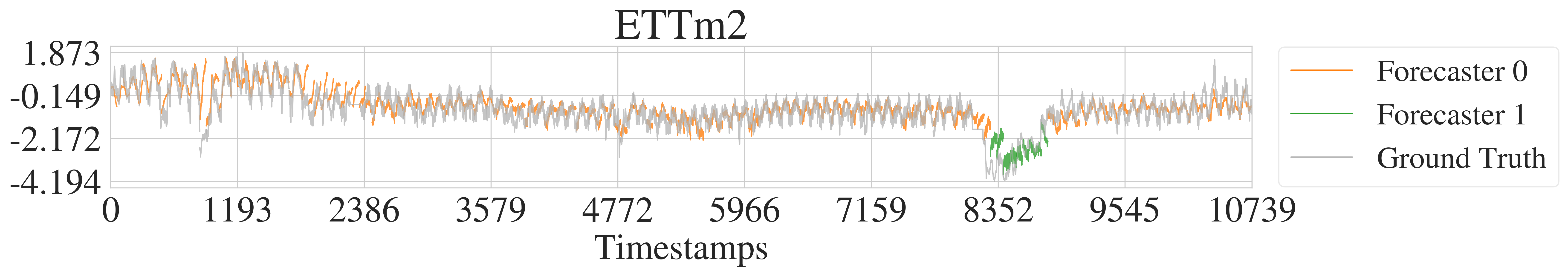}
    }\\[2pt]
    \subfigure[Exchange, Lifecycle]{                    
        \includegraphics[width = 0.48\linewidth]{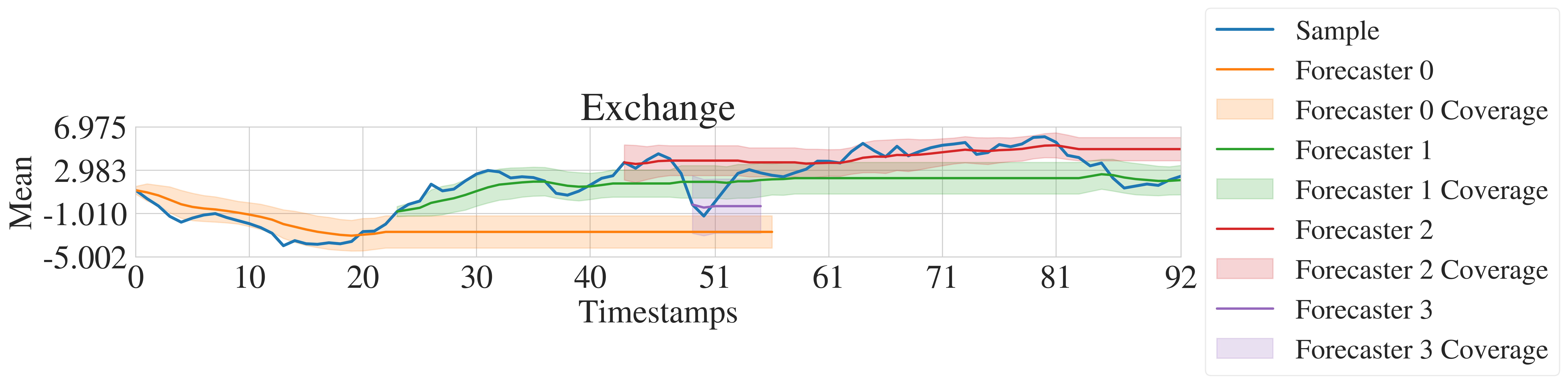}
    }\hfill
    \subfigure[Exchange, Internal Prediction]{                    
        \includegraphics[width = 0.48\linewidth]{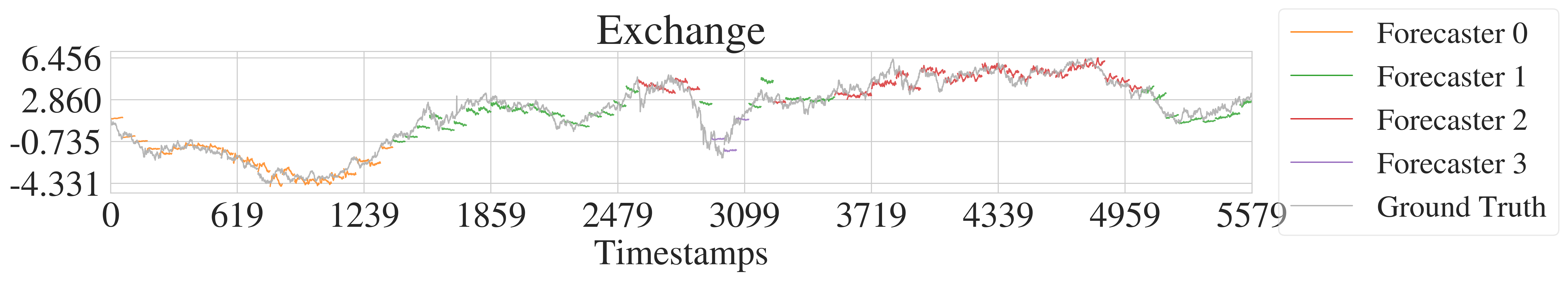}
    }\hfill
    \subfigure[Traffic, Lifecycle]{                    
        \includegraphics[width = 0.48\linewidth]{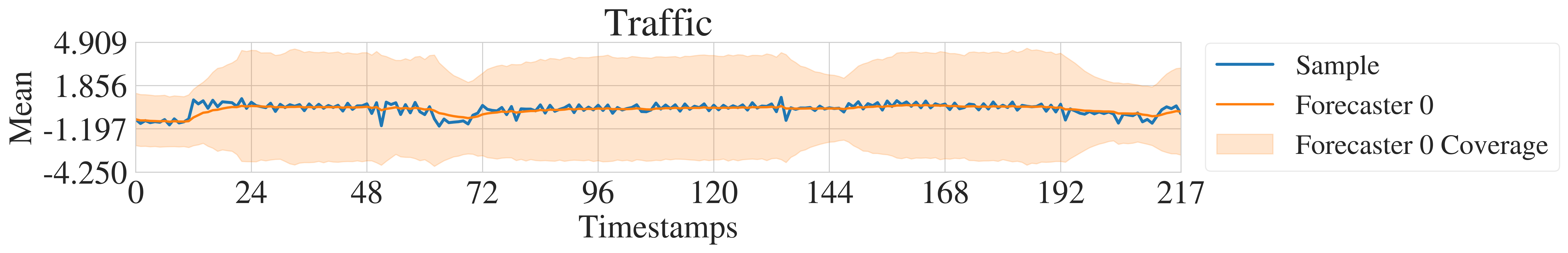}
    }\hfill
    \subfigure[Traffic, Internal Prediction]{                    
        \includegraphics[width = 0.48\linewidth]{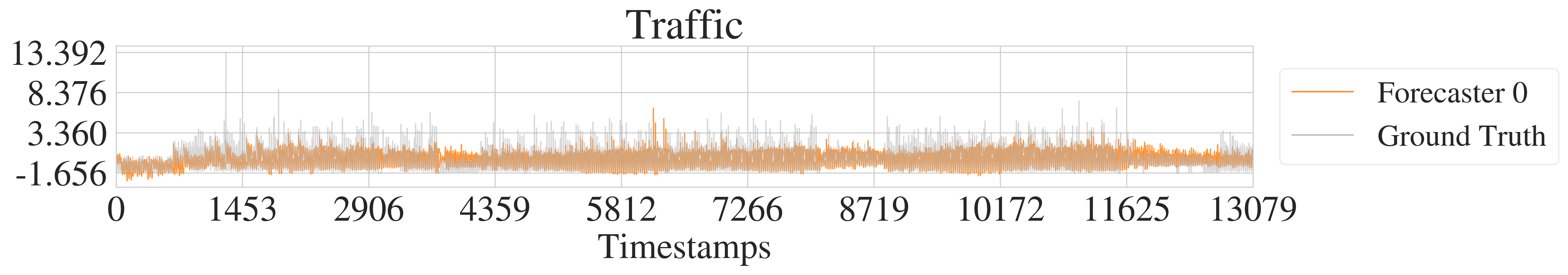}
    }\\[2pt]
    \subfigure[WTH, Lifecycle]{                    
        \includegraphics[width = 0.48\linewidth]{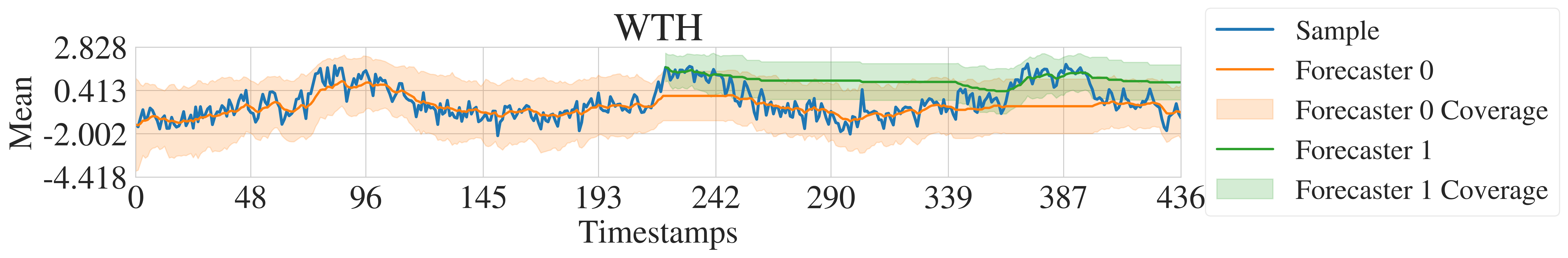}
    }\hfill
    \subfigure[WTH, Internal Prediction]{                    
        \includegraphics[width = 0.48\linewidth]{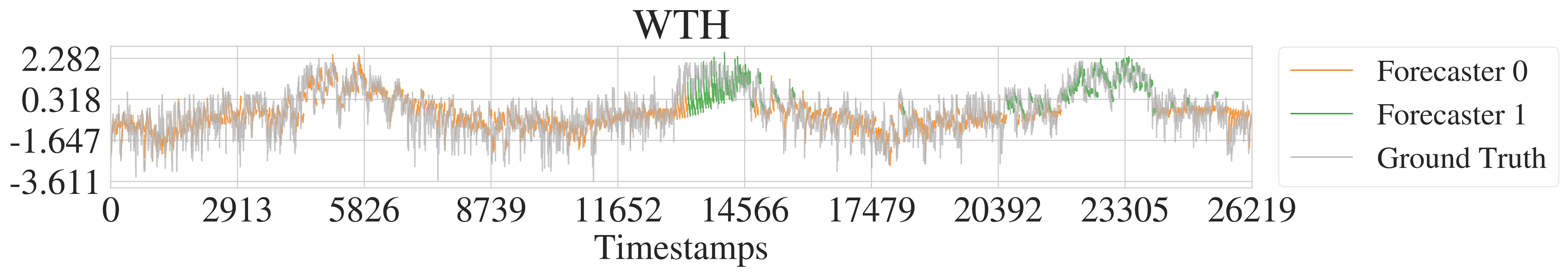}
    }
    \caption{Visualization of forecaster gene trajectories.}
    \label{fig:f_g_traj}
\end{figure*}

\end{document}